\documentclass[journal]{IEEEtran}  % Comment this line out if you need a4paper

 \usepackage{graphics} % for pdf, bitmapped graphics files
\usepackage{epsfig} % for postscript graphics files
\usepackage{mathptmx} % assumes new font selection scheme installed
\usepackage{times} % assumes new font selection scheme installed
\usepackage{amsmath} % assumes amsmath package installed
\usepackage{amsthm}
\usepackage{breqn}
\usepackage{amssymb}  % assumes amsmath package installed
\usepackage{epstopdf}
\usepackage{color}
\usepackage[noadjust]{cite}
\usepackage[ruled,vlined,linesnumbered]{algorithm2e}
\usepackage[normalem]{ulem}
\usepackage{soul}
\usepackage{booktabs}

\theoremstyle{definition}
\newtheorem{ass}{Assumption}

\newcommand{\spr}[1]{{\color{red} #1}}

\definecolor{orange}{rgb}{1,0.5,0}

\newcommand{\fnl}[1]{{\color{black} #1}}

\let\oldtitle\title
\renewcommand\title[1]{%
    \begingroup
        \providecommand{\ttlit}{}%
        \renewcommand{\ttlit}[1]{}%
        \providecommand{\titlenote}{}%
        \renewcommand{\titlenote}[1]{}%
        \hypersetup{pdftitle={#1}}%
        \def\thetitle{#1}%
        \pdfbookmark[0]{#1}{title}
    \endgroup
    \oldtitle{#1}%
}

\usepackage{calc}

\usepackage[shortcuts]{extdash}

\usepackage{graphicx}

\usepackage{booktabs}

\usepackage{verbatim}

\usepackage{amsmath}
\usepackage{psfrag}

\makeatletter
\def\citet{\@ifstar{\citetstar}{\citetnostar}}
\def\Citet{\@ifstar{\Citetstar}{\Citetnostar}}
\def\citetnostar{\@ifnextchar[{\squarecitet}{\simplecitet}}
\def\squarecitet[#1]{\@ifnextchar[{\twocitet[#1]}{\onecitet[#1]}}
\def\Citetnostar{\@ifnextchar[{\squareCitet}{\simpleCitet}}
\def\squareCitet[#1]{\@ifnextchar[{\twoCitet[#1]}{\oneCitet[#1]}}
\def\citetstar{\@ifnextchar[{\squarecitetstar}{\simplecitetstar}}
\def\squarecitetstar[#1]{\@ifnextchar[{\twocitetstar[#1]}{\onecitetstar[#1]}}
\def\Citetstar{\@ifnextchar[{\squareCitetstar}{\simpleCitetstar}}
\def\squareCitetstar[#1]{\@ifnextchar[{\twoCitetstar[#1]}{\oneCitetstar[#1]}}
\makeatother
\def\simplecitet#1{\citeauthor{#1}~\citep{#1}}
\def\onecitet[#1]#2{\citeauthor{#2}~\citep[#1]{#2}}
\def\twocitet[#1][#2]#3{\citeauthor{#3}~\citep[#1][#2]{#3}}
\def\simplecitetstar#1{\citeauthor*{#1}~\citep{#1}}
\def\onecitetstar[#1]#2{\citeauthor*{#2}~\citep[#1]{#2}}
\def\twocitetstar[#1][#2]#3{\citeauthor*{#3}~\citep[#1][#2]{#3}}
\def\simpleCitet#1{\Citeauthor{#1}~\citep{#1}}
\def\oneCitet[#1]#2{\Citeauthor{#2}~\citep[#1]{#2}}
\def\twoCitet[#1][#2]#3{\Citeauthor{#3}~\citep[#1][#2]{#3}}
\def\simpleCitetstar#1{\Citeauthor*{#1}~\citep{#1}}
\def\oneCitetstar[#1]#2{\Citeauthor*{#2}~\citep[#1]{#2}}
\def\twoCitetstar[#1][#2]#3{\Citeauthor*{#3}~\citep[#1][#2]{#3}}
\usepackage[cmex10]{mathtools}             % Loads and extends amsmath
\usepackage{amsfonts,amssymb}
\usepackage{mathrsfs}                      % Gives us \mathscr for \setset

\usepackage{varioref}
\labelformat{subfigure}{\thefigure\textup{(#1)}}
\labelformat{equation}{\textup{(#1)}}
\usepackage[%
        pagebackref=false,
        bookmarks=true,bookmarksopen=true,
        bookmarksnumbered=true,
        breaklinks=true,
        colorlinks=true,
        anchorcolor=blue,citecolor=blue,
        urlcolor=blue,linkcolor=blue,filecolor=blue,
        menucolor=blue,
    ]{hyperref}
\usepackage{doi}

\usepackage{subcaption}

\makeatletter
\@ifpackageloaded{hyperref}{%
\@ifpackageloaded{amsmath}{%
\newcommand{\AMShreffix}[1]{%
        \expandafter\let\csname AMShreffix#1\expandafter\endcsname%
                \csname #1\endcsname%
        \expandafter\renewcommand\csname #1\endcsname{%
                \@hyper@itemfalse\csname AMShreffix#1\endcsname}}
\AtBeginDocument{%
\AMShreffix{equation}
\AMShreffix{align}
\AMShreffix{alignat}
\AMShreffix{flalign}
\AMShreffix{gather}
\AMShreffix{multline}
}}{}}{}
\makeatother

\def\figurename{Fig.}
\begin{document}
\title{Information correlated L\'evy walk exploration and distributed mapping using a swarm of robots}

\author{Ragesh K. Ramachandran$^{1}$, Zahi Kakish$^{2}$, and Spring Berman$^{2}$% <-this % stops a space
\thanks{*This work was supported by the Arizona State University Global Security Initiative.}% <-this % stops a space
\thanks{$^{1}$Ragesh K. Ramachandran is with the Department of Computer Science, University of Southern California, Los Angeles, CA 90089, USA {\tt\small  rageshku@usc.edu}}
\thanks{$^{2}$Zahi Kakish, and Spring Berman are with the School for Engineering of Matter, Transport and Energy, Arizona State University, Tempe, AZ 85287, USA  {\tt\small zkakish@asu.edu, spring.berman@asu.edu}}%
}

\maketitle

\begin{abstract}
In this work, we present a novel distributed method for constructing an occupancy grid map of an unknown environment using a swarm of %inexpensive 
%\spr{resource-constrained}
robots with global localization capabilities and limited inter-robot communication. The robots explore the domain by performing L\'evy walks in which their headings are defined by maximizing the mutual information between the robot's  estimate of its environment in the form of an occupancy grid map and the distance measurements that it is likely to obtain when it moves in that direction. Each robot is equipped with laser range sensors, and it builds its occupancy grid map by repeatedly combining its own distance measurements with map information that is broadcast by neighboring robots. %Using \rag{\autoref{thm: consensus} proved in this paper}, 
Using results on average consensus over time-varying graph topologies,
we prove that all robots' maps will eventually converge to the actual map of the environment.
%We consider a swarm of robots that explore an unknown domain and 
%create local maps based on \spr{their} laser range sensor measurements. 
%Simultaneously, each robot updates its local occupancy grid map using its laser range sensor measurements and using map information broadcast by robots residing in the robot's neighborhood,
%thus eventually converging to the global map of the unknown environment. \spr{We prove this convergence(?)...} 
%We also propose an exploration strategy which combines information-theoretic concepts with L\'evy walks. 
In addition, we demonstrate that a technique based on topological data analysis, developed in our previous work for generating topological maps, can be readily extended for adaptive thresholding of occupancy grid maps.  
%\spr{(Why is this thresholding important?)} \rag{Computing the most likely map from an occupancy grid map is a computationally hard problem. Our way is an alternative to this (explained in Section V)} 
We validate the effectiveness of our distributed exploration and mapping strategy through a series of 2D simulations and multi-robot experiments.
\end{abstract}

\begin{IEEEkeywords}
Distributed robot system, mapping, occupancy grid map, information theory, algebraic topology
\end{IEEEkeywords}

% \fi
\IEEEpeerreviewmaketitle

\section{Introduction}
\label{sec:intro}

\IEEEPARstart{T}{echnological} advances in embedded systems such as highly miniaturized electronic components, as well as significant improvements in  actuator efficiency and sensor accuracy, are currently enabling the development 
%and deployment 
of large-scale robot collectives called  \textit{robotic swarms}. Swarms of autonomous robots have the potential to
%are definitely a viable alternative for 
perform tasks in remote, hazardous, and human-inaccessible locations, such as underground cave exploration, nuclear power plant monitoring, %natural calamity
disaster response, and search-and-rescue operations.
%in mines. 
In many of these applications, a map of the environment where the task should be performed is not available, and it would therefore be necessary to
%A common task involved in most of these applications is 
construct this map using sensor data from the robots.

A widely-used technique for solving this problem is simultaneous localization and mapping (SLAM) \cite{Thrun:2005:PR:1121596}, a class of algorithms for constructing a map of a domain through appropriate fusion of robot sensor data. The numerous algorithms that have been proposed for SLAM can be categorized into the following approaches. 
%\spr{Why do we need to describe all three approaches if we are not using SLAM?} \rag{we are not going deep about any of the techniques. We are mostly just introducing the kinds of  maps. I introduced them because 1) we use occupancy grid map 2) we talk about other kinds of maps in related work. }
\textit{Feature-based mapping} \cite{Smith:1988}, also known as \textit{landmark-based mapping}, is a method in which the environment is represented using a list of global positions of various features or landmarks that are present in the environment. Consequently, the algorithms in this category require feature extraction and data association.
\textit{Occupancy grid mapping} uses an array of cells to represent an unknown environment. This class of algorithms was first introduced in %Elfes et al. 
\cite{Elfes1989} and is the most commonly used method in robotic mapping applications. Occupancy grid maps are very effective at representing 2D environments, but their construction suffers from the curse of dimensionality. The cells in an occupancy grid map are associated with binary random variables that define the probability that each grid cell is occupied by an object.  
\textit{Topological mapping} \cite{Choset01topologicalsimultaneous} procedures generate a \textit{topological map}, which is a compact sparse representation of an environment. A topological map encodes all of an environment's topological features, such as holes that signify the presence of obstacles, and identifies collision-free paths through the environment in the form of a roadmap. This map is defined as %generally 
a graph in which the vertices correspond to particular obstacle-free locations in the domain and the edges correspond to collision-free paths between these locations. 

Size and cost constraints limit individual robots in a swarm from having sufficient sensing, computation, and communication resources to map the entire environment by themselves using existing SLAM-based mapping techniques. In addition, inter-robot communication in a swarm is constrained by restricted bandwidth and random link failures, and the mobility of the robots results in a dynamically changing, possibly disconnected communication network.
However, many existing multi-robot mapping strategies are extensions of single-robot techniques under centralized communication or all-to-all communication among robots. 
%An important contribution to this work is the 
These strategies include approaches based on particle filters, with the assumption that robots broadcast their local observations and controls \cite{Howard2006}, and extensions of the Constrained Local Submap Filter technique, in which robots build a local submap and transmit it to a central leader that constructs the global map \cite{Willams2002}.

%the generalization of approaches based on particle filters to multi-robot systems, with the assumption that robots broadcast their local observations and controls \cite{Howard2006}. 
%Another approach is the 

It is therefore necessary to develop mapping strategies for robotic swarms that can be executed in a decentralized fashion, and that can accommodate the aforementioned constraints on inter-robot communication.
%such as disconnected and dynamically changing communication networks, random link failures, and restricted bandwidth. 
%Distributed approaches are often required to design swarm robotic control strategies, due to the limited communication capabilities of individual robots 
%and the requirement to accommodate disconnected and switching robot interaction graph topologies, restricted communication bandwidth, and random link failures.
%Although distributed approaches have an additional level of complexity in their analysis and design, their gains outweigh the costs.
%There has been a large interest in swarm robotic applications in which teams of robots perform tasks in a cooperative manner. 
In addition, building a map in a distributed manner has the advantage that a team of robots can exhibit more efficient and robust performance than a single robot.
 %Hence, 
 %Researchers have pursued 
 There have been numerous efforts to develop distributed techniques for multi-robot mapping. Various multi-robot SLAM techniques that can be implemented on relatively small groups of robots are surveyed in \cite{Saeedi:2016}.   
 %A notable work on 
In \cite{Aragues2012}, a
distributed Kalman filter expressed in information matrix form is presented and formally analyzed as a solution to distributed feature-based map merging in dynamic multi-robot networks. This approach requires the robots have unique identifiers. 
% In this work, the authors present and formally analyze a fully distributed feature-based map merging problem in dynamic robot networks. In brief, the solution to this problem is a distributed Kalman filter expressed in information matrix form. 
There is also an ample literature on distributed strategies for occupancy grid mapping \cite{gonzalez2013,saeedi2015,jiang2017,Carpin2008}, which focuses on finding approximate relative transformation matrices among robots' occupancy grid maps and fusing the maps using various image processing techniques. 
%However, the matrix computations and map fusion procedures do not scale well with the number of robots.
{These strategies utilize matrix multiplication operations for combining maps from different robots. The most efficient algorithm for multiplying two $N\times N$ matrices (each representing an occupancy grid map) has a computational complexity of $\mathcal{O}(N^{2.373})$ \cite{d2005using}. 
% A common matrix operation used in these strategies for combining maps from different robots is  matrix multiplication. 
In comparison, as we show in \autoref{subsection:Computing mutual information for a single beam of the laser range sensor}, the complexity of our map-sharing procedure is $\mathcal{O}(N^{2})$ in typical cases,
%is linear in the size of the map in typical cases, 
and is linear in the number of robots in the worst case, which occurs when a robot communicates with all other robots during the map-sharing step. Thus, in typical cases, our approach has improved scaling properties 
%(how would you say ``scales better with the map size'' in a more formal way?) 
compared to other distributed methods for occupancy grid mapping.}
%\spr{(Originally, this paragraph said: ``However, the matrix computations and map fusion procedures do not scale well with the number of robots.'' Are the matrix computations used in the map fusion procedure? (i.e., are matrix computations and map fusion not two separate things to analyze?))} \rag{Matrix computations (multiplication, inverse) are generally used in map fusion, we do not use these matrix computations in our approach.}

% In contrast  to our work,  \cite{gonzalez2013,saeedi2015,jiang2017,Carpin2008} 

 %In our work, the robots update their occupancy grid maps based on their laser range sensor measurements and from the occupancy grid maps obtained from their neighbors. 
%%%

%Hence, 

	%%%%%%%%%%%%%%%%%%%%%%%%%%%%%%%%
	\begin{figure}
		\centering
		\includegraphics[width=0.95\linewidth]{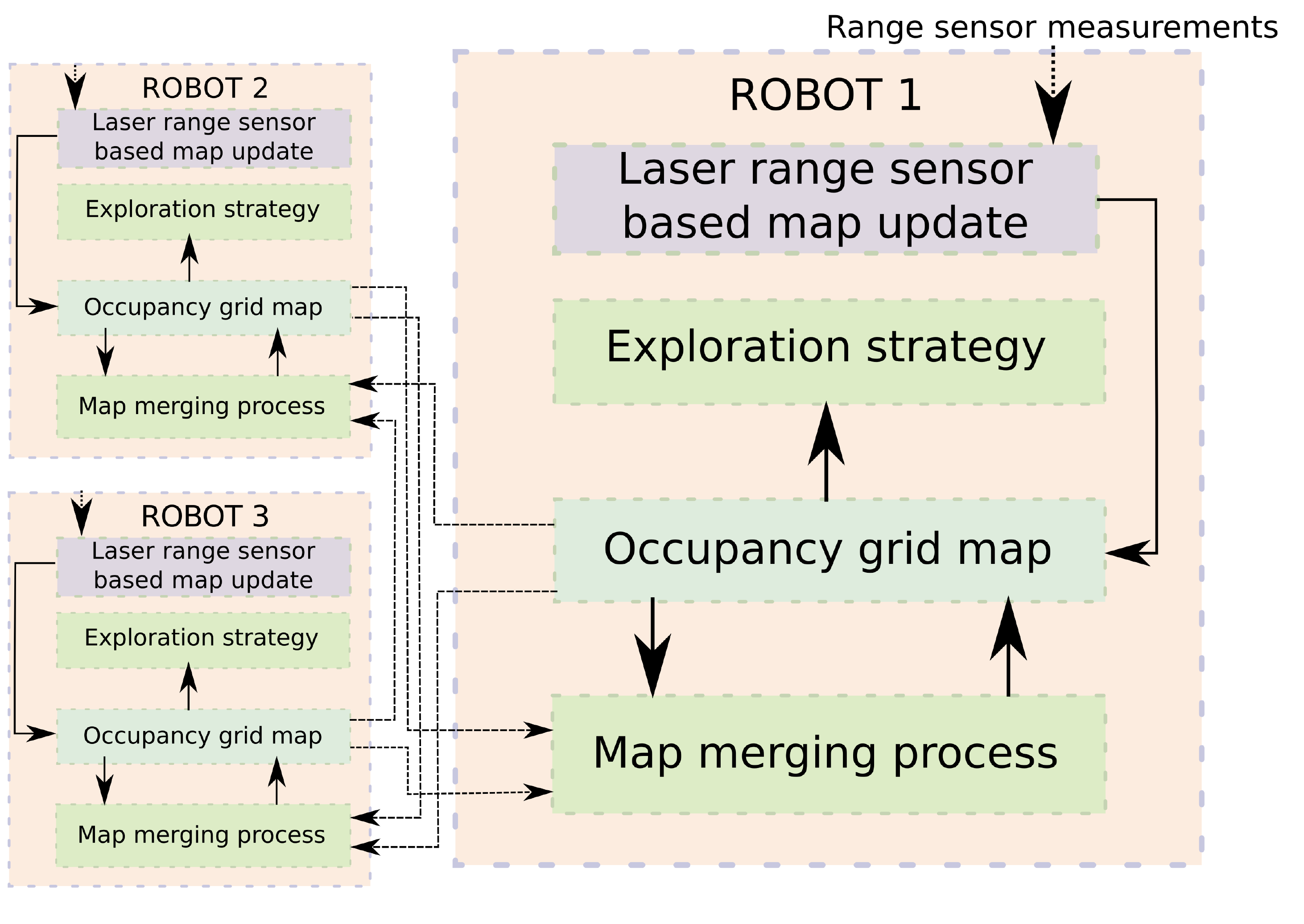}
		\caption{{A diagram of our mapping strategy. The arrows indicate information flow among components in our strategy.}} 
		\label{fig:schematic}       
		\vspace{-8mm}
	\end{figure}
	%%%%%%%%%%%%%%%%%%%%%%%%%%%%%%%%%%%

In this paper, we present a distributed approach to the mapping problem that is  scalable with the number of robots, relies only on local inter-robot communication, and does not require robots to have sophisticated sensing and computation capabilities. \autoref{fig:schematic} gives a schematic diagram of our mapping strategy. % or unique identifiers.
% from a cooperative task performance perspective. 
%, and the robots construct a global map of the environment by fusing their local observations in a logical manner. 
% bvAnother important aspect which is in conjunction with mapping of unknown environment addressed in this paper is exploration of the unknown environment by robots. 
%This is due to the fact that during the mapping process robots are challenged with an exploitation-exploration trade-off. In other words, 
In our approach, robots explore an unknown environment while simultaneously building an occupancy grid map online from their own distance measurements and from maps communicated by other robots that they encounter. As in most occupancy grid mapping strategies, we assume that each robot is either capable of accurately estimating its pose or is equipped with a localization device. We present a distributed algorithm for sharing and fusing occupancy grid maps among robots in such a way that each robot's map eventually converges to the same global map of the entire environment, which we prove using results on achieving consensus on time-varying graphs \cite{Kingston2006}. Our analysis of the distributed mapping algorithm is similar, in spirit, to the ones present in \cite{mangesius2018consensus}.
%Each robot observes a portion of an unknown environment and fuses its local observations with those of other robots to ultimately construct a global map of the entire environment.
The robots do not need to have unique identities that are recognized by other robots in order to implement this map-sharing algorithm. Since all robots ultimately arrive at a consensus on the map, this map can be retrieved from a single arbitrary robot in the swarm, making the strategy robust to robot failures. 
\begin{comment}
It would be an interesting future work to examine how our work translates into the scenario where robots are equipped with only weak localization capabilities, meaning pose information with bounded uncertainty. 
\end{comment}
%extracting stored occupancy map information from a few robots, and theoretically only one. In this regard, our approach is robust to robot failures.  
\fnl{We also introduce an exploration strategy that combines concepts from information theory \cite{Cover:2006:EIT:1146355} with L\'evy walks, a type of random walk with step lengths distributed according to a power law that has been used as a model for efficient search strategies by both animals and robots \cite{Fujisawa2013}. 
%which can also be employed as a swarm robotic exploration strategy. 
%empirically show that a swarm of robots which move according to our information correlated L\'evy walk 
%can enable the robotic swarm to can explore a domain faster than if they use standard L\'evy walks, thereby enabling them to map the domain faster.
We empirically show that this new exploration strategy, which we refer to as an \textit{information correlated L\'evy walk} (ICLW), enables a swarm of robots to explore a domain faster than if they use standard L\'evy walks, thereby enabling them to map the domain more quickly.}   %We defer the details on L\'evy walks and our exploration strategy to \autoref{sec:Exploration based on information Correlated Levy Walk}.} %\spr{(Possibly integrate this:) We combine random exploration with an information-based approach to obtain the advantages of both methods. We follow the information theoretic approach described in \cite{charrowI_ICRA15,Julian2013} and extend the idea to swarm robotic scenario by combining it with L\'evy walks.} 
Finally, we illustrate that a technique based on topological data analysis, used for generating topological maps in our previous work \cite{Ramachandran_Wilson_Berman_RAL_17}, can also be used for adaptive thresholding of occupancy grid maps with a slight modification. The threshold distinguishes occupied grid cells from unoccupied cells in the map by applying tools from \textit{algebraic topology} \cite{Hatcher2002}. A significant difference between our earlier topological approach and the one presented in this paper is our use here of cubical complexes \cite{Kaczynski04a}, a natural choice for grid maps, instead of simplicial complexes \cite{edelsbrunner2008persistent} We validate our mapping approach in 2D simulations of various environments with different sizes and layouts, using the mobile robot simulator \textit{Stage} \cite{Vaughan2008}. We also experimentally validate our approach with the commercially available \textit{TurtleBot 3 Burger} robots. 

\fnl{In contrast to many multi-robot strategies for exploration and mapping, our approach is an {\it identity-free}  strategy in that %In other words, 
the robots are indistinguishable from one another; i.e., robots are agnostic to the individual identities of other robots with which they interact. This property is useful for mapping approaches using robotic swarms, since it facilitates the scalability of the approach with the number of robots and 
%Moreover, an identity-free strategy 
is robust to robot failures during task execution.}
In addition, unlike many multi-robot mapping strategies, our approach is suitable for robotic platforms with limited computational capabilities, which is a typical constraint on robots that are intended to be deployed in swarms since each robot should be relatively inexpensive.
  Notably, although the robots identify other robots within their sensing range as obstacles, our approach filters out these ``false obstacles'' when generating the map of the environment from the robots' data. Also, our map merging approach guarantees that the map of each robot will eventually converge to the actual map of the domain, as long as  the robots' exploration strategy facilitates sufficient communication among the robots and coverage of the domain.

In our previous work, we developed different techniques for addressing swarm robotic mapping problems. In \cite{Ramachandran_Elamvazhuthi_Berman}, we present a method for mapping GPS-denied environments using a swarm of robots with stochastic behaviors. Unlike this paper, the approach in \cite{Ramachandran_Elamvazhuthi_Berman} employs optimal control of partial differential equation models of the swarm population dynamics to estimate the map of the environment. Although the method in \cite{Ramachandran_Elamvazhuthi_Berman} only requires robot data on encounter times with features of interest, it is limited in application to domains with a few sparsely distributed features. The methodologies discussed in our previous works \cite{Ramachandran_Wilson_Berman16} and \cite{Ramachandran_Wilson_Berman_RAL_17} estimate the number of obstacles in the domain and extract a topological map of the domain, respectively. Except for the procedure for adaptive thresholding of occupancy grid maps that is delineated in \autoref{sec:post processing OCG}, the work presented in this paper is novel and does not directly follow any of our previous works. Also, unlike our previous works, this mapping strategy computes the map online rather than offline.

In summary, the contributions of this paper are as follows:

\begin{itemize}
	\item We present a new  exploration strategy for robotic swarms that consists of L\'evy walks influenced by information-theoretic metrics.
	\item We develop a completely decentralized strategy by which a swarm of robots without unique identities can generate an occupancy grid map of an unknown environment. %This strategy does not require the robots to have unique identities.
    \item We prove that each robot's  occupancy grid map  asymptotically converges to a common map at an exponential rate.
    
	\item {We extend
	%demonstrate that 
	our topological data analysis-based technique for adaptive thresholding of occupancy grid maps \cite{Ramachandran_Wilson_Berman16, Ramachandran_Wilson_Berman_RAL_17,ramachandran_berman_2019} %described in our previous work 
	to domains represented as cubical complexes, instead of simplicial complexes as in our prior work.
	%using simplicial complex, can be recast as new method in which cubical complex replaces the simplicial complex.
	}
\end{itemize}

% the post-processing method was introduced and validated in the authors' previous work using filtrations constructed from a simplicial complex. we can extend it to filtrations that are constructed from a cubical complex. Since the fundamental unit of a cubical complex is a cube (a square in two dimensions), and the basic unit of an occupancy grid map is also a cube, the cubical complex is a more natural setting for constructing occupancy grid maps than a simplicial complex.

The remainder of the paper is organized as follows. \autoref{sec:problem statement} formally states the problem that we address and the associated assumptions. We describe our  swarm exploration strategy in \autoref{sec:Exploration based on information Correlated Levy Walk} and our occupancy grid mapping strategy and its analysis in \autoref{sec:occupancy grid map updation}. \autoref{sec:post processing OCG} outlines our TDA-based method for adaptive thresholding of occupancy grid maps and basic concepts from algebraic topology that are required to understand the technique. \autoref{sec:simulation} and \autoref{sec:experiments} present and discuss the results from simulations and robot experiments, respectively. The paper concludes with \autoref{sec:conclusion}. 

%\subsection{Related Work}
%\label{subsec:related work}

%Our exploration and mapping strategies use concepts from %%various fields such as 
%information theory and consensus-based distributed mapping. 
%We dedicate the following subsections to summarize the %respective 
%relevant literature on these topics. %separately.

%\subsubsection{Information Theoretic Exploration}
%\label{subsubsec:information theoretic}

% Our exploration strategy is based on the idea that, rather than moving in random directions as in L\'evy walk it would be more effective to direct the robots in direction with maximum information gain. 
%The methodology is described in detail in \autoref{sec:Exploration based on information Correlated Levy Walk}.

%\section{Related Work on Distributed Mapping}
%\label{sec:Distributed mapping}
 
%We also prove that the robots' occupancy grid maps eventually converge to a common map. 
%An important aspect of our distributed mapping strategy is that it is a robot label-free approach, meaning that the robots need not identify their neighbors based on any identification labels during communication. 

\begin{table}[!t]
	\centering
	\caption{Notation}
	\label{tab:notation}
	\begin{tabular}{@{}ll@{}}
		\toprule
		Symbol         & Description                    \\ \midrule
		$N_R$          & Number of robots               \\
		$R^i$          & Label of robot $i \in \{1,...,N_R\}$ \\
		$x^i$          & $x$-coordinate of $R^i$ w.r.t to global frame\\
		$y^i$          & $y$-coordinate of $R^i$ w.r.t to global frame\\
		$\theta^i$     & Orientation of $R^i$ w.r.t to global frame \\	
		$z^{i,a}_{\tau}$  & Measurement from $a^{th}$ laser range sensor of $R^i$ at time $\tau$\\
		$\mathbf{x}^i_\tau$ & Pose $[x^i ~y^i ~\theta^i]^T$ of $R^i$ at time $\tau$ \\
		$\mathbf{v}^i_\tau$     & Linear velocity vector of $R^i$ at  time instant $\tau$\\ 
		$\mathbf{z}^i_\tau$     & Vector of laser range sensor measurements of $R^i$ at time $\tau$\\
		$M^i$     & Occupancy grid map stored in $R^i$  \\
		$m^i_j$     & $j^{th}$ grid cell of occupancy map $M^i$\\
		 $\mathbb{P}_{m^i_j}$  & Probability that cell $m^i_j$ is occupied ({\it occupancy probability}) \\ 
		 $\mathbb{P}_{M^i}$   & Probability that occupancy map $M^i$ is completely occupied\\
		 $\mathbf{\bar{P}}_{M^i}$  & Set of occupancy probabilities $ \left\lbrace \mathbb{P}_{m^i_j} \right\rbrace_{j=1}^{|M^i|}$\\
		 $s^i$   & Constant speed of $R^i$ \\
		 $\Delta t$  & Time interval $[ \tau ~~ \tau + T]$ \\
		 $\mathbf{X}^i_{\Delta t}$  & Pose sequence $[\mathbf{x}^i_{\tau} ~\cdots~ \mathbf{x}^i_{\tau'} ~\cdots~ \mathbf{x}^i_{\tau + T}]$\\
		 $\mathbf{Z}^i_{\Delta t}$  & Measurement sequence $[\mathbf{z}^i_{\tau} ,\cdots , \mathbf{z}^i_{\tau'} , \cdots, \mathbf{z}^i_{\tau+T}]$\\  
 		 $\mathbf{I}[\mathbf{A}; \mathbf{B} |\mathbf{C}]$ & Mutual information between  random var.'s $\mathbf{A}$, %and 
 		 $\mathbf{B}$ given $\mathbf{C}$\\ 
 		 $<\mathbf{a}>$ & Arithmetic mean of the elements in vector $\mathbf{a}$\\
  		 $<\mathbf{a}>_{gm}$ & Geometric mean of the elements in vector $\mathbf{a}$\\
  		 $u(m_j^i, \mathbf{x}^i_k, \mathbf{z}^i_k) $ & Update rule that $R^i$ uses to assign $\mathbb{P}_{m^i_j}$  \\
  		 $\mathbb{G}(k)$ & Undirected robot communication graph at time step $k$ \\
  		 $\mathbb{V}$ & Vertex set of $\mathbb{G}(k)$,  $\{1, ..., N_R\}$ (robot indices) \\
  		 $\mathbb{E}(k)$ & Edge set of $\mathbb{G}(k)$ (pairs of robots that can communicate)\\
  		 $\mathbb{A}(k)$ & Adjacency matrix associated with $\mathbb{G}(k)$ \\
 		 $ \mathbb{N}_{k}^i  $  & Neighbors of $R^i$ (can communicate with $R^i$) at time step $k$\\ \bottomrule
	\end{tabular}
\end{table}

\section{Problem Statement}
\label{sec:problem statement}

We address the problem of estimating the map of an unknown domain $D \subset \mathbb{R}^d$ using distance measurements  acquired by a swarm of $N_R$ robots while exploring the domain. We consider bounded, closed, path-connected domains that contain static obstacles. Although in this paper we only address the case $d = 2$, it is straightforward to extend our procedure to the case $d = 3$. \autoref{tab:notation} lists 
the definitions of variables that we use throughout the paper for ease of reference.
%We propose an approach in which the robots explore the domain while simultaneously building a map online from their own distance measurements and from maps communicated by other robots that they encounter.
%its laser range sensor data and by communicating with other robots present locally.

\subsection{Robot capabilities}
We assume that the robots have the following capabilities. 
Each robot acquires noisy distance measurements using a laser range sensor such as a SICK LMS200 laser rangefinder \cite{LMS200}. Using this data, a robot can detect its distance to obstacles and other robots within its local sensing radius and perform collision avoidance maneuvers if needed. Each robot broadcasts its stored map information, and other robots that are within a distance $b_r$ of the robot can use this  information to update their own maps. We assume that each robot can estimate its own pose with no uncertainty. It is important to note that the robots are not equipped with any sensors that can distinguish between obstacles and other robots. The robots also do not have unique identifiers.

\subsection{Representation of the domain as an occupancy grid map}

Every robot models the unknown environment as an occupancy grid map, which does not require any {\it a priori} information about the size of the domain and can be expanded as the robot acquires new distance measurements \cite{Thrun:2005:PR:1121596}.
%\spr{An advantage of this type of map is that} the robots do not require any {\it a priori} information about the size of the domain, since they can augment the  map \spr{\sout{size} based on their \sout{laser sensor information} distance measurements} \cite{Thrun:2005:PR:1121596}. 
Each grid cell of an occupancy grid map is associated with a value that encodes the probability of the cell being occupied by an obstacle. Let $M^i_t$ denote the occupancy grid map stored by robot $R^i$ at time $t$, where $i \in \{1,...,N_R\}$. 
%\spr{Each robot in the swarm is denoted by $R^i$, where $i\in \{1,..., N_R\}$.}  
We specify that each robot discretizes the domain with the same resolution. 
At this resolution, a map of the entire domain $D \subset \mathbb{R}^2$ is discretized uniformly into $|D|$ grid cells, labeled $m^i_1,...,m^i_{|D|}$.
During the mapping procedure, each robot augments its map based on its own distance measurements and map information from nearby robots, effectively adding grid cells to its current map. The occupancy grid map of robot $i$ at time $t$ is represented by the grid cells $m^i_1,...,m^i_{|M^i_t|}$, where $|M^i_t|$ denotes the number of grid cells in the robot's map at time $t$. Henceforth, we will usually drop the subscript $t$ from $M_t^i$ to simplify the notation, with the understanding that the map $M^i$ depends on time.
%\spr{(You said that $|M^i|$ is the same for all robots, but can robots have different-sized maps since they explore different regions?)} %\rag{I wasn't precise there, the $|M^i|$ will be same eventually when each robot has the global map which is of the domain D. Probably its better to rephrase it.} 

Let $\mathbf{m}^i_j$, $j \in \{1,...,|M^i|\}$, be a Bernoulli random variable that takes the value 1 if the region enclosed by grid cell $m^i_j$  
%the $j^{th}$ grid cell of $M^i$, 
is occupied by an obstacle, and 0 if it is not. Thus, $\mathbb{P}(\mathbf{m}^i_j=1)$ is the probability that grid cell $m^i_j$ is occupied, called its {\it occupancy probability}.
A standard assumption for %most
occupancy grid maps is the independence of the random variables $\mathbf{m}^i_j$. As a result, the probability that map $M^i$  belongs to a domain which is completely occupied is given by $\mathbb{P}(M^i) = \prod_{j=1}^{|M^i|} \mathbb{P}(\mathbf{m}^i_j=1)$. For the sake of brevity, we will use the notation  $\mathbb{P}_{m^i_j} \equiv \mathbb{P}(\mathbf{m}^i_j=1)$ and $\mathbb{P}_{M^i} \equiv \mathbb{P}(M^i)$ throughout the paper.
%\spr{(Might change $\mathbb{P}_{m^i_j}$ to $\mathbb{P}_{\mathbf{m}^i_j}$)} 
%use $\mathbb{P}_{m^i_j}$ and $\mathbb{P}_{M^i}$ instead of $\mathbb{P}(\mathbf{m}^i_j=1)$ and $\mathbb{P}(M^i)$ respectively in afore coming sections. 
We also define the set $\mathbf{\bar{P}}_{M^i} = \left\lbrace \mathbb{P}_{m^i_j} \right\rbrace_{j=1}^{|M^i|}$, which is the collection of the occupancy probabilities of all grid cells in map $M^i$. Finally, the {\it entropy} $H(M^i)$ of the map $M^i$, which quantifies the uncertainty in the map, is defined as \cite{Thrun:2005:PR:1121596}: 
\begin{align}
\label{eqn:map entropy}
H(M^i) = \sum_{j=1}^{|M^i|}\sum_{k = \left \{ 0,1 \right \}}\mathbb{P}(\mathbf{m}^i_j=k)\log_2\left ( \mathbb{P}(\mathbf{m}^i_j=k) \right )
\end{align}

\subsection{Mapping approach and evaluation}
\label{subsec:framework and performance metric} 

Our mapping approach consists of the following steps.
%, \spr{which the robots execute in parallel}. 
All robots explore the domain simultaneously using the random walk strategy that is defined in \autoref{sec:Exploration based on information Correlated Levy Walk}. 
While exploring, each robot updates its occupancy grid map with its own distance measurements, 
broadcasts this map to neighboring robots, and then modifies its map with the maps transmitted by these neighboring robots using a predefined  %strategy or 
discrete-time, consensus-based protocol, which is discussed in
\autoref{sec:occupancy grid map updation}.  
%(3) broadcasting its occupancy grid map locally; and 
%(4) update its occupancy map using the robot's neighbors' occupancy maps based on a predefined  communication protocol. 
We prove that the proposed protocol %discrete time consensus based map modification protocol  
%for updating each robot's map 
guarantees that every robot's map will eventually converge to a common map. %Forthcoming Sections will explain these steps in great detail. 
A technique for post-processing the occupancy grid map based on topological data analysis (TDA) is presented in \autoref{sec:post processing OCG}.
We evaluate the performance of our mapping approach according to two metrics: (1) the percentage of the entire domain that is mapped after a specified amount of time, and (2) the entropy of the final occupancy grid map, as defined in \autoref{eqn:map entropy}.

%These metrics quantify the 
%effectiveness of our approach \spr{at covering the domain} 
%increasing the coverage 
%and reducing the uncertainty of the occupancy grid map obtained by exploring the unknown environment.

\section{Exploration Based On Information Correlated L\'evy Walks}
\label{sec:Exploration based on information Correlated Levy Walk}

%\spr{Also, since you are using techniques from various references, make sure to emphasize the novel contribution of the work in this section. What is new?  Is it the use of these techniques for computing the robot's heading? Is it the particular combination of the techniques?}

In this section, we describe the motion strategy used by robots to explore the unknown domain. 
%There are three general types of
Exploration strategies 
%in the robotics literature
for robotic swarms generally use {\it random}, {\it guided}, or {\it information-based} approaches
\cite{Kegeleirs2019,Thrun:2005:PR:1121596,gonzalez2013,fricke2016immune}. 
%Exploration techniques for robotic swarms often utilize stochastic processes such as
%rely on stochastic process based walks such as
Random exploration approaches are often based on Brownian motion (e.g., \cite{gelenbeabc1997autonomous,Dimidov2016,Wilson2014})  or L\'evy walks (e.g., \cite{DirafzoonBL15,Sutantyo2013,Fujisawa2013,pang2019swarm,nurzaman2011levy}), 
%The primary reason for choice of these 
%These stochastic exploration strategies 
which facilitate uniform dispersion of the swarm throughout a domain from any initial distribution. Moreover, these approaches do not rely on centralized motion planning or extensive inter-robot communication, which can scale poorly with the number of robots in the swarm. Information-based approaches, such as \cite{charrowI_ICRA15,Julian2013}, guide robots in the direction of maximum information gain based on a specified metric, which can increase the efficiency of exploration compared to random approaches. {\it Mutual information} (or {\it information gain}), a measure of the amount of information that one random variable contains about another \cite{Cover:2006:EIT:1146355},
%In information theory, {\it mutual information} is  
is a common metric used 
%in %all these works 
to assess the information gain that results from a particular action by a robot. This metric can be used to predict the increase in %amount of 
certainty about a state of the robot's environment that is associated with a new sensor measurement by the robot. 

%\rag{L\'evy walk performing robot does not require to identify or use any information from its neighbouring robots therefore it is identity-free. In fact every stochastic strategy (Brownian motion, correlated random walk) is identity-free, we picked L\'evy since it has the highest diffusivity property}
%; multi-robot coverage strategies with these requirements can scale poorly with the number of robots, and thus are unsuitable for swarms.
%Due to the ease of implementation and minimal external information requirement,  
%This is highly desirable in swarm applications involving exploration tasks. 
%Many multi-robot coverage strategies do not scale well for swarm robotic scenarios due to reliance on centralized planning or high levels of communication. 

We specify that each robot in the swarm %explores the domain by performing 
performs a combination of 
random and information-based exploration approaches, in order to benefit from the advantages of both types of strategies. %We refer to this exploration strategy as
%specify that each robot explores the domain by performing 
%an {\it information correlated L\'evy walk} (ICLW) and 
We describe the implementation of ICLW in this section. 
%, which we define in this section. 
{Our exploration strategy does not require the robots to identify or use any information from their neighboring robots while covering the domain; therefore, it is an identity-free strategy.} 

%is used to influence the exploration direction of the robot.  %Instead of moving in random directions as in random exploration strategies, it is more effective to move in the direction with maximum information gain based on some information metric.

To execute a L\'evy walk, a robot repeatedly chooses a new 
%uniformly random 
heading and moves at a constant speed \cite{zaburdaev2015levy} in that direction
over a random distance that is drawn from a heavy-tailed probability distribution function $p(l)$, of the form 
\begin{align}
\label{eqn:levy distribution}
p(l) \propto l^{-\alpha},
\end{align}
where $\alpha$ is the L\'evy exponent. \fnl{The case $1 < \alpha < 3$ signifies a
%, the L\'evy walk enters a 
scale-free \textit{superdiffusive} regime}, in which the expected displacement of a robot performing the L\'evy walk over a given time is much larger than that predicted by random walk models of uniform diffusion. This superdiffusive property  disperses the robots quickly toward unexplored regions. 

%In our exploration strategy, 
%each robot repeatedly moves in
% the direction that maximizes the mutual information between the robot's current
%locally stored 
%occupancy grid map and %its future laser range sensor measurements.
%the distance
%laser range sensor 
%measurements that it is likely to obtain when it moves in that direction. 
%These measurements are expected to decrease the overall entropy of the robot's map, defined in \autoref{eqn:map entropy}.

%\rag{The novelty in our exploration strategy lies in the way we extend the commonly used L\'evy walk based stochastic swarm robotic exploration strategy. Unlike in L\`evy walk based exploration, where the heading of a robot is selected randomly, our strategy computes the direction for a robot by solving an optimization problem and if the robot moves along this computed direction, it is expected to receive maximum information about its unknown environment. To the best of the authors' knowledge, this approach of extending the L\'evy walk based stochastic exploration strategy was never attempted before.}

In contrast to \textit{standard L\'evy walks} (SLW), in which the agent's heading is uniformly random, we define the heading chosen by the robot before each step in the L\'evy walk as the direction that maximizes the robot's information gain about the environment.
%Our exploration strategy utilizes partial information about the environment encoded in the map constructed by the robot.
%and the superdiffusive property of the L\'evy walk. %The proposed approach considers a group of robots exploring the unknown domain 
%The fundamental idea behind this exploration methodology is the utilization of 
%\spr{A robot chooses the heading} that maximizes information about the environment. 
%and move for random length sampled from a L\'evy walk distribution given in \autoref{eqn:levy distribution}. 
%The robots share their occupancy grid with the neighbors according to a predefined communication protocol allowing each robot to update their map based on the information from neighboring robots. 
%In order to find the  direction that maximizes the information gain about the unknown environment for the robots, 
This is computed as the direction that maximizes the mutual information between the robot's current occupancy grid map and 
the distance measurements that it is likely to obtain when it moves in that direction, based on the forward measurement model of a laser range sensor \cite{Thrun:2005:PR:1121596} over a finite time horizon. These measurements are expected to decrease the entropy of the robot's occupancy grid map, defined in \autoref{eqn:map entropy}.
%predicted reduction in the entropy of the robot's occupancy grid map, defined in \autoref{eqn:map entropy}, based on the forward measurement model of a laser range sensor \cite{Thrun:2005:PR:1121596} over a finite time horizon. 
%, which is a property of robots executing L\'evy walks \cite{zaburdaev2015levy}. 
Therefore, the computed robot heading is more likely 
to direct the robot to unexplored regions than a uniformly random heading.
The calculation of this heading is described in the following subsections.
%In this work, we are interested in finding  

\subsection{Laser range sensor forward measurement model}
\label{subsection:Laser range sensor measurement mode}

We assume that the laser range sensor of each robot $R^i$ has $N_l$ laser beams that all lie in a plane parallel to the base of the robot. The distance measurement obtained by the $a^{th}$ laser beam of robot $R^i$ at time $\tau$ is a random variable that will be denoted by $z^{i,a}_{\tau}$. The random vector of all distance measurements obtained by robot $R^i$ at time $\tau$ is represented as $\mathbf{z}^i_{\tau}=[z^{i,1}_{\tau}~\cdots~ z^{i,a}_{\tau}~\cdots~z^{i,N_l}_{\tau}]^T$.  

Define $s_{min}$ and $s_{max}$ as the minimum and maximum possible  distances, respectively, that can be measured by the laser range sensor. In addition, let $\delta$ denote the actual distance of an % the first
obstacle that is intersected by the $a^{th}$ laser beam of robot $R^i$. The Gaussian distribution function with mean $\mu$ and variance $\sigma^2$ will be written as $\mathcal{N}(\mu, \sigma^2)$.
 We define the probability density function of the distance measurement $z^{i,a}_\tau$, given the actual distance $\delta$, as the forward measurement model presented in \cite{charrowI_ICRA15},%\cite{charrowI_ICRA15, Smisek2011},
%The forward sensor model modeling the distribution over ranges from a single beam of a robot is assumed to be the following, adapted from \spr{(or is it exactly the model from these references?)} the \spr{model(?)} in \cite{charrowI_ICRA15, Smisek2011}:
\begin{align}
\label{eqn:forward sensor model}
\mathbb{P}(z^{i,a}_{\tau} ~|~ \delta) = 
\begin{cases}
\mathcal{N}(0, \sigma^2), & \ \delta  \leq s_{min}\\
\mathcal{N}(s_{max}, \sigma^2), & \ \delta  \geq s_{max}\\
\mathcal{N}(\delta, \sigma^2), & \text{otherwise,}
\end{cases} 
\end{align}
where $\sigma^2$ is the variance of the range sensor noise in the radial direction of the laser beam. 
Although this model 
%noise model is simple as it 
does not incorporate range sensor noise in the direction perpendicular to the laser beam, the experimental results in \cite{charrowI_ICRA15} and our results  in \autoref{sec:experiments} demonstrate that the model captures sufficient noise characteristics for generating accurate maps from the sensor data.

\subsection{{Robot headings based on mutual information}}

%Before we dive into the details of our exploration strategy, we are going to 
%We first introduce notation for mutual information between any two random variables $\mathbf{A}$ and $\mathbf{B}$.
The mutual information between two random variables  $\mathbf{A}$ and $\mathbf{B}$ is defined as the {\it Kullback-Leibler distance} \cite{Cover:2006:EIT:1146355}  between their joint probability distribution, %of the random variables $\mathbf{A}$ and $\mathbf{B}$, 
$\mathbb{P}(\mathbf{A},\mathbf{B})$, and the product of their marginal probability distributions, $\mathbb{P}(\mathbf{A})\mathbb{P}(\mathbf{B})$:
\begin{align}
\label{eqn:MI definition}
\mathbf{I}[\mathbf{A};\mathbf{B}] = \mathbb{K}\mathbb{L}\mathbb{D}\left(\mathbb{P}(\mathbf{A},\mathbf{B}) || \mathbb{P}(\mathbf{A})\mathbb{P}(\mathbf{B})\right)
\end{align}
This quantity measures how far $\mathbf{A}$ and $\mathbf{B}$ are from being independent. In other words, $\mathbf{I}[\mathbf{A};\mathbf{B}]$ quantifies the amount of  information that $\mathbf{B}$ contains about $\mathbf{A}$, and vice versa. For example, if $\mathbf{A}$ and $\mathbf{B}$ are independent random variables, then no information about $\mathbf{A}$ can be extracted from the outcomes of $\mathbf{B}$, and consequently, $\mathbf{I}[\mathbf{A};\mathbf{B}] = 0$. On the other hand, if $\mathbf{A}$ is a deterministic function of $\mathbf{B}$, then the entropies of both random variables are equal to the expected value of $-\log_2(\mathbb{P}(\mathbf{A}))$,
%each other 
and $\mathbf{I}[\mathbf{A};\mathbf{B}]$ is equal to this quantity, which is its maximum value.  
%\spr{This is the direction that the robot should take in order to obtain distance measurements that decrease the overall entropy of its map.}
%Since  \rag{maximizing the mutual information between the map and the distance measurements is equivalent to finding the direction a robot should move in order to obtain distance measurements which would decrease the overall entropy of the map, the solution to the maximization problem results in a heading that is more likely to direct the robot to unexplored regions compared to Brownian motion or L\'evy walks with uniformly random headings.} 

During each step in its random walk, every robot performs the following computations and movements. {A new step may be initiated either when the robot completes its previous step, or when the robot encounters an obstacle (or other robot) during its current step. 
%The robot computes an optimal sequence of velocity commands using \autoref{eqn:MI objective},  either when it encounters an obstacle (or other robots) or when it completes maneuvering a previously generated path. 
Suppose that the next step by robot $R^i$ starts at time $\tau$. At this time, the robot computes the duration $T$ of the step by generating a random distance based on the L\'evy distribution (\autoref{eqn:levy distribution}) and dividing this distance by its speed $s^i$, which is constant. Also at time $\tau$, the robot computes the velocity $\mathbf{v}^i_{\Delta t}$ that it will follow during the time interval $\Delta t \ \coloneqq\ [\tau ~~\tau + T]$.} This computation involves several variables, which we introduce here. The pose of robot $R^i$ at time $\tau$ is denoted by $\mathbf{x}^i_\tau$ (see \autoref{tab:notation}). We define a sequence of this robot's poses during the time interval $\Delta t$ as  $\mathbf{X}^i_{\Delta t} \ \coloneqq\ [\mathbf{x}^i_{\tau} ~\cdots~ \mathbf{x}^i_{\tau'} ~\cdots~ \mathbf{x}^i_{\tau + T}]$, where $\tau' \in \Delta t$. We also define $\mathbf{Z}^i_{\Delta t}\ \coloneqq\ [ \mathbf{z}^i_{\tau} ,\cdots , \mathbf{z}^i_{\tau'} , \cdots, \mathbf{z}^i_{\tau+T}]$ %\rag{it's a continuous set} 
as a set of random vectors modeling  laser range sensor measurements that the robot is expected to receive %during its motion 
as it moves during this time interval.
%, depending on its pose at the start of the time interval $\Delta t$. 
 % at each instant of time $\tau(i)$, 
\begin{comment}
We define the sequence of discrete poses of robot $R^i$ over a time interval $t(i)\ \coloneqq\ [\tau(i) ~\tau(i) + T^i_{\tau(i)}]$ as $\mathbf{X}^i_t\ \coloneqq\ [\mathbf{x}^i_{\tau(i)} ~\cdots~ \mathbf{x}^i_{\tau'(i)} ~\cdots~ \mathbf{x}^i_{\tau(i)+T^i_{\tau(i)}}]$.
\end{comment}
%The time horizon $T^i_\tau$  is given by the length of the current step in the robot's L\'evy walk divided by the robot's speed.
%In \autoref{eqn:MI objective}, $\mathbf{M^i}$ represents the occupancy grid map stored by $R^i$. 
%From \autoref{sec:problem statement}, $m^i_l$ denote the random variable which models the occupancy of the $l^{th}$ grid cell of $\mathbf{M^i}$.
%\autoref{subsection:Laser range sensor measurement mode} details the laser sensor measurement model using in this paper. 
%Formally speaking, 
%At time $\tau$, robot $R^i$ computes its velocity $\mathbf{v}^i_\tau$, which it will follow until time $\tau+T$,
%during a time interval $T^i_\tau$, 
At time $\tau$, robot $R^i$ calculates its velocity $\mathbf{v}^i_{\Delta t}$ as the  solution $^{*}\mathbf{v}^i_{\Delta t}$ to the following optimization problem, with the objective function defined as in \cite{charrowI_ICRA15,Julian_PhD_thesis} : 
\begin{align}
\label{eqn:MI objective}
^{*}\mathbf{v}^i_{\Delta t} = \underset{\left \|\mathbf{v}^i_{\Delta t} \right \|=s^i, ~ \angle \mathbf{v}^i_{\Delta t} \in[-\pi, \pi]}{\mbox{arg max}} \frac{\mathbf{I}[{M^i}; ~ \mathbf{Z}^i_{\Delta t}\ |\ \mathbf{X}^i_{\Delta t}]}{C(\mathbf{v}^i_{\Delta t})},
\end{align}
where $\mathbf{I}[{M^i}; ~ \mathbf{Z}^i_{\Delta t}\ |\ \mathbf{X}^i_{\Delta t}]$ represents the mutual information between the robot's occupancy grid map and its distance measurements given a sequence of the robot's poses. 
The term  $C(\mathbf{v}^i_{\Delta t})$ in \autoref{eqn:MI objective}  penalizes the robot for large deviations from its current heading when multiple velocities % command sequences 
%\rag{If we use only $\mathbf{I}[{M^i}; ~ \mathbf{Z}^i_{\Delta t}\ |\ \mathbf{X}^i_{\Delta t}]$ as the objective function we could get multiple velocities which maximizes the objective function. For example, at the start of the exploration every direction is equally likely as the robot has no information about the environment. The term $C(\mathbf{v}^i_{\Delta t})$ endures that the robot choose the heading closest to the current heading from the list of velocities that maximizes $\mathbf{I}[{M^i}; ~ \mathbf{Z}^i_{\Delta t}\ |\ \mathbf{X}^i_{\Delta t}]$ .}
generate different paths with the same mutual information. % reward. 
{We define $C(\mathbf{v}^i_{\Delta t}) = \|\mathbf{v}^i_{\Delta t} -\mathbf{v}^i_\tau \|/s^i+ \varphi$, where $\varphi$ is a small positive constant which ensures that $C(\mathbf{v}^i_{\Delta t})$ is strictly positive. For the computations in the paper, we chose $\varphi = \pi/72$ rad, or 2.5$^\circ$.}
% Check placement of this:
Based on the current occupancy grid map $M^i$ of robot $R^i$ and its set of expected poses $\mathbf{X}^i_{\Delta t}$ under its velocity command $\mathbf{v}^i_{\Delta t}$, $R^i$ can compute the probability distribution of its laser range sensor measurements using the forward measurement model \autoref{eqn:forward sensor model}.

{Before we elaborate on the details of solving the optimization problem described in \autoref{eqn:MI objective}, we present the intuition behind our argument that maximizing the objective function in \autoref{eqn:MI definition} tends
to direct a robot to unexplored regions according to its current map. By the definition of mutual information (see \autoref{eqn:MI formula expression}), $\mathbf{I}[{M^i}; ~ \mathbf{Z}^i_{\Delta t}\ |\ \mathbf{X}^i_{\Delta t}]$ quantifies the expected reduction in {the uncertainty of} a robot's map due to the sensor measurements received by the robot during its motion. In essence,  $\mathbf{I}[{M^i}; ~ \mathbf{Z}^i_{\Delta t}\ |\ \mathbf{X}^i_{\Delta t}]$ captures the expected amount of entropy reduced in the robot's map $M^i$ based on the likely measurements ($\mathbf{Z}^i_{\Delta t}$) that the robot would obtain if it followed a trajectory $\mathbf{X}^i_{\Delta t}$. Therefore, maximizing  $\mathbf{I}[{M^i}; ~ \mathbf{Z}^i_{\Delta t}\ |\ \mathbf{X}^i_{\Delta t}]$ for different trajectories ($\mathbf{X}^i_{\Delta t}$) yields the trajectory that {decreases} the uncertainty in the robot's map {by the largest amount.}
There is an equal chance for a grid cell in an unexplored region to be free or occupied by an obstacle. Since the entropy of a Bernoulli random variable is highest when its probability of occurrence is $0.5$, the occupancy probabilities of the grid cells in the unexplored region have the highest entropy (uncertainty). Thus, maximizing $\mathbf{I}[{M^i}; ~ \mathbf{Z}^i_{\Delta t}\ |\ \mathbf{X}^i_{\Delta t}]$ will generate trajectories that {direct  the robot } toward unexplored regions based on the information contained in its map.  }

\subsection{Computing mutual information }
\label{subsection:Computing mutual information for a single beam of the laser range sensor}

In this section, we describe the computation of the objective function in \autoref{eqn:MI objective} and discuss techniques for solving the associated optimization problem.
We first focus on computing $\mathbf{I}[M^i; z^{i,a}_{\tau}]$, the mutual information between the measurement $z^{i,a}_{\tau}$ obtained by the $a^{th}$ laser beam  of robot $R^i$ at  time  $\tau$ and the robot's current occupancy grid map $M^i$.  Grid cells in the map that do not intersect the beam do not to contribute to the mutual information. Hence, the task of computing $\mathbf{I}[M^i; z^{i,a}_{\tau}]$ reduces to computing $\mathbf{I}[\mathbf{c}^{i,a}_{\tau}; z^{i,a}_{\tau}]$, where $\mathbf{c}^{i,a}_{\tau}$  is the collection of Bernoulli random variables $\mathbf{m}_j^i$ modeling the occupancy of grid cells in the map of robot $R^i$ that are intersected by the $a^{th}$ beam at time $\tau$. 
%\spr{Should $\mathbf{c}^{i,a}_{\tau}$  be defined as the Bernoulli random variables $\mathbf{m}_j^i$ corresponding to these grid cells?} 
%$z^{i,a}_{\tau}$. 
%$\mathbf{I}[\mathbf{c}^{i,a}; z^{i,a}_{\tau}]$ 
This quantity is defined as \cite{Julian_PhD_thesis}:
\begin{align}
\label{eqn:MI definition2}
\mathbf{I}[\mathbf{c}^{i,a}_{\tau}; z^{i,a}_{\tau}] = \int_{ z\in z^{i,a}_{\tau}}\sum_{c\in\mathbf{c}^{i,a}_{\tau}}\mathbb{P}(c,z)\log_2\left( \frac{\mathbb{P}(c,z)}{\mathbb{P}(c)\mathbb{P}(z)}\right) dz,
\end{align}
where $\mathbb{P}(c,z)$ is the joint probability distribution of $c$ and $z$, and $\mathbb{P}(c)$ and $\mathbb{P}(z)$ are the probability distributions of the occupancy probabilities of the intersected grid cells and the range sensor distance measurements, respectively.  
%\sout{Note that here, we use $\mathbb{P}(z)$as  shorthand notation for $\mathbb{P}(z^{i,a}_{\tau} | \delta)$, defined in} \autoref{eqn:forward sensor model}.
We show in \hyperref[app: formula derivation]{Appendix~\ref*{app: formula derivation}} that $\mathbf{I}[\mathbf{c}^{i,a}_\tau; z^{i,a}_{\tau}]$ can be expressed as: 
\begin{align}
\label{eqn:MI formula expression}
\mathbf{I}[\mathbf{c}^{i,a}_{\tau}; z^{i,a}_{\tau}] = -\int_{  z^{i,a}_{\tau}}\mathbb{P}(z)\log_2( \mathbb{P}(z)) dz + K,
\end{align}
where $K=-\log(\sqrt{2\pi}\sigma) -0.5$.  Since $K$ is not a function of the map or the distance measurements, it does not affect the solution to the optimization problem in \autoref{eqn:MI objective} and therefore does not need to be included in this problem. %can be ignored during computation.

%can be ignored during optimization.

The effect of $\mathbf{c}^{i,a}_{\tau}$ on $\mathbf{I}[\mathbf{c}^{i,a}_\tau; z^{i,a}_{\tau}]$ is through the probability distribution $\mathbb{P}(z)$ in \autoref{eqn:MI formula expression}.
%As mentioned earlier, $\mathbf{c}^{i,a}$ is the list of grids cells intersected by a beam within the maximum range of the laser range sensor. 
We now compute this distribution.
From the forward measurement model \autoref{eqn:forward sensor model}, $\mathbb{P}(z)$
% Check this:
%\spr{(Okay?:) $\mathbb{P}(z)$ (which we use as shorthand notation for $\mathbb{P}(z^{i,a}_{\tau} | \delta)$)} \rag{$\mathbb{P}(z)$ is not a shorthand for $\mathbb{P}(z^{i,a}_{\tau} | \delta)$)} \spr{So how are they related?} \rag{ by equation 8 one is a conditional probability other is the full probability}
%the \spr{probability distribution of a laser beam's distance measurement} %over range of a beam 
is completely determined by the distance $\delta$ from the laser range sensor to the  closest occupied cell in $\mathbf{c}^{i,a}_{\tau}$. %\spr{\sout{which we denote as $e_p$.  Therefore, we can exclude from $\mathbf{c}^{i,a}_{\tau}$ all cells  whose distance from the laser range sensor exceeds $\delta$.}} 
Let $e_p$ denote a binary sequence of length $|\mathbf{c}^{i,a}_{\tau}|$ in which each of the first $p-1$ elements is  0 and the $p^{th}$ element is 1. The remainder of the elements in the sequence can be either 0 or 1. This sequence is a possible realization of $\mathbf{c}^{i,a}_{\tau}$, in which the first $p-1$ intersected grid cells are unoccupied, the $p^{th}$ cell is occupied, and the remaining cells may or may not be occupied.
%\rag{a set of $|\mathbf{c}^{i,a}_{\tau}|$ adjacent grid cells \spr{Are these cells all in $\mathbf{c}^{i,a}_{\tau}$? What do you mean by ``adjacent''? in which the  $p^{th}$ grid cell is the first occupied cell. }
For %convenience, consistency and
compactness  of notation, we define $e_0$ as the sequence in which all elements are 0; that is, %the realization of $\mathbf{c}^{i,a}_{\tau}$ in which 
no intersected grid cells are occupied.
%\rag{as the set of grid cells representing the case when}  none of the cells in $\mathbf{c}^{i,a}_{\tau}$ are occupied.
Then, we have that
\begin{align}
\label{eqn:probability of a range}
\mathbb{P}(z)=\sum_{p=0}^{|\mathbf{c}^{i,a}_{\tau}|}\mathbb{P}(z~|~\mathbf{c}^{i,a}_{\tau}=e_p)\mathbb{P}(\mathbf{c}^{i,a}_{\tau}=e_p)
\end{align}
%\rag{Since we do not care about the occupancy of the grids cells after the $p^{th}$ one in $e_p$, all realizations of $\mathbf{c}^{i,a}_{\tau}$ are contained in the set $\{e_0, e_1, \cdots e_p \cdots e_{|\mathbf{c}^{i,a}_{\tau}|}\}$.}
We direct the reader to \cite{charrowI_ICRA15,Julian2013} for a detailed description of such sensor models.

We can now extend our computation of the mutual information for a single distance measurement at a given time to $\mathbf{I}[M^i; \mathbf{Z}^i_{\Delta t}\ |\ \mathbf{X}^i_{\Delta t}]$, the mutual information for all distance measurements taken by robot $R^i$ over a sequence of times. Since the exact computation of this quantity is intractable, we adopt a common technique used in the robotics literature: we select several laser beams on the robot and assume that the measurements from these beams are independent of one another \cite{Kretzschmar2012, Julian2013}.  
We define $\mathcal{Z}^i_{\Delta t}$ as the set of distance measurements obtained at times $\tau' \in \Delta t$ from the selected laser beams on robot $R^i$, indexed by $a' \in \{1,...,N_l\}$.
Then, we can approximate $\mathbf{I}[M^i; \mathbf{Z}^i_{\Delta t}\ |\ \mathbf{X}^i_{\Delta t}]$ as the following sum over $\mathcal{Z}^i_{\Delta t}$:
\begin{align}
\label{eqn:MI multiple beams}
\mathbf{I}[M^i; \mathbf{Z}^i_{\Delta t}\ |\ \mathbf{X}^i_{\Delta t}] ~\approx \sum_{z^{i,a'}_{\tau'} \in \mathcal{Z}^i_{\Delta t}} \mathbf{I}[M^i; z^{i,a'}_{\tau'}]
\end{align}
%The dependence on pose of the robot in \autoref{eqn:MI multiple beams} is ignored for the sake of brevity.
In general, finding $\mathcal{Z}^i_{\Delta t} \subseteq \mathbf{Z}^i_{\Delta t}$ that best approximates the formula in \autoref{eqn:MI multiple beams} is an NP-hard problem \cite{charrowI_ICRA15}. Therefore, no approximation algorithm can be designed to find this $\mathcal{Z}^i_{\Delta t}$ 
%completes the task 
in polynomial time. 
%In spite of this, 
However, generating $\mathcal{Z}^i_{\Delta t}$ using greedy algorithms has shown promising results \cite{Kretzschmar2012, Julian2013, charrowI_ICRA15}.

Here, we specify the following procedure for a robot to solve the optimization problem in \autoref{eqn:MI objective} in order to compute the heading that maximizes its information gain.
The robot implements the greedy algorithm in \cite{Kretzschmar2012}, which selects the laser beams having an information gain above a predefined threshold, to find $\mathcal{Z}^i_{\Delta t}$ in \autoref{eqn:MI multiple beams}. The expression in \autoref{eqn:MI multiple beams} is used to approximate $\mathbf{I}[M^i; \mathbf{Z}^i_{\Delta t}\ |\ \mathbf{X}^i_{\Delta t}]$, where each term $\mathbf{I}[M^i; z^{i,a'}_{\tau'}]$ is equal to $\mathbf{I}[\mathbf{c}^{i,a'}_{\tau'}; z^{i,a'}_{\tau'}]$, defined in \autoref{eqn:MI formula expression}. The expression for $\mathbb{P}(z)$ in \autoref{eqn:MI formula expression} is computed from \autoref{eqn:probability of a range}, and \autoref{eqn:MI formula expression} is numerically integrated.  The robot employs a greedy algorithm to find a suboptimal solution to the optimization problem in \autoref{eqn:MI objective}. Specifically, the value of the objective function in \autoref{eqn:MI objective} is computed along different headings, and the robot selects the heading corresponding to the maximum value. {
%For the computations performed in our paper, we used eight different headings. 
It is straightforward to show that the computational cost for evaluating the objective function along a heading has an upper bound 
$\mathcal{O}(|M^i|\cdot|\mathcal{Z}^i_{\Delta t}|)$, where $|\mathcal{Z}^i_{\Delta t}|$ is the cardinality of $\mathcal{Z}^i_{\Delta t}$. {For the simulations and experiments in this paper,
the objective function was computed along eight different headings, 
%since we found that 
%computing the maximum value of the objective function along eight possible robot headings 
which enabled the robots to accurately map the domain without requiring an excessive computational load on each robot.}}

%\autoref{eqn:MI multiple beams} is used to approximate the value of $\mathbf{I}[M^i; \mathbf{Z}^i_{\Delta t}\ |\ \mathbf{X}^i_{\Delta t}]$ along each heading of interest. 

%which prompted us to use a similar technique of  selecting the laser beams having an information gain above a predefined threshold \cite{Kretzschmar2012}. 
%\rag{we have used Kretzschmar2012 idea of rejecting measurements with low information gain. } 

%\spr{It would help to summarize how you would compute the objective function from the approximations that you just derived; that is, cite the equations that the robot uses to compute the objective function. Would the robot then use a greedy algorithm to solve the optimization problem? Should there be a discussion of computational complexity?}

An alternate approach to solving the optimization problem in \autoref{eqn:MI objective} is to compute the gradient of the objective function and define the robot's heading as the direction of gradient ascent. 
However, since the computations are performed on a discrete occupancy grid map, it is not clear that the objective function has a well-defined gradient.  Although prior attempts have been made to compute the gradients of information-based objective functions under particular assumptions \cite{Julian_PhD_thesis,Charrow2015}, the gradient computation relies on numerical techniques such as finite difference methods.

\section{Occupancy grid map updates by each robot}
\label{sec:occupancy grid map updation}

%As precedented, robots explore the unknown domain by moving along the direction which gives maximum information gain based on its laser range sensor measurement model and the current knowledge of the environment stored in its occupancy grid map.  

While exploring the environment, each robot updates its %internally stored 
occupancy grid map based on its laser range sensor measurements and the occupancy grid map information broadcast by robots that are within a distance $b_r$.  In this section, we describe how robot $R^i$ updates $\mathbf{\bar{P}}_{M^i}$, the collection of occupancy probabilities of all cells in its map, using both its distance measurements and the sets $\mathbf{\bar{P}}_{M^{\hat{n}}}, \hat{n} \in \mathbb{N}_{\tau}^i$, where $\mathbb{N}_{\tau}^i$ denotes the set of robots that are within distance $b_r$ of robot $R^i$ at time $\tau$.  We present a discrete-time, consensus-based protocol 
 for modifying the occupancy map of each robot and prove that this protocol guarantees that all robots eventually arrive at a consensus on the map of the environment. %they are exploring. 
 As explained in \autoref{subsec:Protocol for occupancy grid sharing}, our  method for updating the occupancy map is resilient to false positives, meaning that even if a robot incorrectly assigns a high occupancy probability $\mathbb{P}_{m^i_j}$ to a free grid cell $j$ due to noise in its distance measurement, the impact of this noisy measurement on  $\mathbb{P}_{m^i_j}$ is eventually mitigated due to the averaging effect of our %discrete time consensus based 
 map modification protocol. Since occupancy grid mapping algorithms  require the robots' pose information,
 %pose information and the algorithms usually do not aid in robot pose estimation. 
 we assume that each robot can estimate its own pose using an accurate localization technique. 

\subsection{Occupancy map updates based on distance measurements}
% laser range sensor

%In \autoref{subsection:Laser range sensor measurement mode}, we discussed about the forward sensor model for the laser range sensor. 
The forward sensor measurement model \autoref{eqn:forward sensor model} represents the probability that a robot obtains a particular distance measurement given the robot's map of the environment and the robot's pose. The parameter $\delta$ in the  model can be computed from the robot's map and pose. Commonly used occupancy grid mapping algorithms \cite{Thrun:2005:PR:1121596,Elfes1989} use an inverse sensor measurement model to update the occupancy probabilities of the grid cells. %An inverse sensor measurement 
This type of model gives the probability that a grid cell is occupied, given the laser range sensor measurements and the pose of the robot. Although forward sensor measurement models can be easily derived for any type of range sensor, inverse sensor measurement models are more useful for occupancy grid algorithms \cite{Thrun:2005:PR:1121596}. Methods such as supervised learning algorithms and neural networks have been used to derive inverse sensor models based on a range sensor's forward model \cite{Thrun2003AR}. Pathak et al. \cite{Pathak2007} describe a rigorous approach to deriving an analytical inverse sensor model for a given forward sensor model. Although inverse sensor models derived from forward sensor models can be used to efficiently estimate an occupancy grid map, %of an environment, 
it is difficult to develop a distributed version of such models, %. This is mainly because either 
since either their computation is performed offline \cite{Thrun2003AR} or  
 the mapping between the forward and inverse sensor models is nonlinear \cite{Pathak2007}. These difficulties preclude us from exploiting these techniques in our mapping approach.  

Instead, we propose a heuristic inverse range sensor model for which a distributed version can be easily derived. We specify that each robot estimates its pose and obtains distance measurements at discrete time steps, to reflect the fact that sensor measurements are recorded at finite sampling rates. Let $\mathbf{x}^i_k$ denote the pose of robot $R^i$ at time step $k$, and let $\mathbf{z}^i_k$ be the vector of its distance measurements at this time step. Our inverse sensor model, which we refer as an {\it update rule}, is %basically 
a function $u: (m^i_j, \mathbf{x}^i_k,\mathbf{z}^i_k) \rightarrow [0, 1]$. This function assigns an occupancy probability to grid cell $m_j^i$ based on the robot's pose and all of its distance measurements at time step  $k$. 
% which helps to reason about the probability of occupancy of a grid cell in regard to the pose and all distance measurements received by a robot at a particular time step. 
%that maps the tuple $(m^i_j, \mathbf{x}^i_k,\mathbf{z}^i_k)$  to the interval $[0, 1]$.
%and is denoted as $u(m^i_j, \mathbf{x}^i_k,\mathbf{z}^i_k)$. 
%Note that,  $u(m^i_j, \mathbf{x}^i_k,\mathbf{z}^i_k)$ is defined to admit inputs at discrete time step to reflect the fact that in reality sensor measurements are obtained at finite sampling rates. 
Robot $R^i$ uses this function to modify  $\mathbf{\bar{P}}_{M^i}$ based on its distance measurements. We define the update rule in terms of a function $l: (m^i_j, \mathbf{x}^i_k,z^{i,a}_{k}) \rightarrow [0, 1]$,
%, which we call the {\it occupancy probability assignment function} for a single laser beam. 
which assigns an occupancy probability to grid cell $m^i_j$ based on the robot's pose and its $a^{th}$ laser beam's distance measurement at time step $k$. 
%define a single laser beam occupancy probability assignment function: $l(m^i_j, \mathbf{x}^i_\tau,z^{i,a}_{\tau})$. 
%This function assigns a value from the interval $[0,1]$ to the grid cell $m^i_j$ which can be interpreted as the probability of occupancy of the grid cell $m^i_j$ based on the distance measurement $z^{i,a}_{\tau}$. 
The function can be applied only to those grid cells $m^i_j$ that are intersected by the $a^{th}$ beam at time step $k$.
%that contain any part of the horizontal projection of the path of the laser beam which yields the measurement $z^{i,a}_{\tau}$. Horizontal projection of a path in the current context means projection of the path to the plane of the occupancy grid map. 
%Therefore, it is important to check this condition before evaluating the function $l(m^i_j, \mathbf{x}^i_\tau,z^{i,a}_{\tau} )$. 

We define $l$ as one of two functions, $l_r$ and $l_u$, depending on whether the robot estimates that its $a^{th}$ laser beam is reflected ($l_r$) or not reflected ($l_u$).
These functions depend on $s^a_{m^i_j}$, the distance from the center of cell $m^i_j$ to the $a^{th}$ laser range sensor of robot $R^i$, and constants $p_a$, $p_f$, and $p_{hit}$. %\spr{(Where do these constants come from?)} \rag{ we assign them to the model} 
The functions $l$, $l_r$, and $l_u$ are defined as follows:

%\spr{(Note that I added the superscript $a$ to $s_{m_j^i}$. Also, I think that the $s_{max}$ should be dropped as an argument of $l_u$, since it's a constant.)}

%\spr{(How did you come up with the shapes of these functions? \rag{my creativity :), idea was if  a beam is not reflected then we have information about its occupancy ($p_a=0.5$)  } Why does the occupancy probability linearly increase with distance of the cell from the sensor?)}

\begin{figure}[!t]
	
\begin{tabular}{cc}
	\centering	
	\hspace{-3mm}
\subcaptionbox{unreflected model\label{subfig:nonreflectance}}{\includegraphics[width=.40\linewidth, height=.40\linewidth]{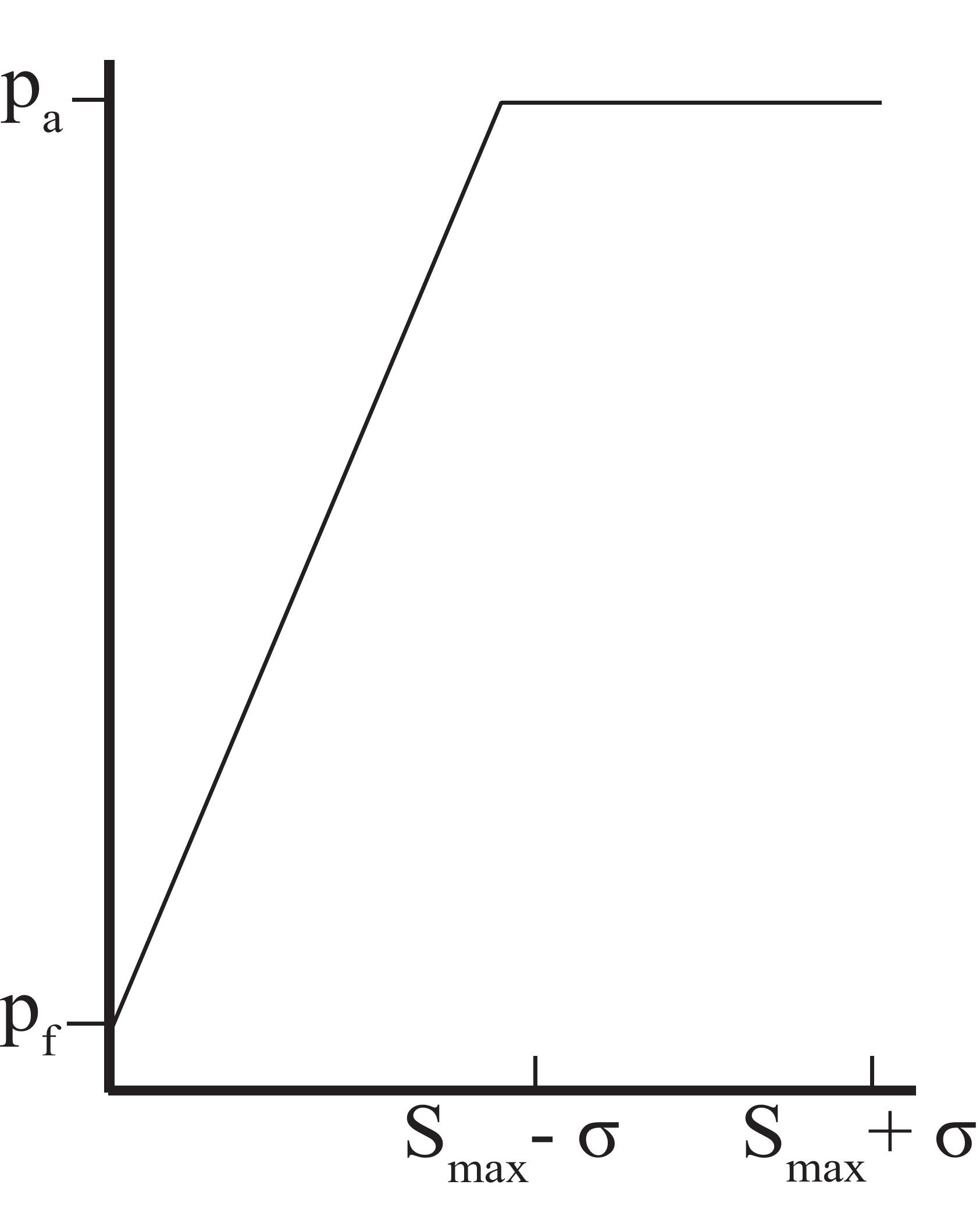}
}
&
\subcaptionbox{reflected model\label{subfig:reflectance}}{\includegraphics[width=.40\linewidth, height=.40\linewidth]{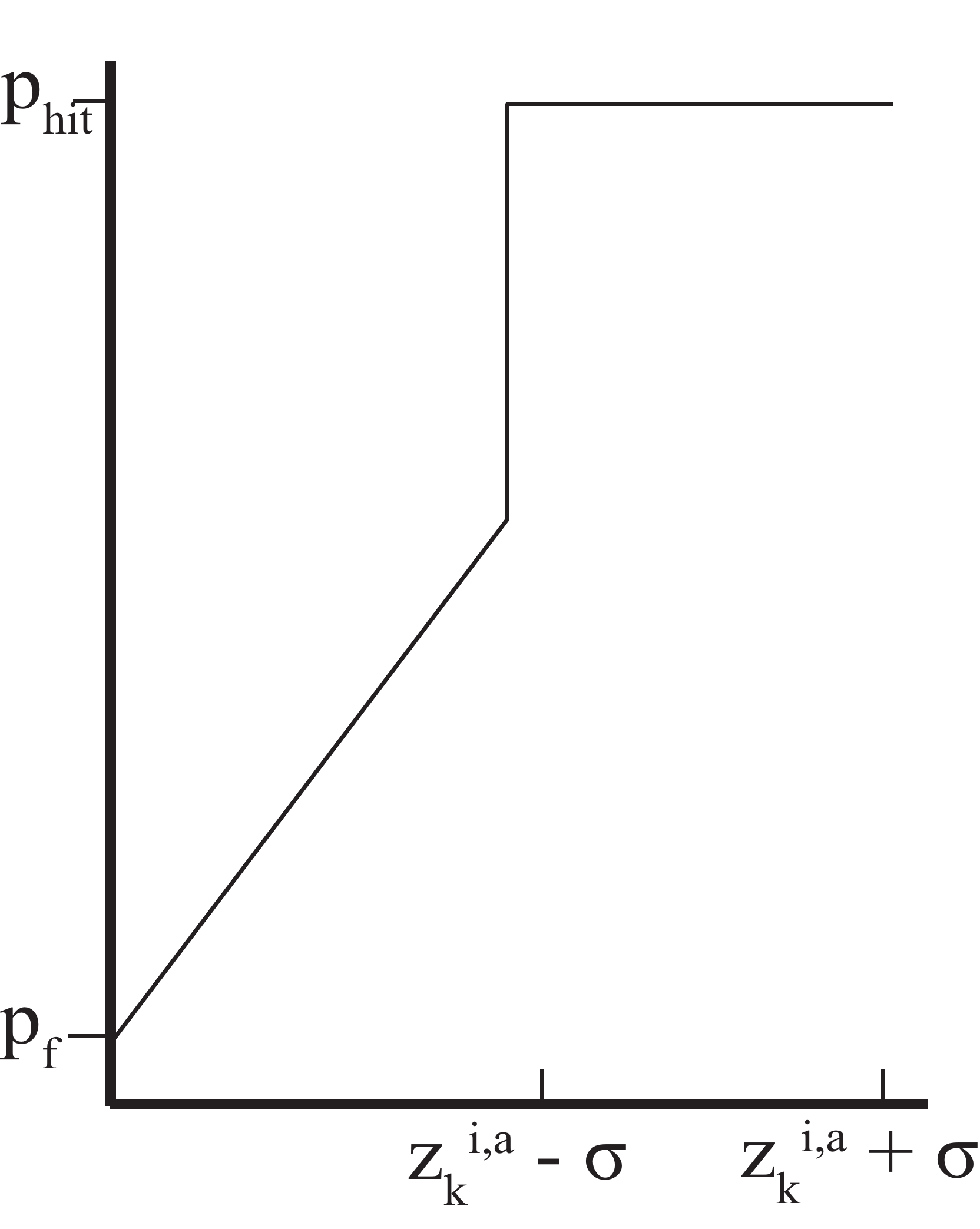}
}
\end{tabular}

\caption{%Sub-functions used in $l(m^i_j, \mathbf{x}^i_\tau,z^{i,a}_{\tau} )$ when the laser beam is (a) not reflected; (b) reflected. 
Illustrative plots of the functions (a) $l_u$ and (b) $l_r$.
%, defined in \autoref{eqn:reflectance model} and \autoref{eqn:unreflectance model}, respectively.}
The $x$-axis in both plots
%of both \autoref{subfig:nonreflectance} and \autoref{subfig:reflectance} 
measures $s^a_{m^i_j}$, the distance between the $a^{th}$ laser range sensor of robot $R^i$ and any grid cell $m^i_j$ that intersects the beam from this laser, which yields the distance measurement $z^{i,a}_{k}$ at time step $k$.  
%is traced by the laser beam corresponding to the measurement $z^{i,a}_{\tau}$. %(a) Sub-function used if the laser beam is not reflected. (b) Sub-function used if the laser beam is reflected.
}  
\label{fig:sub function}
\end{figure}

%length of the horizontal projection of the line segment connecting the center of $m^i_j$ to the laser range sensor of $R^i$. 
%$l(m^i_j, \mathbf{x}^i_\tau,z^{i,a}_{\tau} )$ has two sub-functions: 
%reflected and unreflected, each sub-function is triggered based on whether the laser beam corresponding to $z^{i,a}_{\tau}$ was reflected or not.  
%We define the single laser beam occupancy probability assignment function  as:
	\begin{align}
	\label{eqn: inverse model update rule}
	l(m^i_j, \mathbf{x}^i_k,z^{i,a}_{k} ) = 
	\begin{cases}
	l_r(s^a_{m^i_j},z^{i,a}_{k}) & \ z^{i,a}_{k}  \leq s_{max}-\sigma\\
	l_u(s^a_{m^i_j}) & \ z^{i,a}_{k}  > s_{max} -\sigma\\
	\end{cases} 
	\end{align}
	
	\begin{align}
	\label{eqn:reflectance model}
	l_r(s^a_{m^i_j},z^{i,a}_{k}) = 
	\begin{cases}
	\frac{p_a-p_f}{s_{max}}s^a_{m^i_j} + p_f & s^a_{m^i_j} <  z^{i,a}_{k}-\sigma\\
	p_{hit} &  z^{i,a}_{k}-\sigma \leq s^a_{m^i_j} \leq z^{i,a}_{k}+\sigma
	\end{cases}
	\end{align}
	
	\begin{align}
	\label{eqn:unreflectance model}
	 l_u(s^a_{m^i_j}) = 
	\begin{cases}
	\frac{p_a-p_f}{s_{max}}s^a_{m^i_j} + p_f & s^a_{m^i_j} <  s_{max}-\sigma\\
	p_a &  s_{max}-\sigma \leq s^a_{m^i_j} \leq s_{max}+\sigma\\
	\end{cases}
	\end{align}
%\spr{Since $s_{max}$ is a constant, I don't think that it should be an argument of $l_u$}
%$R^i$ assumes that the laser beam corresponding to the measurement $z^{i,a}_{\tau}$ is a reflected laser beam if  $z^{i,a}_{\tau} \leq s_{max}-\sigma$. 
\autoref{fig:sub function} illustrates the functions $l_r$ and $l_u$ that are defined in \autoref{eqn:reflectance model} and \autoref{eqn:unreflectance model}, respectively.  

{Our heuristic inverse sensor model can model laser range sensors for which the noise in the range measurements is predominantly in the radial direction of the laser beams. We defined our inverse sensor model  based on the physically realistic assumption that given a range measurement, the grid cells closer to the robot have a lower occupancy probability than the grid cells farther away. In our model, if a laser range measurement $z^{i,a}_{k}$ is obtained, then the occupancy probability of each grid cell along the radial direction of the laser beam increases with its distance from the sensor, until this distance is within $\sigma$ units of the measurement $z^{i,a}_{k}$. The model assumes that a grid cell whose distance from the sensor is in the interval $[z^{i,a}_{k}-\sigma, z^{i,a}_{k}+\sigma]$ belongs to an obstacle. On the contrary, if no laser range measurement is obtained, then the model assumes that a grid cell at a distance less than $s_{max}-\sigma$ from the sensor is likely to be unoccupied by an obstacle. We selected parameter values for the model for which the robots generated reasonably accurate maps in our simulations and  experiments. Alternatively, one could estimate the parameter values by fitting  the model to real measurements from laser range sensors. }

Given the function $l$ in \autoref{eqn: inverse model update rule}, 
       % the single laser beam occupancy probability assignment function, 
        we  now specify the update rule function $u$ according to the following procedure.
        \begin{comment}
        We describe the mapping from the input to the output of the update rule in the pseudocode in Algorithm \ref{algo:update rule}.
        \end{comment}
        %The procedure described in \autoref{algo:update rule} is self explanatory and can be easily followed.
%We set $u(m^i_j, \mathbf{x}^i_k,\mathbf{z}^i_k) = 1$ if the occupancy probability of $m^i_j$ can be inferred from $\mathbf{x}^i_k$ and $\mathbf{z}^i_k$. (So if it can't be inferred, then $u < 1$? Why?)
We define $\zeta$ as the set of distance measurements $z_k^{i,a}$ at time step $k$ that are recorded by laser beams that either intersect grid cell $m^i_j$ or are reflected by an obstacle that covers $m^i_j$.
%Steps 2 to 4 in the algorithm identify and store the distance  measurements in a set $\zeta$, for which the corresponding laser beams' path's horizontal projection intersects with the input grid cell $m^i_j$. 
If $\zeta = \emptyset$, 
%meaning that the cell $m_j^i$ does not intersect any laser beam from $R^i$ (and therefore 
meaning that none of the measurements in $\mathbf{z}^i_k$ provide any information about $m^i_j$, then we set $u(m^i_j, \mathbf{x}^i_k,\mathbf{z}^i_k ) = 1$ to make the update rule well-defined for such grid cells. 
%any %input 
%grid cell, we  define $u(m^i_j, \mathbf{x}^i_k,\mathbf{z}^i_k ) = 1$ if none of the measurements in $\mathbf{z}^i_k$ provide any information about $m^i_j$. %(Step 5). 
Otherwise, we set  $u(m^i_j, \mathbf{x}^i_k,\mathbf{z}^i_k)$ to the maximum value of $l(m^i_j, \mathbf{x}^i_k,z)$ over all measurements $z \in \zeta$.
%Finally, Step 5 of the algorithm returns the maximum value of $l(m^i_j, \mathbf{x}^i_k,z )$ when evaluated over all elements $z$ in set $\zeta$.

\subsection{Consensus-based occupancy grid map sharing}
 \label{subsec:Protocol for occupancy grid sharing}
 
 In this section, we describe a discrete-time, consensus-based protocol by which each robot modifies its occupancy map using the maps that are broadcast by neighboring robots. We prove that each robot's map asymptotically converges to the occupancy grid map that best represents the domain by using analysis techniques for linear consensus protocols over time-varying graphs, which have been well-studied in the literature \cite{Olfati-Saber2004, Ren2004, Moreau2005,Kingston2006}. The main results of these works assume the existence of a time interval over which the union of the graphs 
%during which the union of communication graphs 
contains a spanning tree, which is required  in order to reach consensus.  % \rag{if the union of graph is connected then it contains a spanning tree}
 In particular, we apply results from \cite{Kingston2006} on {\it average consensus} over time-varying graph topologies in a discrete-time setting.
 
 We begin with an overview of average consensus over time-varying graphs, using graph-theoretic notation from \cite{GodsilRoyle2001}. 
 %to explain the graph-theoretic concepts.
 %This forms the basis for our discrete-time, consensus-based protocol for map modification. 
 %We use notation from \cite{GodsilRoyle2001} to explain the associated graph-theoretic concepts.
 We define $\mathbb{G}(k)\ =\ \left( \mathbb{V}, \mathbb{E}(k) \right)$ as an undirected time-varying graph with $n$ vertices, $\mathbb{V} = \{1,...,n\}$, and a set of undirected edges $\mathbb{E}(k)$ at time step $k$. 
 In our scenario, $\mathbb{G}(k)$ defines the communication network of the robots at time step $k$, in which the vertices represent the robots, $\mathbb{V} = \{1,...,N_R\}$, and each edge $(i, \hat{n}) \in \mathbb{E}(k)$ indicates that robots $R^i$ and $R^{\hat{n}}$ are within broadcast range of each other at time step $k$ and can therefore exchange information.  
  
 %Let $\mathbb{G}(k)\ =\ \left( \mathbb{V}, \mathbb{E}(k) \right)$ be the undirected time-varying communication graph.  
 %The nodes represent robot indices, $\mathbb{V} = \{1, ..., i, ..., N_R\}$. 
 %At a time step $k$, if two robots $R^i,R^{\hat{n}}$ are in broadcasting range of each other and exchange information, then there is an edge $(i,\hat{n}) \in \mathbb{E}(k)$.
 Let $\mathbb{A}(k) = [a_{ij}(k)] \in \mathbb{R}^{n \times n}$ be the adjacency matrix associated with graph $\mathbb{G}(k)$ at time step $k$, where 
$ a_{ij}(k)$ denotes the element in the $i^{th}$ row and $j^{th}$ column of $\mathbb{A}(k)$. In this matrix, 
$a_{ij}(k) \ne 0$ if and only if an edge exists between vertices $i$ and $j$ at time step $k$, and $a_{ij}(k) = 0$ otherwise. 
 %Let $\mathbb{A}[k] = [a_{ij}(k)]$ be the adjacency matrix associated with the time-varying undirected communication graph $G(k)={V, E(k)}$  at time step $k$. %$V = {1, ..., n}$ is the vertex set of $G$  
 %edge from node $i$ to $j$ exists at time step $k$ if and only if $a_{ij} \ne 0$. 
The set of neighbors of vertex $i$ at time step $k$, defined as $\mathbb{N}_{k}^i  = \left\lbrace \hat{n} ~|~ (i,\hat{n}) \in \mathbb{E}(k),  ~i \ne \hat{n}\right\rbrace$, contains the vertices $j$ for which $a_{ij}(k) \neq 0$.
 %can be determined by the nonzero elements in row $i$ of $\mathbb{A}[k]$. 
 % We formally define $ \mathbb{N}_{k}^i$, the neighbor robots' indices set of $R^i$:  $ \mathbb{N}_{k}^i  = \left\lbrace \hat{n} | (i,\hat{n}) \in \mathbb{E}(k),  i \ne \hat{n}\right\rbrace$.
 Suppose that at time step $k$, each vertex $i$ is associated with   %stores 
 a real scalar variable %information 
 $x_i(k)$. At every time step, the vertex updates its value of $x_i(k)$ to a weighted linear combination of its neighbors' values and $x_i(k)$, where the weights are the corresponding values of $a_{ij}(k)$. Then the vector $\mathbf{x}(k) = [x_1(k)~...~x_n(k)]^T$ evolves according to the discrete-time dynamics $\mathbf{x}(k+1) = \mathbb{A}(k)\mathbf{x}(k)$. If $\lim_{k \rightarrow \infty} x_j(k) = \frac{1}{n}\sum_{i=1}^{n} x_i(0)$ for all $j \in \mathbb{V}$, then the vertices are said to have achieved {\it average consensus}. It is proved in  \cite[Theorem 1]{Kingston2006} that the dynamics of $\mathbf{x}(k)$ converge asymptotically to average consensus if $\mathbb{A}(k)$ is a doubly stochastic matrix, meaning that each of its rows and columns sums to 1, and if there exists a time interval for which the union of graphs over this interval is connected. We will use these results to prove an important result on our protocol for occupancy map sharing.

We now define the protocol by which robot $R^i$ updates $\mathbf{\bar{P}}_{M^i}$, the occupancy probabilities of all grid cells in its map $M^i$, based on its neighbors' occupancy maps, its current occupancy map, and its distance measurements. %\rag{along with its current occupancy probabilities of all grid cells in its map and its distance measurements}. 
It is important to note that the maps do not contain any information about the robots that broadcasted them.
%each robot receives do not contain any information about the robots that broadcasted those maps.
We specify that the occupancy probability $\mathbb{P}_{m^i_j}(k)$ of every grid cell $m_j^i$ in the map $M^i$ at time step $k$ is updated at the next time step as follows:
% The following equation rigorously formulates our  discrete-time, consensus-based map modification protocol when applied to the map's grid cell $m^i_j$:
 \begin{align}
 \label{eqn:grid cell update rule}
 \mathbb{P}_{m^i_j}(k + 1) = u(m^i_j, \mathbf{x}^i_k,\mathbf{z}^i_k) ~ \cdot \prod_{\hat{n} \in \mathbb{N}^i_{k} \cup i}\left( \mathbb{P}_{m^{\hat{n}}_j}(k) \right)^{a_{i \hat{n}}(k)}. %\times u(m^i_j, \mathbf{x}^i_k,\mathbf{z}^i_k )
 \end{align}
 
 We introduce the vector $\mathbf{u}_j[k] = [u(m^1_j, \mathbf{x}^1_k,\mathbf{z}^1_k)~\cdots~u(m^{N_R}_j, \mathbf{x}^{N_R}_k,\mathbf{z}^{N_R}_k)]^T$; that is, each entry of $\mathbf{u}_j[k]$ is the value of the function $u$ that a robot uses to compute the occupancy probability of the $j^{th}$ cell  in its map at time step $k$.
 %If we stack the outputs of $u(m^i_j, \mathbf{x}^i_k,\mathbf{z}^i_k ) $ corresponding to every robot at %epoch
 %time step $k$, we obtain a vector which we denote as $\mathbf{u}_j[k]$. 
{We use the notation $\mathbf{u}_j[k] \ne \mathbf{1}$ to indicate that at least one element of $\mathbf{u}_j[k]$ is not 1; i.e., at least one robot obtains distance measurements at time step $k$ that yield information about the $j^{th}$ cell.  We also define the set %$\mathbf{d}$ as a subsequence of the discrete infinite sequence $\{0, 1, ..., \infty\}$, such that 
 $\mathbf{d}$ as the sequence of time steps $d \in \{0, 1, ..., \infty\}$ for which $\mathbf{u}_j[d] \ne \mathbf{1}$. }
 
 %to obtain the geometric mean of its elements
 %$= \{d \in \{0, 1, ..., \infty\} ~\lvert\ \mathbf{u}_j[d] \ne 1 \}$.} 
 %By using the notation $\mathbf{u}_j[d] \ne 1$ we mean that at least one of the elements in vector $\mathbf{u}_j[d]$ is not unity.
 %$<\cdot>_{gm}$ is a geometric mean operator which takes a vector as an input and returns the geometric mean of its elements. If $\mathbf{q} = [q_1, q_2, ..., q_n]$, then $<\mathbf{q}>_{gm} = \sqrt[n]{q_1\cdot q_2 \cdot\cdot\cdot q_n}$. 
 
 We make the following  assumptions about the robots' communication network, map updates, and distance measurements. These assumptions are required to prove the main theoretical result of this paper, \autoref{thm: consensus}.
 
 \begin{ass}
	\label{ass:graph connectedness}
	There exists a time interval over which the union of robot communication graphs $\mathbb{G}(k)$
	%, where each time step $k$ is in this interval, 
	is connected.
\end{ass}

\begin{comment}
\hyperref[ass:graph connectedness]{Assumption~\ref*{ass:graph connectedness}} is required to apply the results from \cite[Theorem 1]{Kingston2006}. \spr{(Did those results require the union of the graphs to contain a spanning tree, or to be connected?)} 
\end{comment}

In reality, it is difficult to prove that this assumption holds true for robots that explore arbitrary domains.  However, it is reasonable to suppose that the assumption is satisfied for scenarios where a relatively high density of robots results in frequent robot interactions, or where the robots' communication range is large with respect to the domain area.  

%\spr{(Which of these occurs in our simulations and robot experiments?) (\autoref{sec:simulation} and \autoref{sec:experiments}).}
%\rag{I am not sure. But my best guess is frequent robot interactions happens during simulations and large communication range aids the experiments}
%due to the increase in the number of interactions as the number of robots in the swarm increases, one could suppose that \hyperref[ass:graph connectedness]{Assumption~\ref*{ass:graph connectedness}} is 
%most certainly 
%satisfied for large robot populations $N_R$. 

%\spr{The next two assumptions can be enforced } 

\begin{ass}
	\label{ass: pairwise interaction}
	%instant of time, 
 Suppose that at time step $k$, robot $R^i$ has a nonempty set of neighbors $\mathbb{N}^i_{k}$, indexed by $r = \{1,...,|\mathbb{N}^i_{k}|\}$. Then robot $R^i$ modifies its occupancy map according to \autoref{eqn:grid cell update rule} using the map from its $r^{th}$ neighbor $R^{\hat{n}}$ at each time step $k+r$, and setting the nonzero $a_{ij}$ values at time steps $k' = k+r-1$ to $a_{ii}(k') = 0.5$, $a_{i\hat{n}}(k') = 0.5$.
 If robot $R^i$ has no neighbors at time step $k$, i.e., $\mathbb{N}^i_{k} = \emptyset$, then it sets $a_{ii}(k) = 1$ in \autoref{eqn:grid cell update rule}.
 %pairwise with its neighbors and with $a_{i \hat{n}}(k) = 0.5$ or  $a_{i \hat{n}}(k) = 1$ (if $\mathbb{N}^i_{k} = \emptyset$). %\spr{\sout{In other words, $\mathbb{N}^i_{k}$ is a singleton or a null set at each time step $k$.}}
\end{ass}

\begin{comment}
This assumption
%pairwise interaction assumption (\hyperref[ass: pairwise interaction]{Assumption~\ref*{ass: pairwise interaction}}) 
is one way to ensure that adjacency matrix corresponding to $\mathbb{G}(k)$ is doubly stochastic, which is required to prove \autoref{thm: consensus}.
\end{comment}
%This assumption will hold if each robot modifies its occupancy map using only one of its neighbor's maps per time step.  
\begin{comment}
Suppose at a time step $k$ there are multiple robots around a particular robot within its information transmission range, then in order to make the \hyperref[ass: pairwise interaction]{Assumption~\ref*{ass: pairwise interaction}} hold, the robot stores the maps of its neighbors in buffer and modifies its map based on the maps in its buffer by using one map at a time step for every subsequent time step until all the maps in the buffer are used. 
\end{comment}
To illustrate, if robot $R^i$ has three neighbors $\mathbb{N}_k^i = \{R^p, R^q, R^r\}$ at time step $k$ that all communicate their maps $M^p, M^q, M^r$ to $R^i$, then $R^i$ can use \autoref{eqn:grid cell update rule} to update its map $M^i$ with map $M^p$ at time step $k+1$, map $M^q$ at time step $k+2$, and  map $M^r$ at time step $k+3$. If a fourth robot $R^s$ enters the neighborhood of $R^i$ between time steps $k$ and $k+3$, $R^i$ updates its map with map $M^s$ at time step $k+4$. The choice of $a_{ij}(k)$ values ensures that the adjacency matrix $\mathbb{A}(k)$ is doubly stochastic at each time step $k$. 
\begin{comment}
at time step 1 there are three neighbors for a robot then, it will store the maps of all its neighbors in a buffer. At time step 1, it will modify its map based the map of its first neighbor, after which it would use its next neighbor's map at time step 2 and finally uses the last neighbors map at time step 3 to modify its map information. 
\end{comment}

\begin{ass}
	\label{ass: finite support}
	The set $\mathbf{d}$ is finite.
\end{ass}

\begin{comment}
\hyperref[ass: finite support]{Assumption~\ref*{ass: finite support}} is used in the proof in \hyperref[app: proof of consensus theorem]{Appendix~\ref*{app: proof of consensus theorem}} to establish convergence of a sequence summation. 
\end{comment}
{This assumption is made to enable the proof of \autoref{thm: consensus}, rather than to describe an inherent property of the mapping strategy.}
In practice, the assumption can be realized by programming the robots to ignore their laser range sensor measurements after receiving a predefined number of them, although we did not need to enforce this assumption explicitly in our simulations and experiments.

\vspace{2mm}

To facilitate our analysis, we include an additional assumption on the robots' distance measurements. 
We define the {\it accessible grid cells} for robot ${R}^i$ as the set of grid cells in ${M}^i$ 
%(or in the entire domain? Otherwise, the index $a$ could depend on the particular robot's map $M^i$) 
whose occupancy probabilities can be inferred by $R^i$ from its laser range sensor measurements.
%while exploring the environment.  
This set includes all  unoccupied grid cells and the
grid cells along the periphery of obstacles.
%outer border grid cells of obstacles. 
The $a^{th}$ cell in this set %accessible grid cells 
is denoted by $m^i_a$. The set of \textit{inaccessible grid cells} contains all other cells in $M^i$,
%is the complement of the accessible grid cells 
and the $\bar{a}^{th}$ cell in this set is denoted by $m^i_{\bar{a}}$.

\vspace{2mm}

\noindent\textbf{Assumption 3A.} %For the ease of analysis, 
In the limiting case where the robots explore the environment for an infinite amount of time, 
%\rag{this for the sake of analysis, in the analysis we are considering asymptotic converges, which means the map information dynamics \autoref{eqn:grid cell update rule} is happening for an infinite time, we don't enforce it in practice} 
each robot $R^i$ uses the value of the function $u(m^i_a, \mathbf{x}^i_k,\mathbf{z}^i_k)$ at exactly one time step $k$ per accessible grid cell $m^i_a$ to infer the occupancy probability of that cell. We define $\bar{u}^i_a = u(m^i_a, \mathbf{x}^i_k,\mathbf{z}^i_k)$ for this $k$, which may be chosen as any time step at which $u(m^i_a, \mathbf{x}^i_k,\mathbf{z}^i_k) \neq 1$. We assume that there always exists such a time step $k$; i.e., that each robot obtains distance measurements that provide information about every accessible grid cell. 
%distance measurement 
%\spr{(Is this assumption about the number of {\it distance measurements}, or about the number of {\it values of the update function $u$} that the robot uses to compute the cell occupancy probabilities?}
%\rag{Although technical both are same, we are talking here number of values of the update function} 
%Let each entry in the vector $[\bar{u}^1_a ~ \cdots ~ \bar{u}^{N_R}_a]^T$ denote the value of the function $u$, not equal to one, that a robot uses to calculate the occupancy probability of the $a^{th}$ accessible grid cell. 
%(it can be a random choice, easier one to implement would be to choose the first value.)

\begin{comment}
The results presented in \autoref{sec:simulation} and \autoref{sec:experiments} throw light on the fact that, in practice it is not required to impose \hyperref[ass: finite support]{Assumption~\ref*{ass: finite support}} explicitly, as we did not enforce any such condition while conducting simulations and experiments for this paper.
\end{comment}
 
 \vspace{2mm}
 
 We can now state the main result of this paper, which uses the following definitions. The vector of all robots' occupancy probabilities for the $j^{th}$ cell at time step $k$ is written as $\mathbb{P}_{m_j}[k] = [\mathbb{P}_{m^1_j}(k) ~...~ \mathbb{P}_{m^{N_R}_j}(k)]^T$. %$\mathbb{P}_{m^i_j}(k)\  \forall i \in \mathbb{V}$. 
 In addition, $<\cdot>_{gm}$ denotes the geometric mean operator, defined as $<\mathbf{q}>_{gm} = \sqrt[n]{q_1 \cdot q_2 \cdot ... \cdot q_n}$ for a vector $\mathbf{q} = [q_1 ~ q_2 ~ ... ~ q_n]^T$.
 %the following theorem, which is an important result of this paper.

\newtheorem{theorem}{Theorem}
\begin{theorem}
	\label{thm: consensus}
	If each robot $R^i$ updates its occupancy grid map according to \autoref{eqn:grid cell update rule}, then under \hyperref[ass:graph connectedness]{Assumption~\ref*{ass:graph connectedness}}-\hyperref[ass: finite support]{Assumption~\ref*{ass: finite support}} (excluding \hyperref[ass: finite support]{Assumption~\ref*{ass: finite support}A}), %stated below, 
	we have that
	\begin{align}
	\label{eqn: asymptotic result}
	\lim_{k \rightarrow \infty} \mathbb{P}_{m^i_j}(k) ~=~  <\mathbb{P}_{m_j}[0]>_{gm} \hspace{1mm} \cdot \hspace{1mm} \prod_{d \in \mathbf{d}} <\mathbf{u}_j[d]>_{gm}.
	\end{align}
	
	For an inaccessible grid cell $m^i_{\bar{a}}$, \autoref{eqn: asymptotic result} reduces to 
	\begin{align}
	    \label{eqn: inaccessible asymptotic}
	    \lim_{k \rightarrow \infty} \mathbb{P}_{m^i_{\bar{a}}}(k) ~=~  <\mathbb{P}_{m_{\bar{a}}}[0]>_{gm}.
	\end{align}
\end{theorem}
In addition, under \hyperref[ass: finite support]{Assumption~\ref*{ass: finite support}A}, the asymptotic value of $\mathbb{P}_{m^i_{a}}$ for an accessible grid cell $m^i_a$ can be derived from \autoref{eqn: asymptotic result} as 
\begin{equation}
\label{eqn: accessible asymptotic}
\lim_{k \rightarrow \infty} \mathbb{P}_{m^i_a}(k)  ~=~   <\mathbb{P}_{m_a}[0]>_{gm}  \cdot  <[\bar{u}^1_a ~ \cdots ~ \bar{u}^{N_R}_a]^T>_{gm},
\end{equation}

Furthermore, \autoref{eqn: asymptotic result}, \autoref{eqn: inaccessible asymptotic}, and \autoref{eqn: accessible asymptotic} converge exponentially to their respective limits.
\begin{proof}
	See \hyperref[app: proof of consensus theorem]{Appendix~\ref*{app: proof of consensus theorem}}.
\end{proof}

\autoref{thm: consensus} states that under the map modification protocol \autoref{eqn:grid cell update rule}, the occupancy probability $\mathbb{P}_{m^i_j}$ of every grid cell $m^i_j$ in the map of each robot $R^i$
\begin{comment}
if each robot $R^i$ updates its \spr{occupancy map}
%$\mathbb{P}_{m^i_j}$ 
according to the 
protocol \autoref{eqn:grid cell update rule}, then
$\mathbb{P}_{m^i_j} \forall i \in \{1, ..., N^R\}, j \in \{1, ..., M^i\}$ 
\end{comment}
will converge exponentially to  a value that is proportional to $\prod_{d \in \mathbf{d}} <\mathbf{u}_j[d]>_{gm}$.  For each inaccessible grid cell, \autoref{eqn: inaccessible asymptotic} dictates that the occupancy probability  converges to the  constant $<\mathbb{P}_{m_{\bar{a}}}[0]>_{gm}$, If we set $\mathbb{P}_{m^i_j}(0) = 1$ for each robot $R^i$, then this constant equals one, which is an accurate occupancy probability since the cell is occupied.
%(Actually, it's not the same proportionality constant as in \autoref{eqn: asymptotic result}, since it only includes the initial occupancy probabilities from the inaccessible cells. And as stated later in the paragraph, it will be initialized to 1.) 
By \autoref{eqn: accessible asymptotic}, the occupancy probability of each accessible grid cell $m^i_a$ will asymptotically tend to the geometric mean of $[\bar{u}^1_a ~ \cdots ~ \bar{u}^{N_R}_a]^T$ if the proportionality constant $<\mathbb{P}_{m_a}[0]>_{gm}$ is 1, which also occurs if  
%The proportionality constant, $<\mathbb{P}_{m_j}[0]>_{gm}$, will be 1 if 
we initialize $\mathbb{P}_{m^i_j}(0) = 1$ for each robot $R^i$.
%as the initial occupancy probability of $m^i_a$ in each robot's map.  
%condition for the update rule \autoref{eqn:grid cell update rule}. 
%In this manner, the asymptotic behavior of update rule \autoref{eqn:grid cell update rule} is only dependent on the laser range sensor measurements made by the robots. 
%The asymptotic value of $\mathbb{P}_{m^i_j}$ is proportional to the geometric mean of the laser range sensor measurements made by all robots at various time steps. 
%the probability of occupancy of a grid of ever robot would eventually converge to a value computed based on all the information for all the robots over all the time step. Therefore, the 
%\rag{Probability of occupancy converges to a value that  best represents the occupancy of the grid cell, meaning for an occupied cell $m^i_j$, $\mathbb{P}(m^i_j = 1)$ will be close to one and vice versa.} 
%\spr{Does ``the value that best represents the occupancy'' mean ``the occupancy probability that is closest to the true occupancy (0 or 1)''?  Why does the geometric mean produce this value?}
Since the occupancy probability $\mathbb{P}_{m^i_j}$ of each grid cell $m^i_j$ ultimately converges to the geometric mean of occupancy probabilities computed by every robot, and the effect of outliers in the data is greatly dampened in the geometric mean  \cite{Rani2014},  the resulting $\mathbb{P}_{m^i_j}$  reasonably represents the true occupancy of the grid cell, even if a few robots record 
highly noisy or inaccurate measurements.

{We note that although our analysis specifically guarantees {\it asymptotic} convergence of the robots' maps to the actual map, 
%that the maps of the robots converge asymptotically, 
 our simulation and experimental results  in \autoref{sec:simulation} and \autoref{sec:experiments} show that the maps indeed converge in finite time within reasonable accuracy. }

% \autoref{thm: consensus} delineates that if each robot $R^i$ updates its $\mathbb{P}_{m^i_j}$ according to the discrete time consensus based map modification protocol \autoref{eqn:grid cell update rule}, then  $\mathbb{P}_{m^i_j} \forall i \in \{1, ..., N^R\}, j \in \{1, ..., M^i\}$ would eventually reach a value which is proportional to $\prod_{d \in \mathbf{d}} <\mathbf{u}_j[d]>_{gm}$. The proportionality constant is the  geometric mean of  elements of the vector $\mathbb{P}_{m_j}[0]$. The proportionality constant can be made unity if we choose $\mathbb{P}_{m^i_j} = 1$ as initial condition for the update rule \autoref{eqn:grid cell update rule}. In this manner, the asymptotic behavior of update rule \autoref{eqn:grid cell update rule} is only dependent on the laser range sensor measurements made by the robots. The asymptotic value of $\mathbb{P}_{m^i_j} $ is proportional to the geometric mean of the laser range sensor measurements made by all robots at various time steps. As a result, the asymptotic value of $\mathbb{P}_{m^i_j} $ will converge to a probability value indicating the true occupancy of the grid cell even if a few robots report highly noisy or incorrect measurements.
 
 \section{Post-Processing of Occupancy Grid Maps}
\label{sec:post processing OCG}

%\spr{Is this part done off-line, i.e., not by each robot? That should be mentioned.}

\begin{comment}
In this section, we propose a technique for post-processing the occupancy grid map generated by the robots \spr{(after it has converged to a common map?)}. %Post processing of occupancy grid map refers to 
\spr{By post-processing, we mean inferring}
\end{comment}

Since all robots' occupancy grid maps eventually converge to a common occupancy map, in theory only a single robot needs to be retrieved to obtain this map. 
In this section, we propose a technique for post-processing the grid cell occupancy probabilities %that are computed by 
from the retrieved robot(s) to infer the most likely occupancy grid map of the environment. We note that this technique can be applied to occupancy grid maps that are generated through any  mapping procedure.
%This technique can be applied to the occupancy probabilities retrieved from the robots after they have explored the environment for (how long? long enough for their maps to converge to a common map?).
A common approach to this 
%the occupancy grid map 
inference problem is the \textit{Maximum A Posterior (MAP)}  mapping procedure \cite{Thrun:2005:PR:1121596}, which computes the occupancy grid map with the maximum probability of occurrence based on the occupancy probability of each grid cell in the map. In general, the MAP procedure is posed as an optimization problem, and the solution is computed using gradient-based hill climbing  methods. This approach is computationally expensive, since gradient ascent must be performed  from different initial conditions to escape local {maxima} and the search space is exponential in the number of grid cells (for a given set of $n$ grid cells, there  are
$2^n$ possible occupancy grid maps \cite{Thrun:2005:PR:1121596}).

Here, we present an alternative approach that is based on techniques from \textit{topological data analysis} (TDA) \cite{edelsbrunner2010}, which uses the mathematical framework of algebraic topology \cite{Hatcher2002}.
%to characterize the topological structure of data.
%an applied version of algebraic topology. 
%concepts from algebraic topology \cite{Hatcher2002}.
%studying topological and geometric attributes of noisy data in a coordinate-free manner.
In practice, the time complexity of our procedure
%and for most practical purpose
is linear in the number of grid cells $(\mathcal{O}(|M^i|))$ \cite{Ramachandran_Wilson_Berman_RAL_17}. In \autoref{subsec: AT and TDA}, we present an overview of relevant concepts from TDA and algebraic topology. %required to understand our procedure. 
An in-depth treatment of these subjects can be found in  \cite{edelsbrunner2010,Hatcher2002,Kaczynski04a,Ghrist2008barcodes}.  We then describe our TDA-based occupancy grid mapping procedure in \autoref{subsec: adaptive thresholding}.

\subsection{Algebraic topology and Topological Data Analysis (TDA)}
\label{subsec: AT and TDA}

In recent years, considerable progress has been made in using tools from algebraic topology to estimate the underlying structure and shape of data \cite{carlsson2009topology}, %Understanding the underlying shape of data a priori would 
which aids in efficient analysis of the data using statistical techniques such as regression \cite{edelsbrunner2008persistent}. Topological data analysis (TDA) is a collection of algorithms for performing coordinate-free topological and geometric analysis of noisy data.  In most applications, the data consists of %obtained by 
noisy samples of an intensity map that is supported on a Euclidean domain. The set of these %noisy sampled 
data is referred as a \textit{point cloud}. The dominant topological features of the Euclidean domain associated with the point cloud can be computed using {\it persistent homology} \cite{edelsbrunner2008persistent}, a key concept in TDA. 
%Persistent homology can be used to filter topological features that persist over a wide range of scales.
A compact graphical representation of this information can be presented using  barcode diagrams \cite{Ghrist2008barcodes} and persistence diagrams \cite{edelsbrunner2008persistent}.

%In contrast to our previous works \cite{Ramachandran_Wilson_Berman16,Ramachandran_Wilson_Berman_RAL_17}, here we focus on cubical homology rather than simplicial homology.  

%Let $\mathbf{T}$ be a topological space that admits a  cubical decomposition.
A topological space $\mathbf{T}$ can be associated with an infinite sequence of vector spaces called \textit{homology groups}, denoted by  $H_t(\mathbf{T})$, $t = 0, 1, 2, \dots$. 
%\spr{\sout{Every topological feature information regarding $\mathbf{T}$ is encoded in every one of these vector spaces.}}
Each of these vector spaces encodes information about a particular topological feature of $\mathbf{T}$.
The dimension of $H_t(\mathbf{T})$, defined as the \textit{Betti number} $\beta_t$ \cite{Ghrist2008barcodes}, is a topological invariant that represents %quantifies 
the number of independent topological
features encoded by $H_t(\mathbf{T})$. %These topologically invariant quantities, denoted by $\beta_t$, \spr{are called} \textit{Betti numbers} \cite{Ghrist2008barcodes}. 
 Additionally, $\beta_t$ gives the number of
independent  $t$-dimensional cycles in $\mathbf{T}$. %the topological space. 
For example, if $\mathbf{T}$ is embedded in $\mathbb{R}^2$, denoted by $\mathbf{T} \hookrightarrow \mathbb{R}^2$, then $\beta_0$ and $\beta_1$ are the number of connected components in $\mathbf{T}$ and number of holes in $\mathbf{T}$, respectively.

In contrast to our previous works \cite{Ramachandran_Wilson_Berman16,Ramachandran_Wilson_Berman_RAL_17}, here we consider topological spaces that admit a cubical decomposition rather than a simplicial decomposition and use cubical homology rather than simplicial homology.
The fundamental unit of a cubical complex is an \textit{elementary interval} \cite{Kaczynski04a}, a closed interval $\mathtt{I} \subset \mathbb{R}$ of the form $\mathtt{I}=[l, ~l+1]$ (a {\it nondegenerate} interval) or $\mathtt{I}=[l, ~l]$ (a {\it degenerate} interval) for some $l\in\mathbb{Z}$. 
%An elementary  interval is called {\it degenerate} if it contains only one point and {\it nondegenerate} if it has unit length.  
A \textit{cube} or \textit{elementary cube} $\mathbb{Q} \subset \mathbb{R}^d$ is constructed from a finite product of elementary intervals $\mathtt{I}_t$, $\mathbb{Q} = \prod_{t=1}^{d} \mathtt{I}_t$ %where each $\mathtt{I}_t$ is an elementary interval
\cite{Kaczynski04a}.  If $\mathbb{Q}$ and $\mathbb{O}$ are elementary cubes and $\mathbb{Q} \subset \mathbb{O}$, then $\mathbb{Q}$ is a {\it face} of $\mathbb{O}$. For a topological space $\mathbf{T}$, let a $t$-cube $\square_t$ be a continuous map $\square_t : [0,1]^t \rightarrow \mathbf{T}$ \cite{LaValle2006}. 
%As mentioned earlier one can create 
A $t$-cube has %consists of 
$2t$ faces, each of which is a ($t-1$)-dimensional cube.  A cubical complex $\mathbb{K}$ is a union of $t$-cubes 
%whose faces are all in $\mathbb{K}$, 
%for which every face $\square_{t-1}$ of each cube $\square_t\in \mathbb{K}$ is in $\mathbb{K}$, 
for which the faces of each cube are all in $\mathbb{K}$ and the intersection of any two cubes $\square_t$ and $\square'_t$ is either the empty set or a common face of both $\square_t$ and $\square'_t$.
%\rag{A cubical complex can be created by taking the union of $t$-cubes of different dimensions $t$.}
%it must meet the following requirements: 1) 

\begin{figure}[!t]
	\vspace{2mm}
	\hspace{3mm}
	\includegraphics[width=0.9\linewidth, height=.6\linewidth]{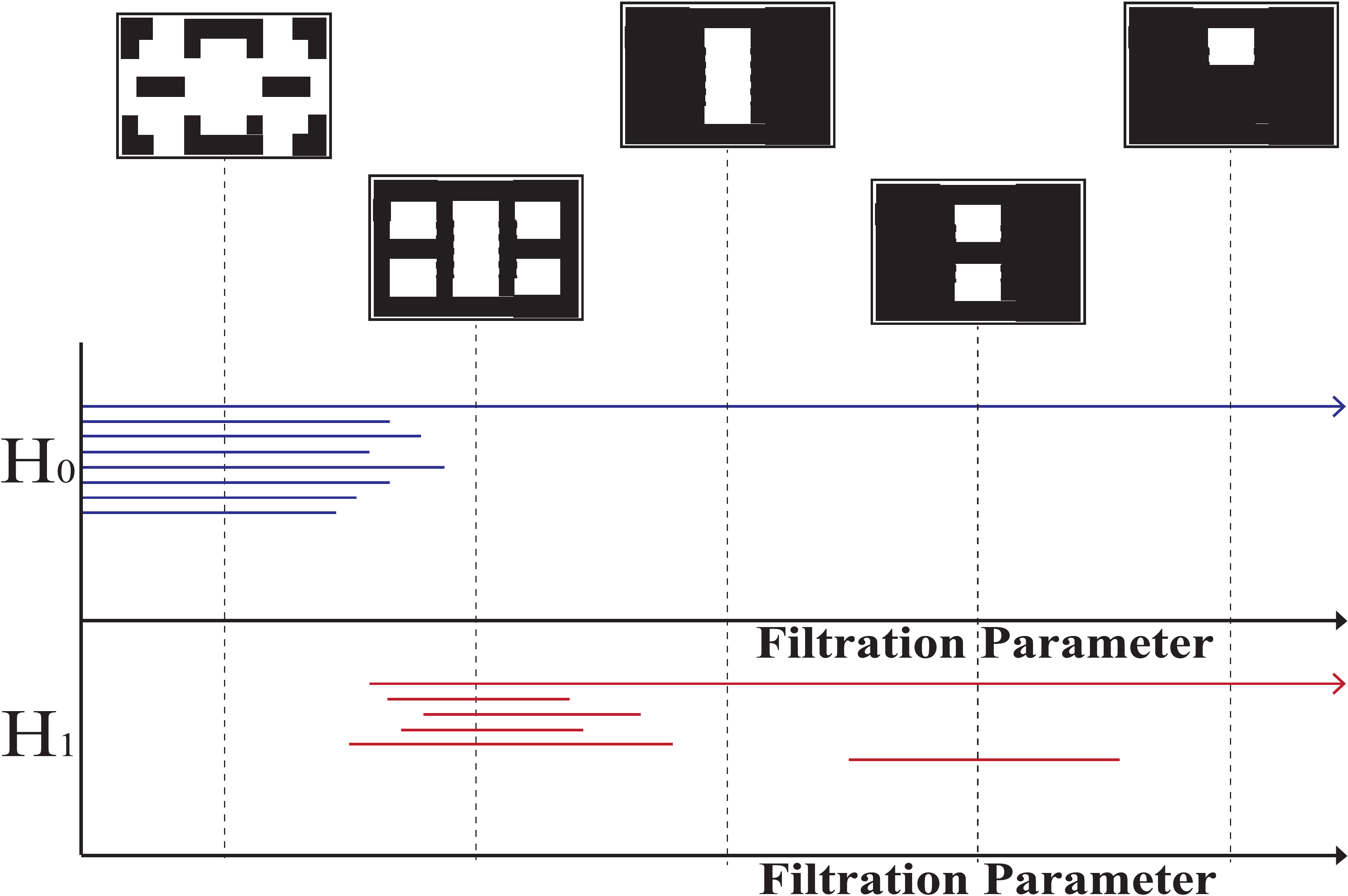}
	\caption{An example barcode diagram of a filtration constructed from a cubical complex. The shaded regions contain the two-dimensional \textit{elementary cubes} (squares). The arrows in $H_0$ and $H_1$ indicate the persistent topological features over a range of values %variation 
	of the filtration parameter. 
	The arrows show that the cubical complex has one persistent topological feature corresponding to each homology group $H_0$ and $H_1$.}
	\label{fig:barcode_example}       % Give a unique label
\end{figure}

Suppose that $\iota, \eta \in \mathbb{K}$. We use the notation $\eta \leq \iota$ to indicate that $\eta$ is a face of $\iota$. Let $f: \mathbb{K} \to \mathbb{R} $ be a function for which $\eta \leq \iota$  implies that $f(\eta) \leq f(\iota)$. Then $f^{-1}((-\infty, \varpi])$ is a cubical complex denoted by $\mathbb{K}_\varpi$, and $\varpi_1 \leq \varpi_2$ implies that  $\mathbb{K}_{\varpi_1} \subseteq \mathbb{K}_{\varpi_2}$, yielding a \textit{filtration} \cite{Kaczynski04a} of cubical complexes with $\varpi$ as its \textit{filtration parameter}. The persistent homology can be generated by varying the value of $\varpi$ %the filtration parameter 
and computing the basis of the homology group vector spaces (the homology generators) for each cubical complex corresponding to the 
%filtration parameter 
value of $\varpi$. A \textit{barcode diagram} represents $H_t(\mathbf{T})$ in terms of its homology generators  and can be used to determine persistent topological features of the topological space $\mathbf{T}$. 
%A barcode diagram helps to recognize the persistent topological features of a topological space. 
\autoref{fig:barcode_example} gives an example of a barcode diagram for a cubical complex. The diagram plots a set of horizontal line segments whose $x$-axis spans a range of filtration parameter values 
and whose $y$-axis shows the homology generators %depicts the homology group vector space basis 
in an arbitrary ordering. The number of arrows in the diagram indicates the count of persistent topological features of $\mathbf{T}$. Specifically, the number of arrows in each homology group corresponds to the number of topological features %in the domain 
that are encoded by that %particular 
group.

\subsection{Classifying occupied and unoccupied grid cells %from free ones 
with adaptive thresholding}
\label{subsec: adaptive thresholding}

We now describe our technique for distinguishing occupied grid cells from unoccupied grid cells by applying the concept of persistent homology %\cite{edelsbrunner2008persistent}
to automatically find a threshold based on the occupancy probabilities $\mathbb{P}_{m^i_j}$ in a map $M^i$. This TDA-based technique provides an adaptive method for thresholding an occupancy grid map of a domain that contains obstacles at various length scales.   In this approach, we threshold  $\mathbb{P}_{m^i_j}$ at various levels, compute the numbers of topological  holes (obstacles) in the domain corresponding to each level of thresholding, and identify the threshold value above which topological features persist. 

As described in \autoref{subsec: AT and TDA},
%\cite{Kaczynski04a}, 
a filtration of cubical complexes
$\mathbb{K}_{\varpi}$ 
%based on a parameter called the 
with filtration parameter $\varpi$ can be used to compute the persistent homology. In order to be consistent with the definition of a filtration, we set $\mathbb{P}_{m^i_j} = 1$ for unexplored grid cells $m^i_j$. We define the filtration parameter $\varpi$ as a threshold for identifying unoccupied %obstacle-free 
grid cells $m^i_j$ according to $\mathbb{P}_{m^i_j} < \varpi$.  
A filtration is constructed by creating  cubical complexes $\mathbb{K}_{\varpi}$ 
for a sequence of increasing $\varpi$ values. Each complex $\mathbb{K}_{\varpi}$ is defined as the union of all $2$-cubes $\square_2$ whose vertices $\square_0$ are the centers of grid cells $m^i_j$ for which $\mathbb{P}_{m^i_j} < \varpi$.

%The method starts by adding those $2$-cubes (square $\square_2$) to a cubical complex whose all four vertices ($\square_0$) belong to the center of grid cells with probability of occupancy less than $\varpi$. 
%This process is repeated for an increasing sequence of $\varpi$ values, which results in a filtration. %It is to be noted that 
%Once a filtration is constructed, 
Next, a barcode diagram is extracted from the filtration and used to identify the number of topological features in the domain, which is given by the number of arrows in each homology group. The threshold $\varpi_{cls}$ for classification of the grid cells as occupied or unoccupied is defined as the minimum
value of $\varpi$ for which all the topological features are captured by the corresponding cubical complex $\mathbb{K}_\varpi$. In the barcode diagram, %$\varpi_{cls}$ is the value of $\varpi$ for which
there exists no horizontal line segment other than the arrows in any of the homology groups for all $\varpi > \varpi_{cls}$. 
%$\varpi_{cls}$ is the value of $\varpi$ for which all barcode segments except the arrows are annihilated for all %values of 
%$\varpi > \varpi_{cls}$. % greater than this value. 
%This computation is done in practice 
The value of $\varpi_{cls}$ can be computed as the maximum value of $\varpi$ that is spanned by the terminating barcode segments in all the homology groups. 

The persistent homology computations were performed using the C++ program   Perseus \cite{Perseus}, and the barcode diagrams were generated using MATLAB. Since Perseus only accepts integers as filtration parameters, the  $\mathbb{P}_{m^i_j}$ values of each map $M^i$ were scaled between 0 and 255 prior to being used as input to the computations.
%before inputing the data into software. 
%The results presented in this paper 
In our simulations and experiments, we restricted the persistent homology computations to dimensions zero and one since the domains being mapped were two-dimensional.

	%%%%%%%%%%%%Worlds time%%%%%%%%%%%%%%%%%%%%%%%%%%%%%%%%%%%%%%%%

\begin{figure*}[!t]
	
	\begin{tabular}{ccccc}
		\centering	
		
		\subcaptionbox{ \label{subfig:plain}}{\includegraphics[width=.18\linewidth, height=.16\linewidth]{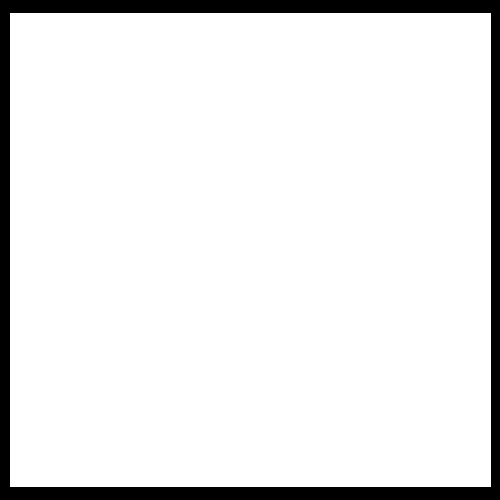}}
		
		&
		
		\subcaptionbox{ \label{subfig:mycave}}{\includegraphics[width=.18\linewidth, height=.16\linewidth]{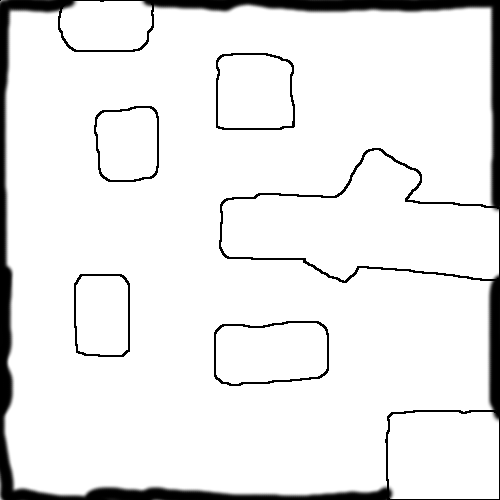}}
			
		&		
		
		\subcaptionbox{ \label{subfig:uoa_robotics_lab}}{\includegraphics[width=.18\linewidth, height=.16\linewidth]{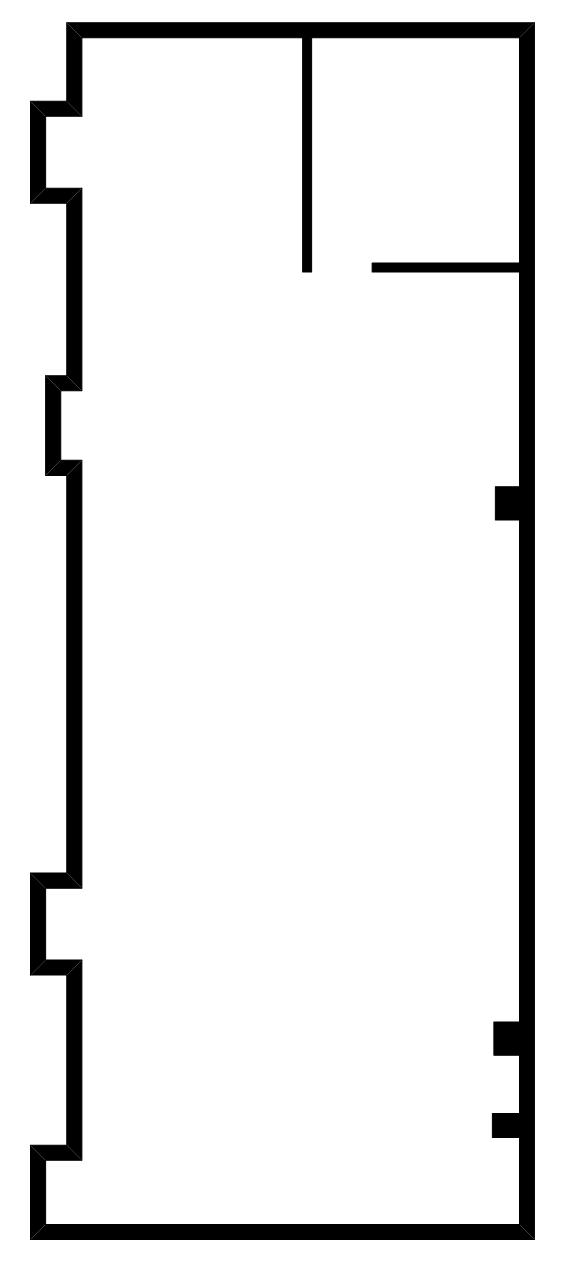}}
		
		& 
		
		\subcaptionbox{ \label{subfig:autolab}}{\includegraphics[width=.18\linewidth, height=.16\linewidth]{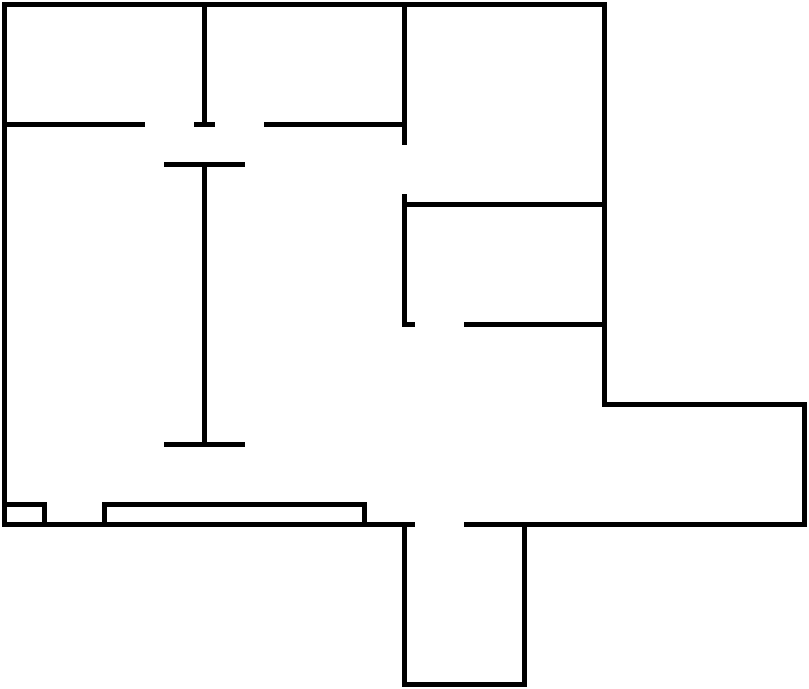}}
		
		& 
		
		\subcaptionbox{\label{subfig:frieburg}}{\includegraphics[width=.18\linewidth, height=.16\linewidth]{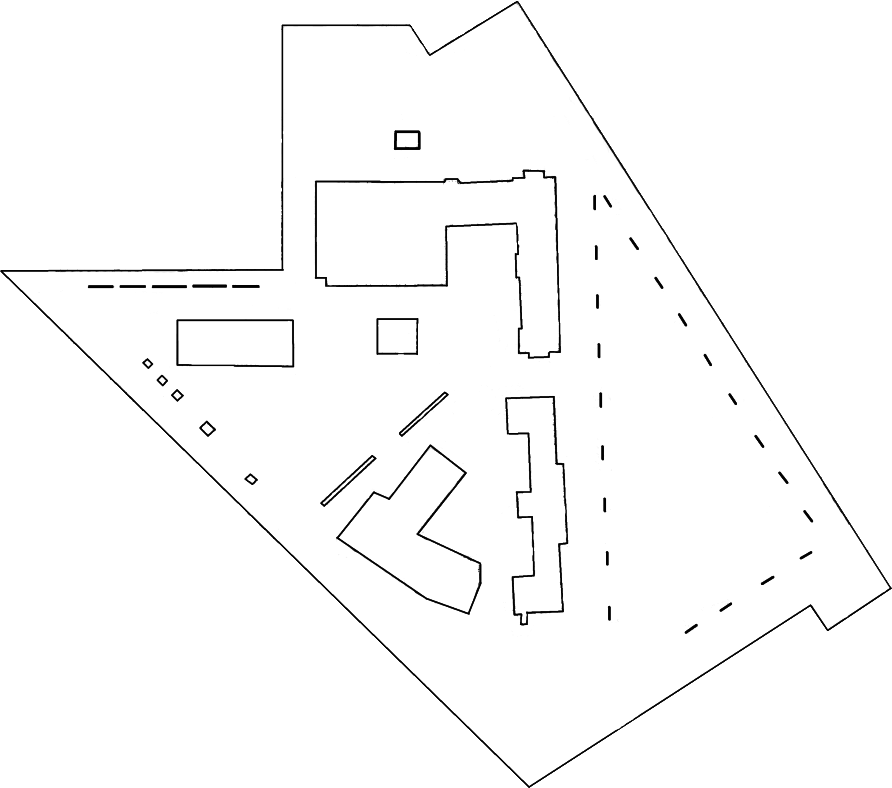}}
		
	\end{tabular}
	\caption{Simulated environments, all  obtained from \cite{Stage}. The dimensions of the minimum bounding rectangle of each environment are given in parentheses. (a) An unobstructed environment (16 m $\times$ 16 m); (b) a cave environment (16 m $\times$ 16 m); (c) floor plan of a robotics laboratory at the University of Auckland (40 m $\times$ 20 m); (d) floor plan of an autonomy laboratory (40 m $\times$ 30 m); (e) University of Freiburg campus (90 m $\times$ 80 m).}  
	\label{fig:worlds}
\end{figure*}

%%%%%%%%%%%%%%%%%%%%%%%%%%%%%%%%%%%%%%%%%%%%%%%%%%%%%%%%%%%%%%%%%%%%%%%%%%%

%%%%%%%%%%%%generated maps time%%%%%%%%%%%%%%%%%%%%%%%%%%%%%%%%%%%%%%%%

\begin{figure*}[!t]
	
	\begin{tabular}{ccccc}
		\centering

		\subcaptionbox{ \label{subfig:plain map}}{\includegraphics[width=.18\linewidth, height=.16\linewidth]{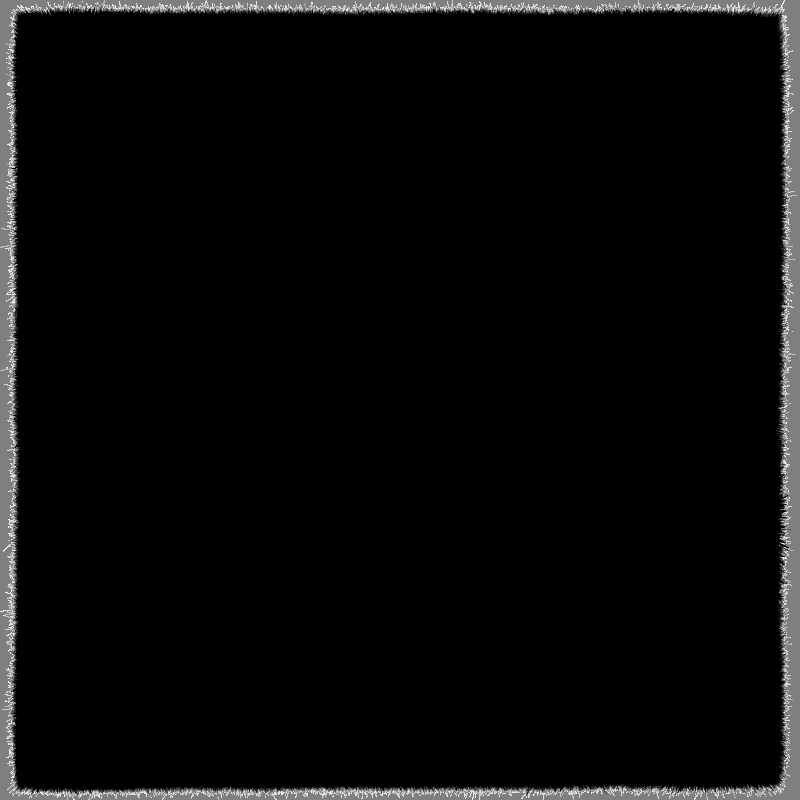}} 
		
		&	
		
		\subcaptionbox{ \label{subfig:mycave map}}{\includegraphics[width=.18\linewidth, height=.16\linewidth]{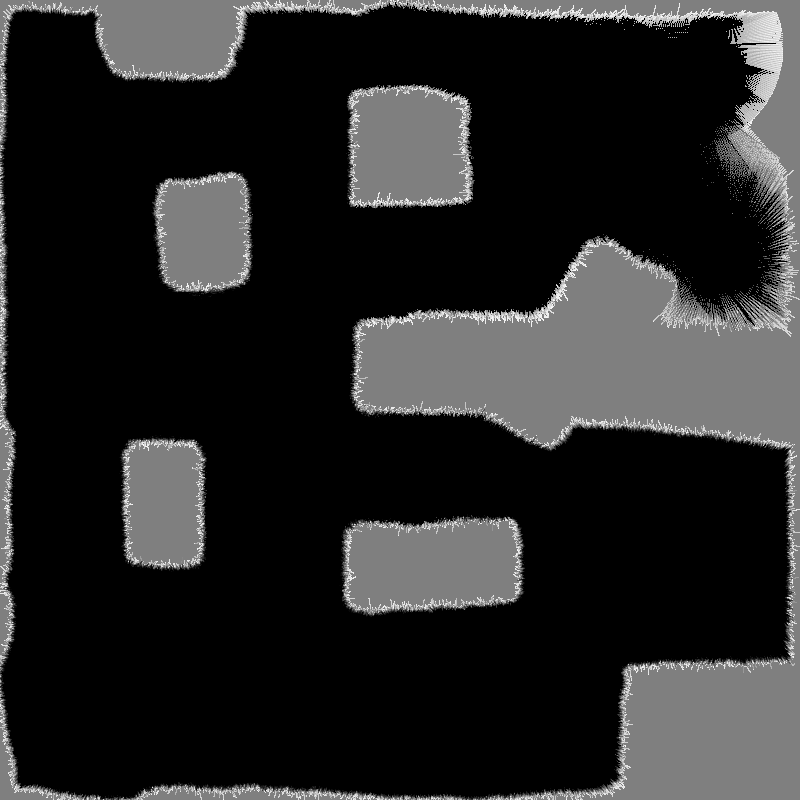}}
		
		&
		
		\subcaptionbox{ \label{subfig:uoa_robotics_lab map}}{\includegraphics[width=.18\linewidth, height=.16\linewidth]{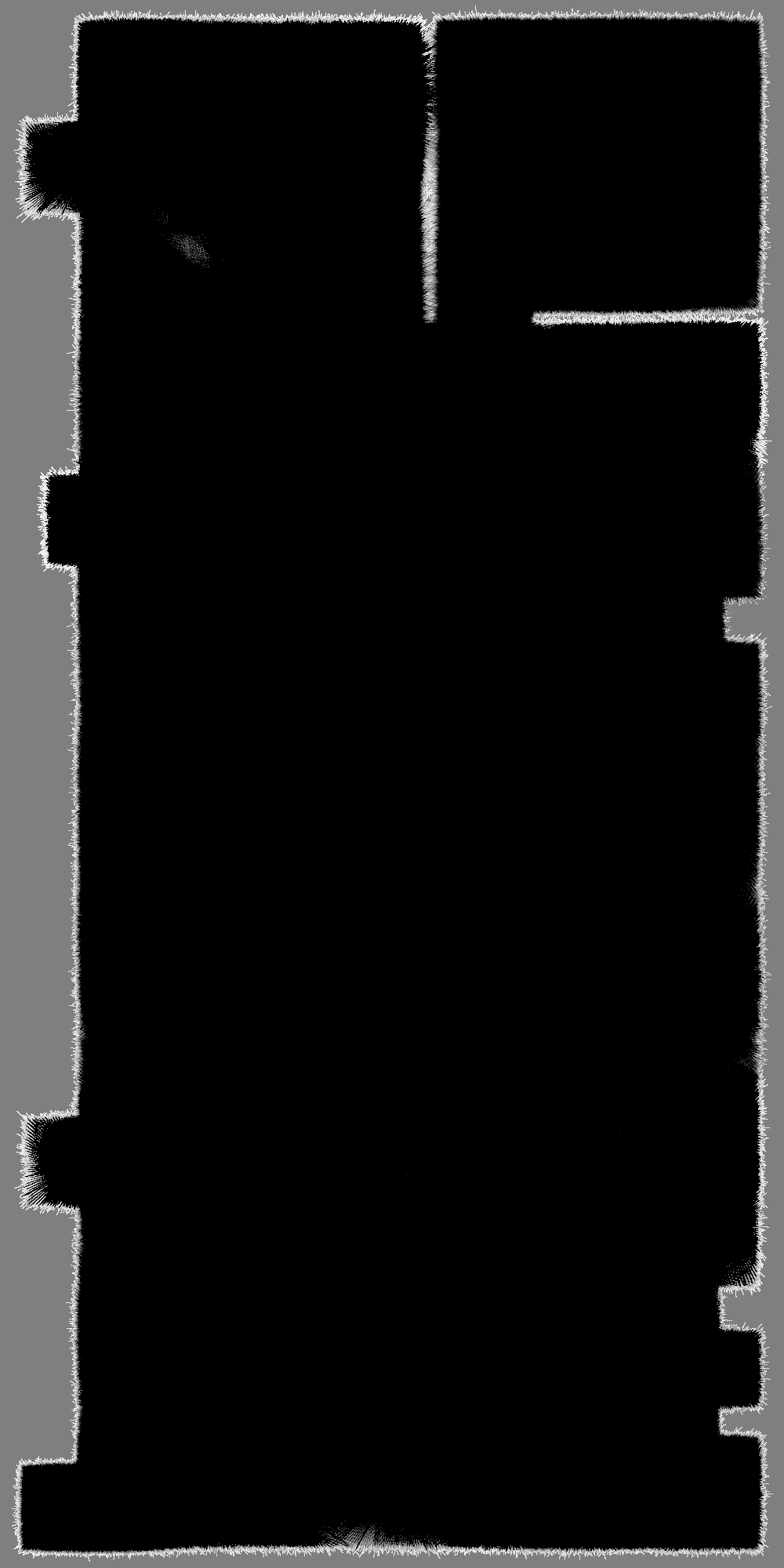}}
		
		&
		
		\subcaptionbox{ \label{subfig:autolab map}}{\includegraphics[width=.18\linewidth, height=.16\linewidth]{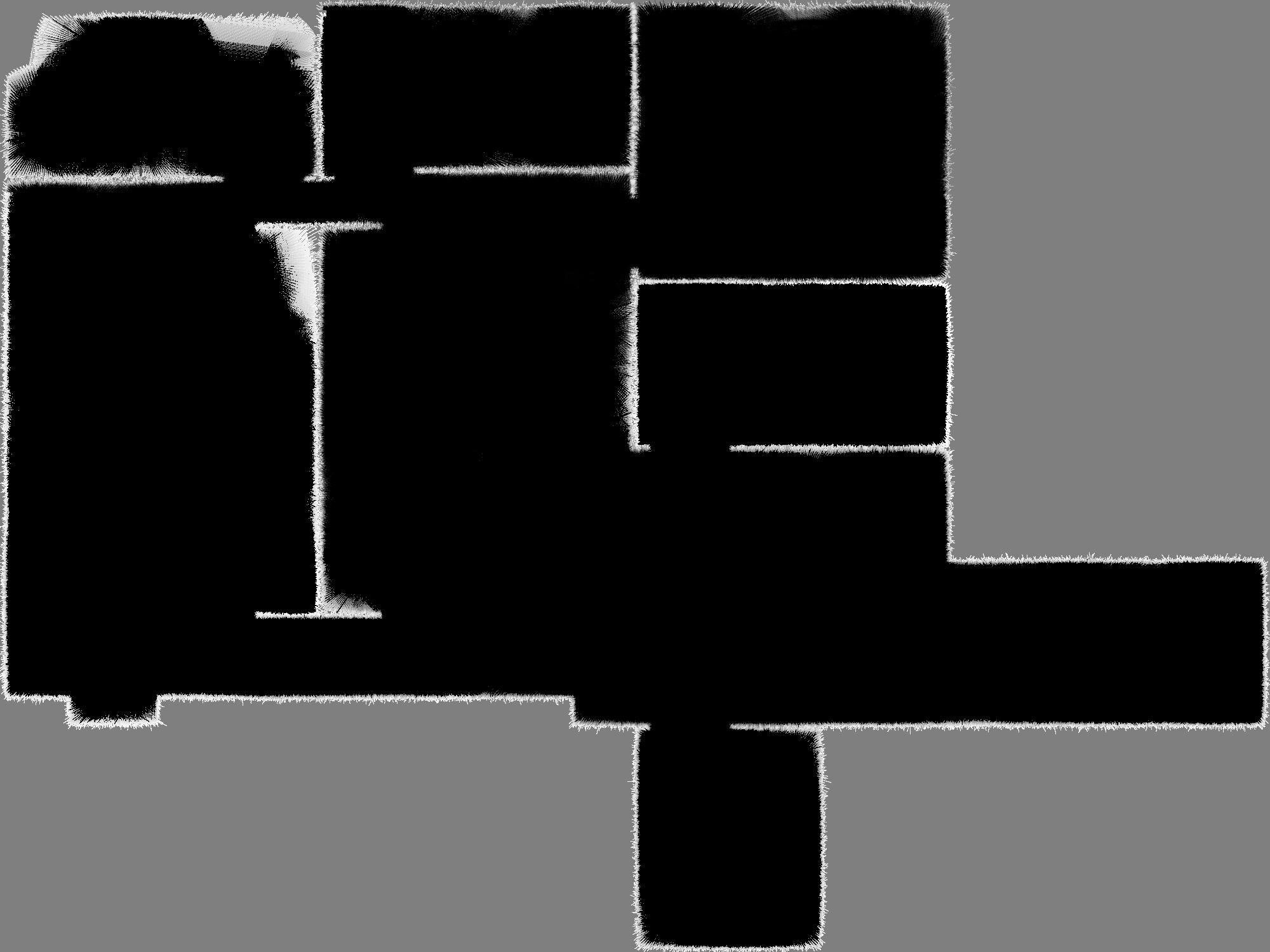}}
		&

		\subcaptionbox{ \label{subfig:frieburg map}}{\includegraphics[width=.18\linewidth, height=.16\linewidth]{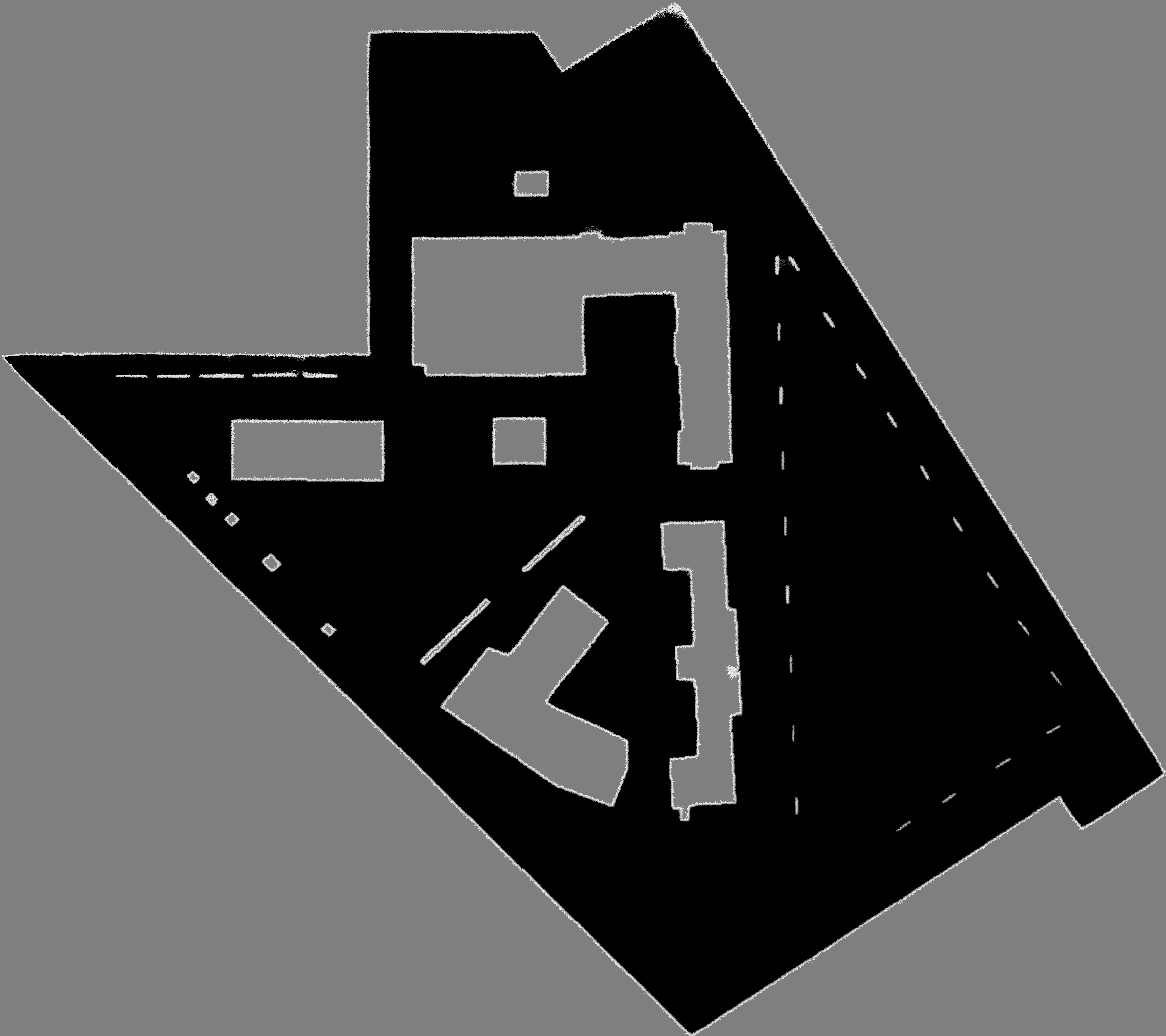}}

	\end{tabular}
	\caption{Occupancy grid maps of the environments in (a) \autoref{subfig:plain}, (b) \autoref{subfig:mycave}, (c) \autoref{subfig:uoa_robotics_lab}, (d) \autoref{subfig:autolab}, and (e) \autoref{subfig:frieburg}. Each map is generated by an arbitrary member of a swarm of $N_R$ robots that explore the environment for $t_f$ seconds: (a) $N_R = 1$, $t_f = 1800$ s; (b) $N_R = 5$, $t_f = 1200$ s; (c) $N_R = 10$, $t_f = 1500$ s; (d) $N_R = 20$, $t_f = 900$ s; and (e) $N_R = 50$, $t_f = 3600$ s.
	%through information based exploration and our distributed mapping strategy.  
	Gray regions that correspond to obstacles and the exterior of the domain are unexplored. Within the explored parts of the domain, dark regions contain grid cells with low occupancy probabilities  ($\mathbb{P}_{m^i_j} \approx 0.1$ in black regions), and light regions have high occupancy probabilities ($\mathbb{P}_{m^i_j} \approx 0.9$ in white regions).} 
	%map generated when a single robot explored a domain with the layout depicted in \autoref{subfig:plain} for $1800s$; 
	\label{fig:generated maps}
\end{figure*}

\begin{figure}[!t]
	
	\begin{tabular}{ccc}
		\centering

		\subcaptionbox{$0$ s  \label{subfig:scr_t_0}}{\includegraphics[width=.3\linewidth]{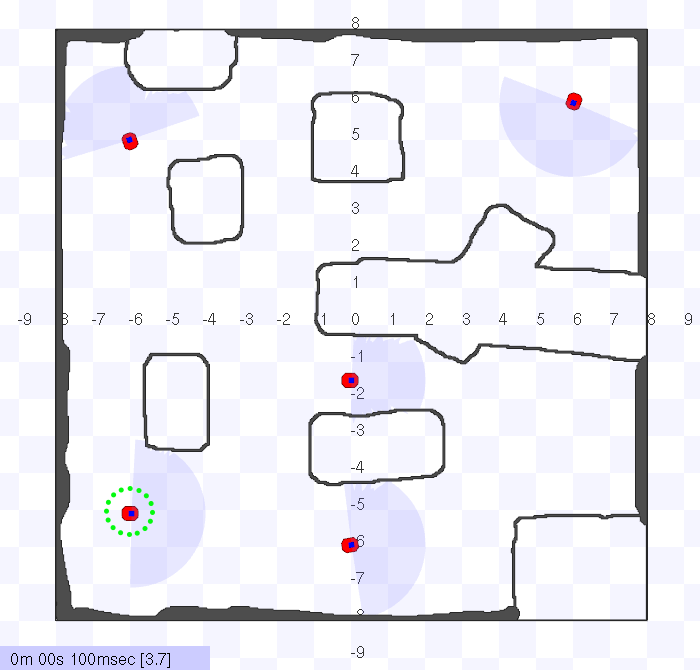}}
		&

		\subcaptionbox{$133$ s \label{subfig:scr_t_133}}{\includegraphics[width=.3\linewidth]{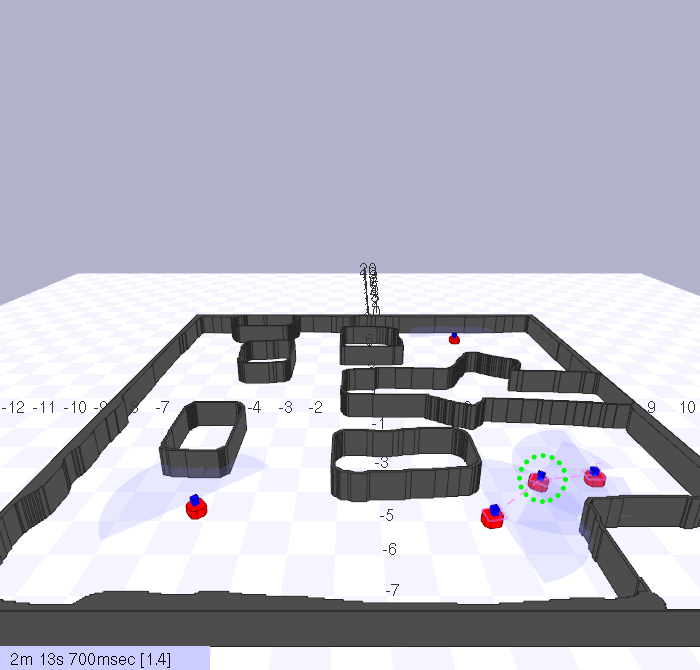}}
		
		&

		\subcaptionbox{ $897$ s \label{subfig:scr_t_897}}{\includegraphics[width=.3\linewidth]{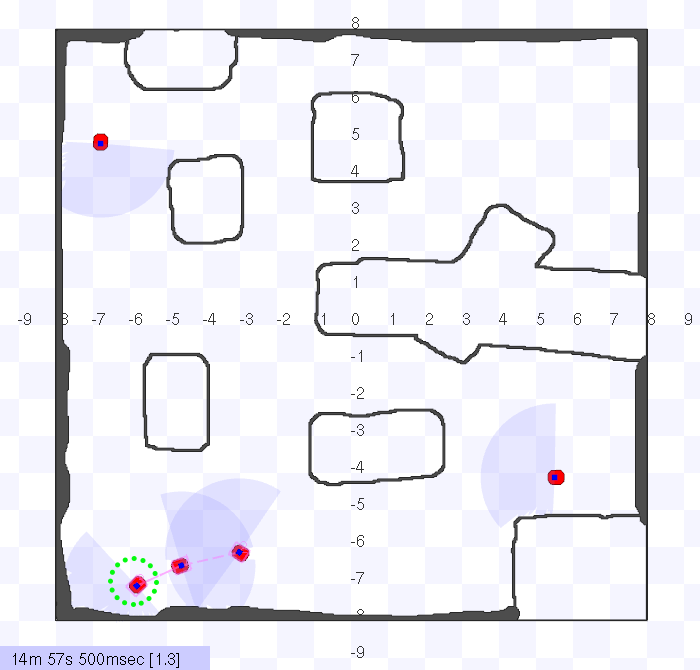}} 
		
	\end{tabular}
	\caption{Snapshots of a simulation of five robots exploring the cave environment  (\autoref{subfig:mycave}) at the times given in the captions.  \autoref{subfig:scr_t_0},  and \autoref{subfig:scr_t_897} show overhead views of the environment, while \autoref{subfig:scr_t_133} shows perspective view. The blue shaded region emanating from each robot displays the range of its laser sensors.  A red dotted line between two robots in \autoref{subfig:scr_t_133} indicates that the robots can communicate with each other. }  
	\label{fig:screenshots}
	\vspace{-5mm}
\end{figure}

\begin{figure}[!t]
	
	\begin{tabular}{ccc}
		\centering

		\subcaptionbox{ $0$ s \label{subfig:updt_t_0}}{\includegraphics[width=.3\linewidth]{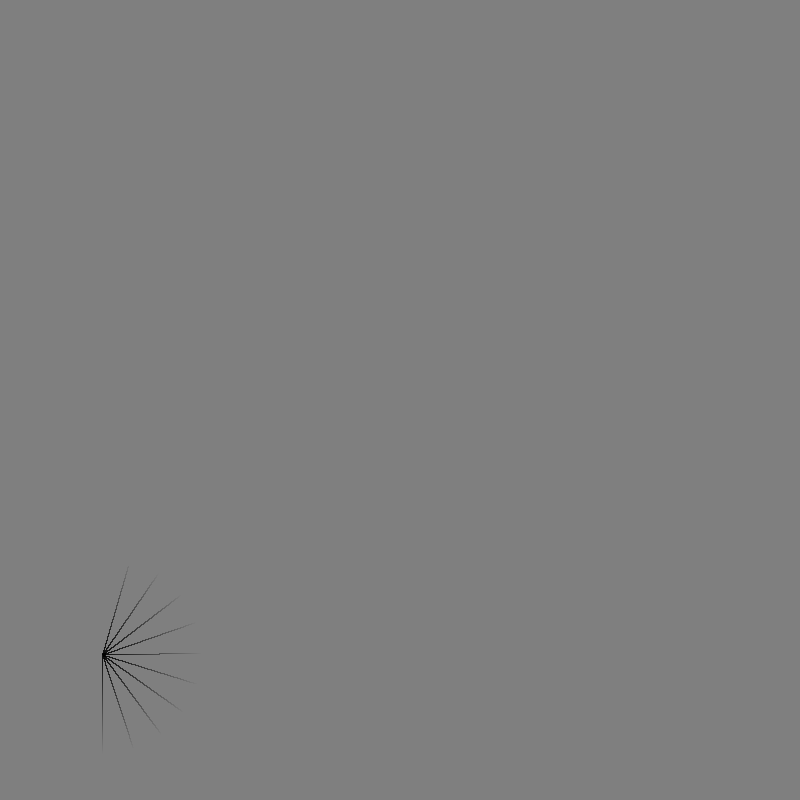}}
		&

		\subcaptionbox{ $133$ s \label{subfig:updt_t_133}}{\includegraphics[width=.3\linewidth]{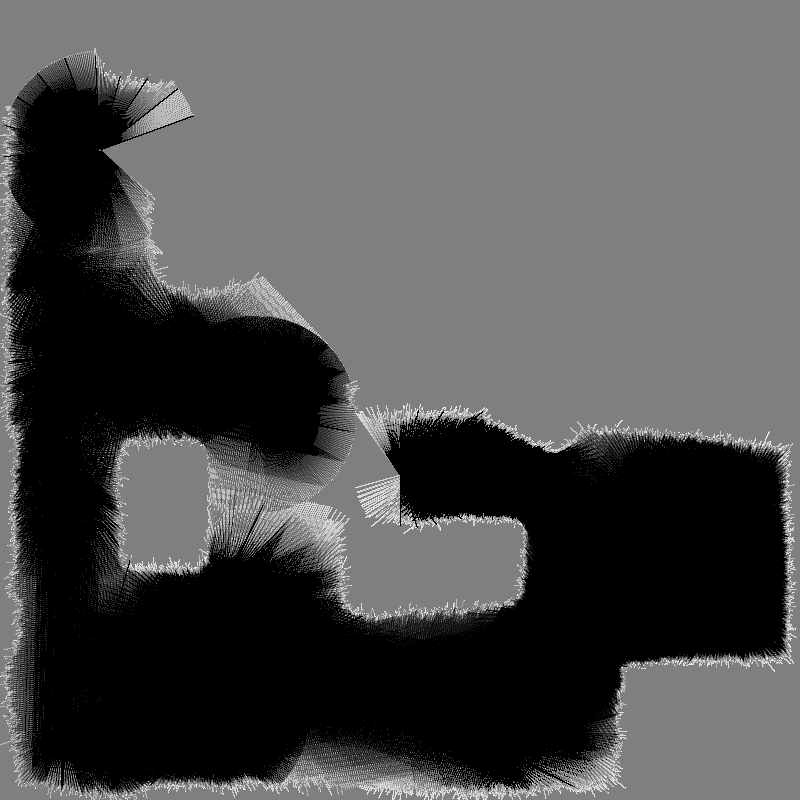}}
		
		&

		\subcaptionbox{ $897$ s \label{subfig:updt_t_897}}{\includegraphics[width=.3\linewidth]{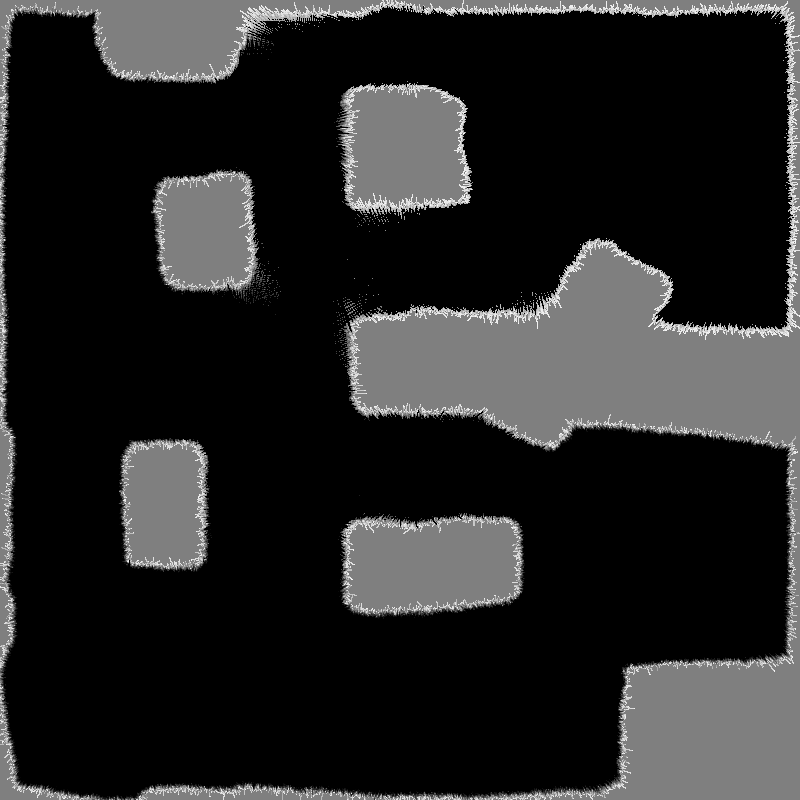}} 
		
	\end{tabular}
	\caption{The occupancy grid map generated by the robot inside the green dotted circle in the snapshots in \autoref{fig:screenshots} at the times of these snapshots.} 
	\label{fig:snapshots}
	\vspace{-5mm}
\end{figure}

%%%%%%%%%%%%%%%%%%%%%%%%%%%%%%%%%%%%%%%%%%%%%%%%%%%%%%%%%%%%%%%%%%%%%%%%%%%

%%%%%%%%%%%%collision%%%%%%%%%%%%%%%%%%%%%%%%%%%%%%%%%%%%%%%%

\begin{figure}[!t]
	
	\begin{tabular}{cc}
		\centering

		\subcaptionbox{  \label{subfig:robot collide}}{\includegraphics[width=.45\linewidth]{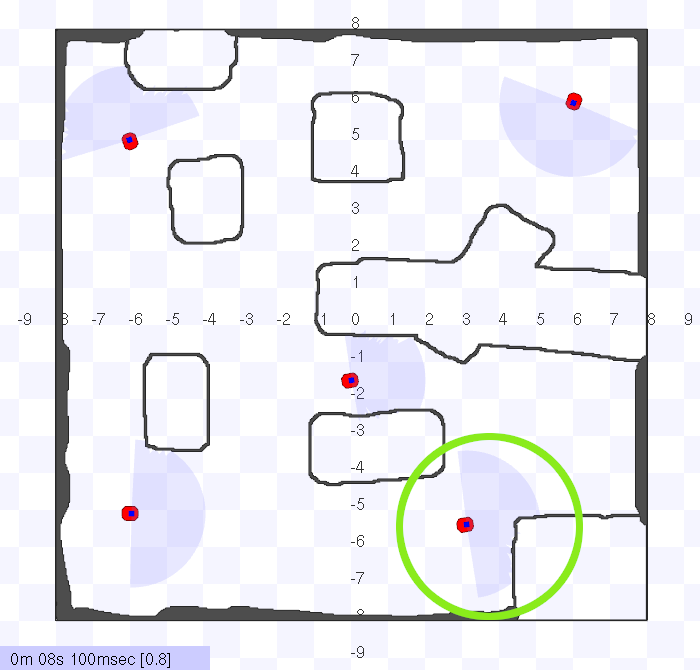}}
		&

		\subcaptionbox{  \label{subfig:robot avoid}}{\includegraphics[width=.45\linewidth]{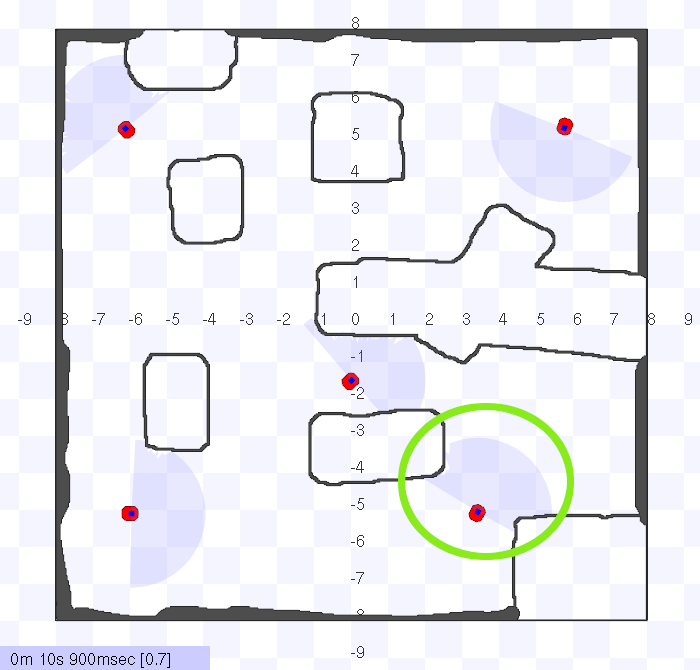}}

	\end{tabular}
	\caption{{The robot inside the green circle (a) detects an obstacle, then (b) follows a new heading computed from \hyperref[eqn:MI objective]{Eq.~\ref{eqn:MI objective}}.}} 
	\label{fig:collision}
	\vspace{-5mm}
\end{figure}

\begin{figure}[!t]

		\centering	
		%\hspace{-3mm}	
		\includegraphics[width=\linewidth, height=.60\linewidth]{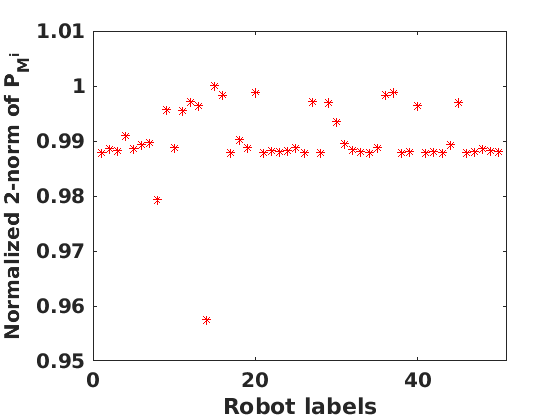}

\caption{Consensus over maps. The metric $\|\mathbf{P}_{M^i}\|_2 / \max_i \|\mathbf{P}_{M^i}\|_2$, where $\mathbf{P}_{M^i}$ is the set $\mathbf{\bar{P}}_{M^i}$ in matrix form, for each robot $R^i$ in a swarm of 50 robots that explore the environment in \autoref{subfig:frieburg} for $t_f = 3600$ s.}

\label{subfig:map_cons}
\end{figure}

%%%%%%%%%%%% Map coverage metric%%%%%%%%%%%%%%%%%%%%%%%%%%%%%%%%%%%%%%%%

\begin{figure}[!t]

		\centering	
		%\hspace{-3mm}	
		\includegraphics[width=\linewidth, height=.60\linewidth]{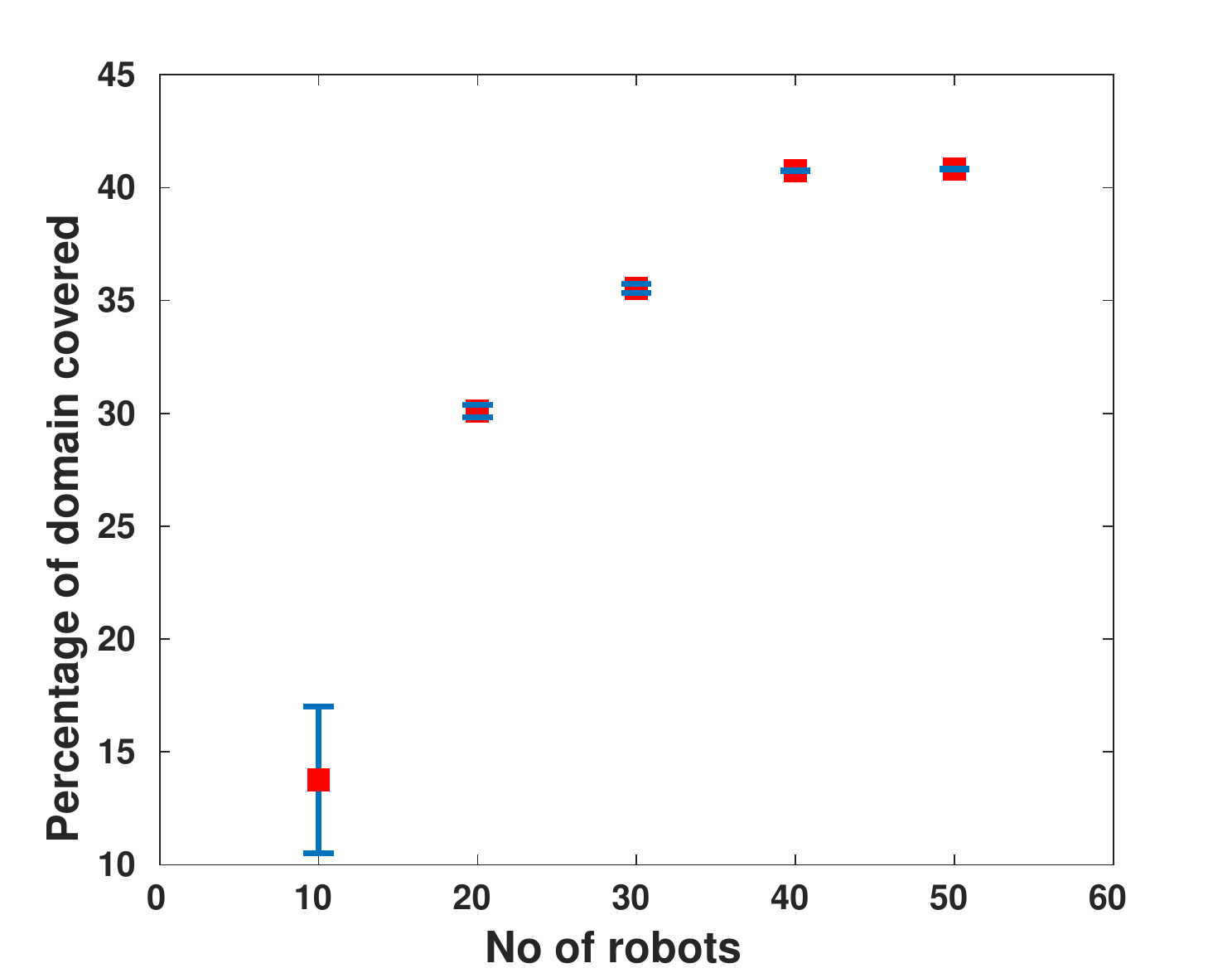}

\caption{Coverage vs. swarm size. Percentage of the environment in \autoref{subfig:frieburg} that is covered by swarms of different sizes after $t_f = 3600$ s. Red dots indicate the median  percentage over 10 simulations, and error bars show the $25^{th}$ and $75^{th}$ percentiles.}

\label{subfig:map_cov}
\end{figure}

%%%%%%%%%%%% Map Consensus%%%%%%%%%%%%%%%%%%%%%%%%%%%%%%%%%%%%%%%%

\begin{figure*}[!t]
	
	\begin{tabular}{cccc}
		\centering

		\subcaptionbox{Map from robot $R^1$ \label{subfig:map_cons_r1}}{\includegraphics[width=.22\linewidth, height=.20\linewidth]{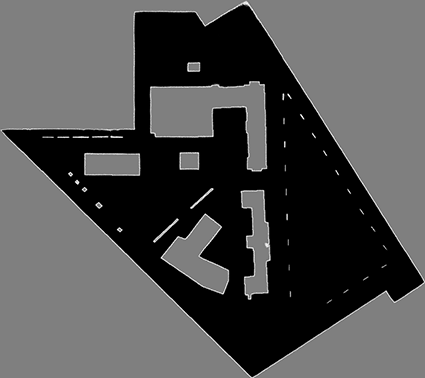}}
		&

		\subcaptionbox{Map from robot $R^4$ \label{subfig:map_cons_r4}}{\includegraphics[width=.22\linewidth, height=.20\linewidth]{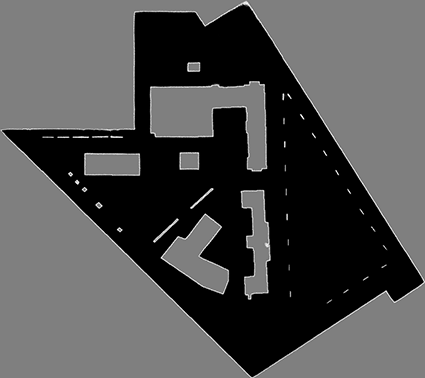}} 
		
		&

		\subcaptionbox{Map from robot  $R^8$\label{subfig:map_cons_r8}}{\includegraphics[width=.22\linewidth, height=.20\linewidth]{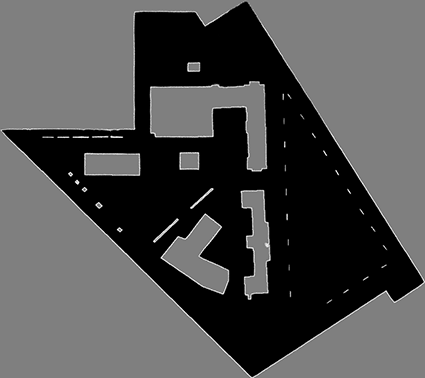}}
		&

		\subcaptionbox{Map from robot  $R^9$\label{subfig:map_cons_r9}}{\includegraphics[width=.22\linewidth, height=.20\linewidth]{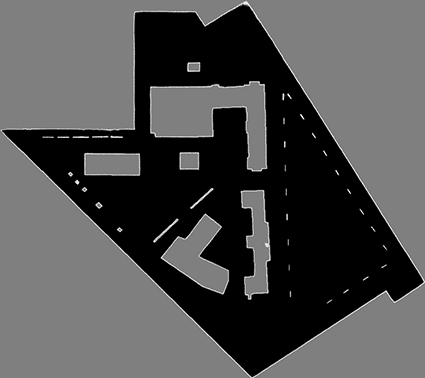}}

		\\
		
		\subcaptionbox{Map from robot  $R^{12}$\label{subfig:map_cons_r12}}{\includegraphics[width=.22\linewidth, height=.20\linewidth]{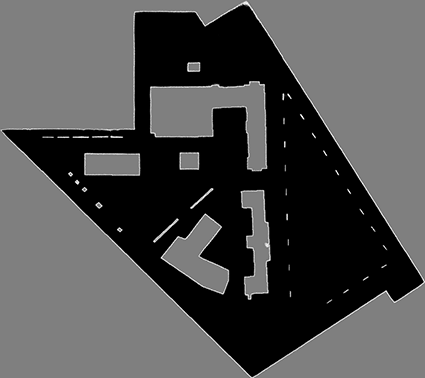}}
		&

		\subcaptionbox{Map from robot  $R^{14}$\label{subfig:map_cons_r14}}{\includegraphics[width=.22\linewidth, height=.20\linewidth]{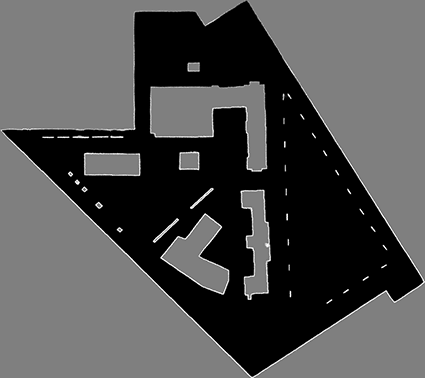}} 
		
		&

		\subcaptionbox{Map from robot $R^{37}$\label{subfig:map_cons_r37}}{\includegraphics[width=.22\linewidth, height=.20\linewidth]{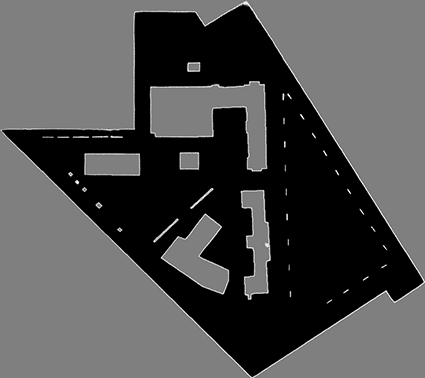}}
		&

		\subcaptionbox{ Map from robot $R^{45}$\label{subfig:map_cons_r45}}{\includegraphics[width=.22\linewidth, height=.20\linewidth]{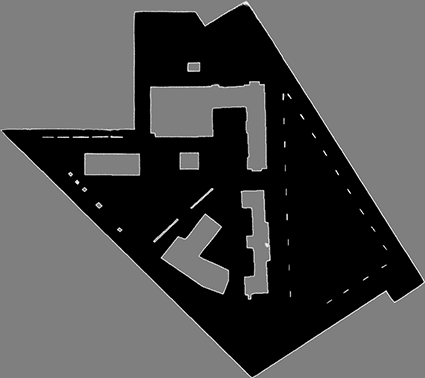}}
		
	\end{tabular}
	\caption{%Consensus in the 
	Occupancy grid maps computed by 8 members of a swarm of $50$ robots after exploring the environment in \autoref{subfig:frieburg} for 3600 s. 
	%A swarm of fifty robots explored an environment of size $90m\times80m$ having a layout as shown in \autoref{subfig:frieburg}. 
	%The captions of under each figure indicates label of the robot that constructed the map.  
	}  
	\label{fig:map consensus}
\end{figure*}

%%%%%%%%%%%%%%%%%%%%%%%%%%%%%%%%%%%%%%%%%%%%%%%%%%%%%%%%%%%%%%%%%%%%%%%%%%%%

%%%%%%%%%%%% Entropy %%%%%%%%%%%%%%%%%%%%%%%%%%%%%%%%%%%%%%%%

\begin{figure*}[!t]
	
	\begin{tabular}{ccc}
		\centering
		\subcaptionbox{Unobstructed environment\label{subfig:entro plain}}{\includegraphics[width=.32\linewidth, height=.25\linewidth]{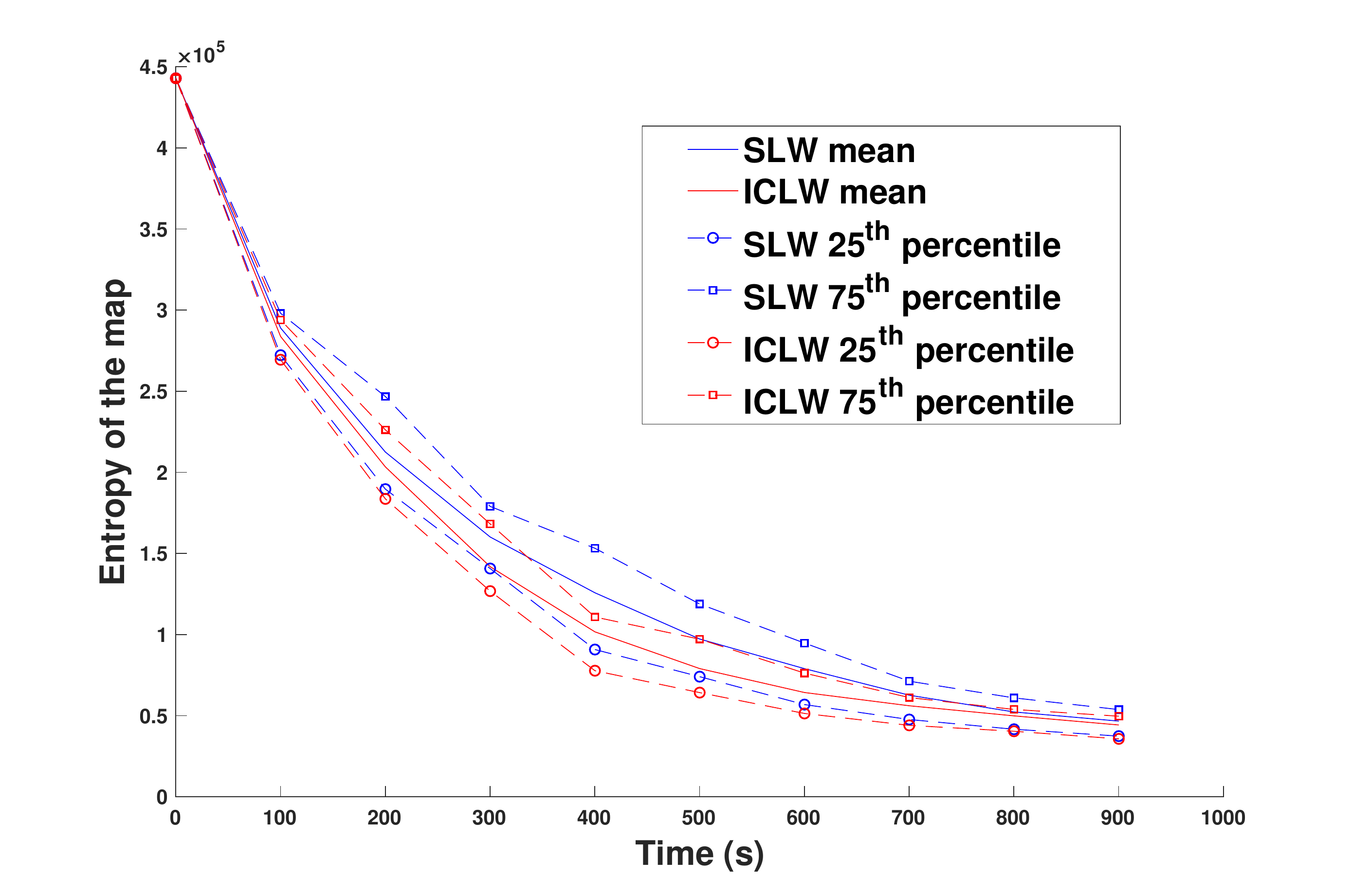}}
		
		& 
		
		\subcaptionbox{Cave environment \label{subfig:entro cave}}{\includegraphics[width=.32\linewidth, height=.25\linewidth]{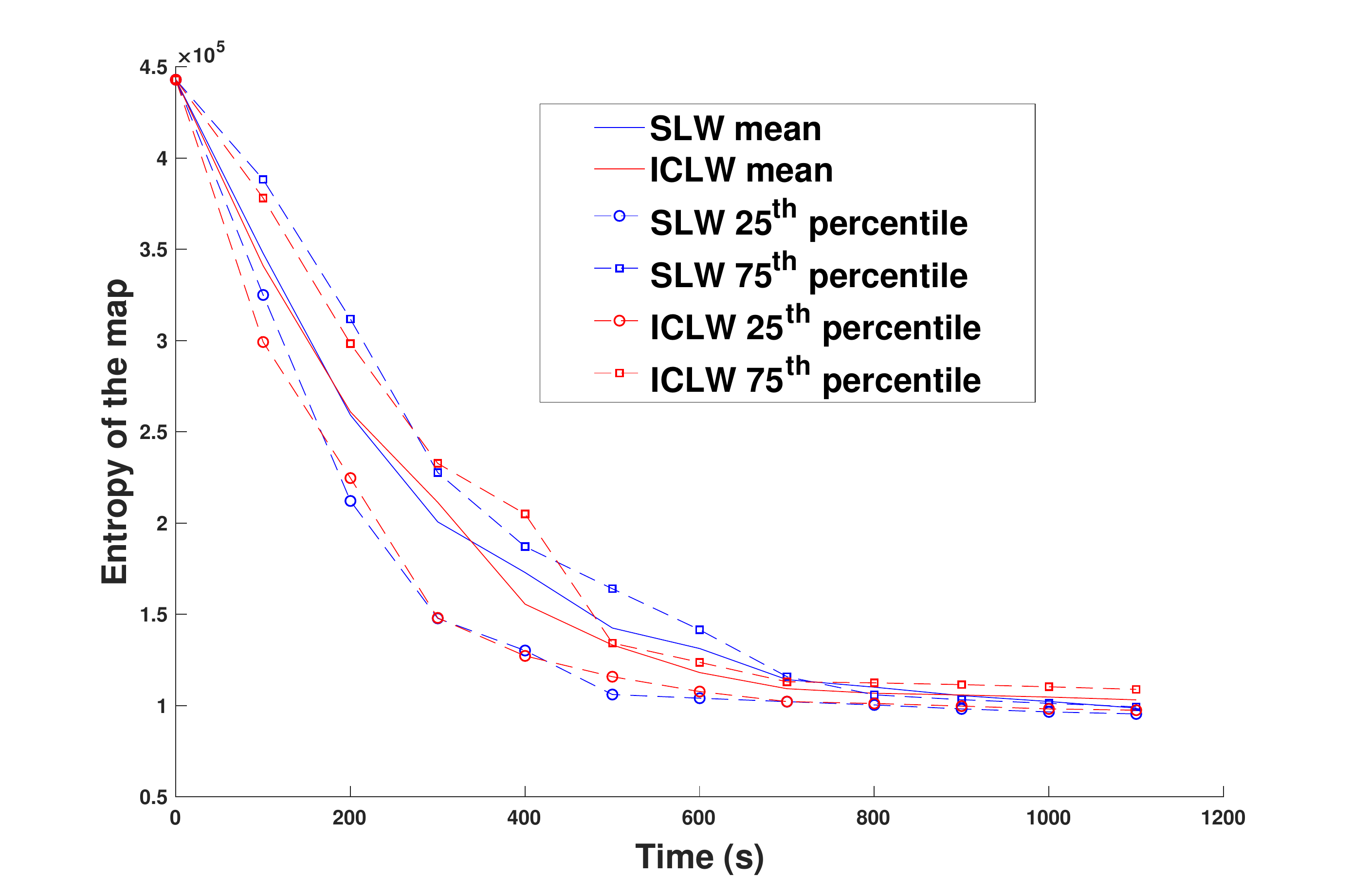}}
		&

		\subcaptionbox{Autonomy laboratory\label{subfig:entro autolab}}{\includegraphics[width=.32\linewidth, height=.25\linewidth]{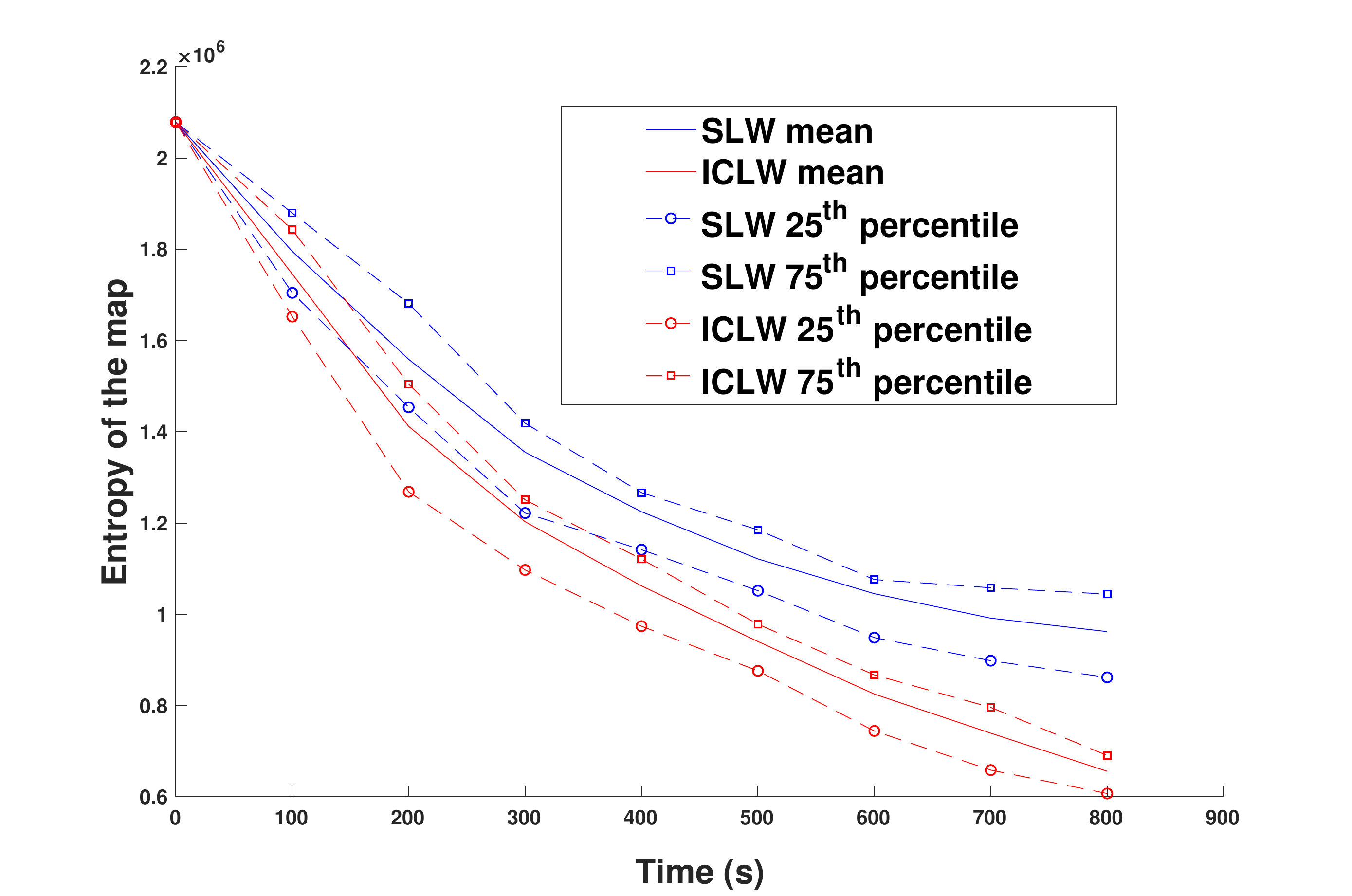}} 
		
	\end{tabular}
	\caption{Time evolution of the entropy $H(M^i)$, computed using \autoref{eqn:map entropy}, of occupancy maps $M^i$, $i =  1,...,N_R$,  
	%\spr{(So there are 5 \rag{ (no, $5\times 30$, 5 robots performing 30 trails)} plots for each type of line in (a) and (b) and 20 plots for each type of line in (c)?)}  
	from simulations of: (a) $N_R = 5$ robots mapping the unobstructed environment % with layout 
	in \autoref{subfig:plain};
	%and size $16m\times16m$, 
	%using a swarm of $5$ robots; 
	(b) $N_R = 5$ robots mapping the cave environment %with layout
	in \autoref{subfig:mycave}; and
	%and size $16m\times16m$, 
	%using a swarm of $5$ robots; and 
	(c) $N_R = 20$ robots mapping the autonomy laboratory %with layout
	in \autoref{subfig:autolab}.
	%and size $40m\times30m$, 
	%using a swarm of $20$ robots.
	For each environment, 30 simulation trials were run in which the robots performed a standard L\'evy walk {(SLW)}, and 30 trials were run in which they performed the information correlated L\'evy walk {(ICLW)} presented in \autoref{sec:Exploration based on information Correlated Levy Walk}.
	%variation of occupancy grid map's entropy %\cite{Thrun:2005:PR:1121596}
	%over time of exploration, 
	%when robots explored the domain using L\'evy walk strategy and with our exploration strategy (\autoref{sec:Exploration based on information Correlated L\'evy Walk}). 
	%The simulated experiments were repeated $30$ times. 
	%The x-axis of the plots indicates time of exploration.
	%Y-axis represents the entropy of occupancy grid map. 
%	\spr{The line types in the plot legends are defined as follows. }
	The following plots are shown in each subfigure:
	%Legends of the plots are as follows: 
	mean entropy over 30 trials with the SLW {\it (solid blue lines)}; 
	%for standard L\'evy walk strategy; 
	mean entropy over 30 trials with the ICLW {\it (solid red lines)};
	%for our information correlated exploration and mapping strategy (indicated as MI L\'evy walk in plot legends);  
	 $25^{th}$ percentile of the 30 trials 
	 with the SLW
	 %in the case of standard L\'evy walk 
	 {\it (blue dashed lines with circles)};  $75^{th}$ percentile of the 30 trials 
	 %in the case of standard L\'evy walk 
	 with the SLW {\it (blue dashed lines with squares)};  $25^{th}$ percentile of the 30 trials with the ICLW 
	 %in the case of our strategy (MI L\'evy walk) 
	 {\it (red dashed lines with circles)};  $75^{th}$ percentile of the 30 trials with the ICLW
	 %in the case of our strategy (MI L\'evy walk) 
	 {\it (red dashed lines with squares)}.}
	\label{fig:entropy}
\end{figure*}

%%%%%%%%%%%%%%%%%%%%%%%%%%%%%%%%%%%%%%%%%%%%%%%%%%%%%%%%%%%%%%%%%%%%%%%%%%%%

\begin{figure}[!t]

		\centering	
		%\hspace{-3mm}	
		\includegraphics[ height=0.5\linewidth]{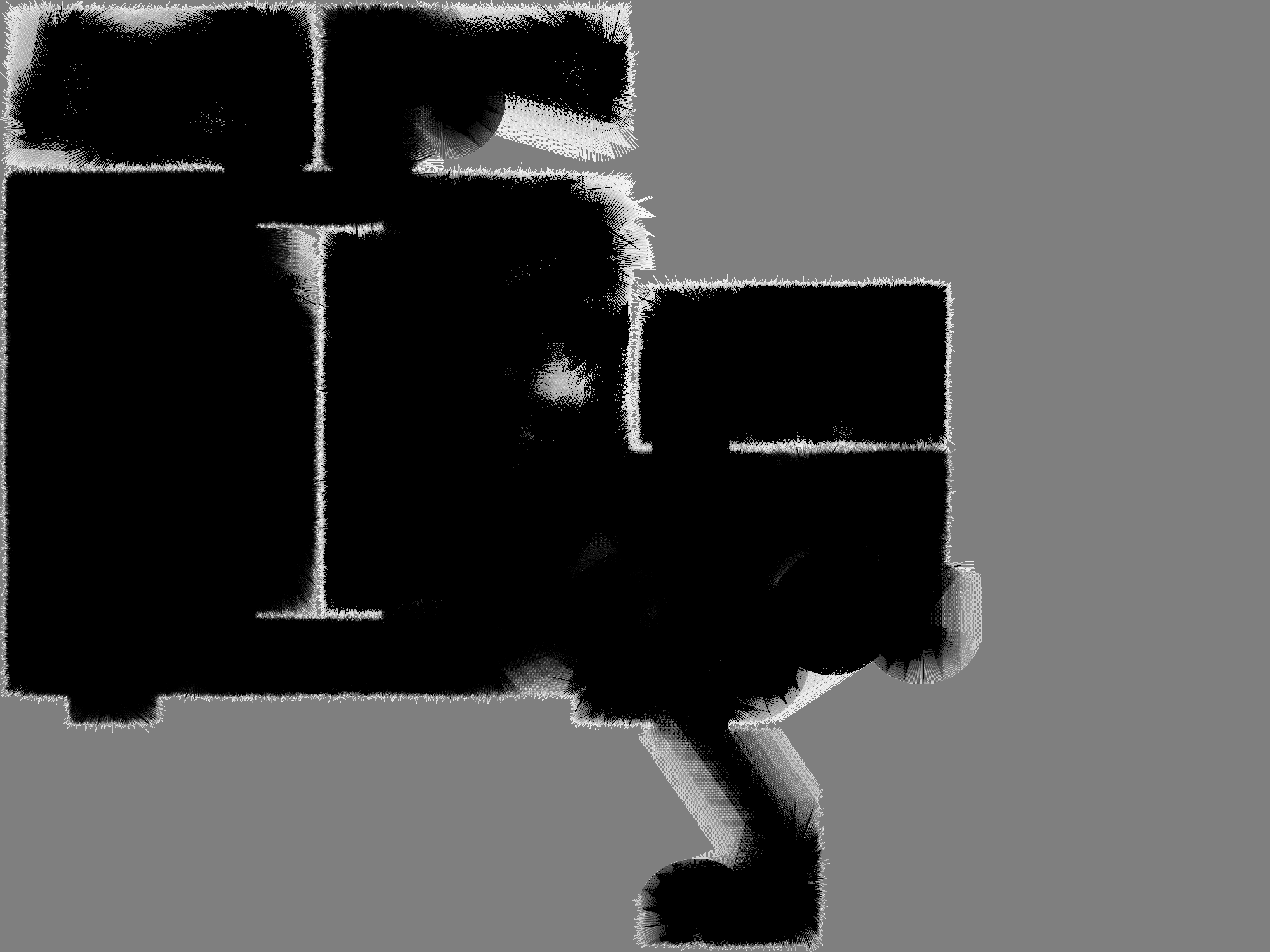}

\caption{{Map of the autonomy laboratory environment (\autoref{subfig:autolab}) generated by robots performing a SLW.}}

\label{fig:autolab_levy}
\end{figure}

%%%%%%%%%%%%%%%%%%%%%%%%%%%%%%%%%%%%%%%%%%%%%%%%%%%%%%%%%%%%%%%%%%%%%%%%%%%%

%%%%%%%%%%%%%%%%%%%%%%%%%%%%%%%%%%%%%%%%%%%%%%%%%%%%%%%%%%%%%%%%%%%%%%%%%%%%

\begin{figure}[!t]

		\centering	
		%\hspace{-3mm}	
		\includegraphics[width=0.9\linewidth,height=0.5\linewidth]{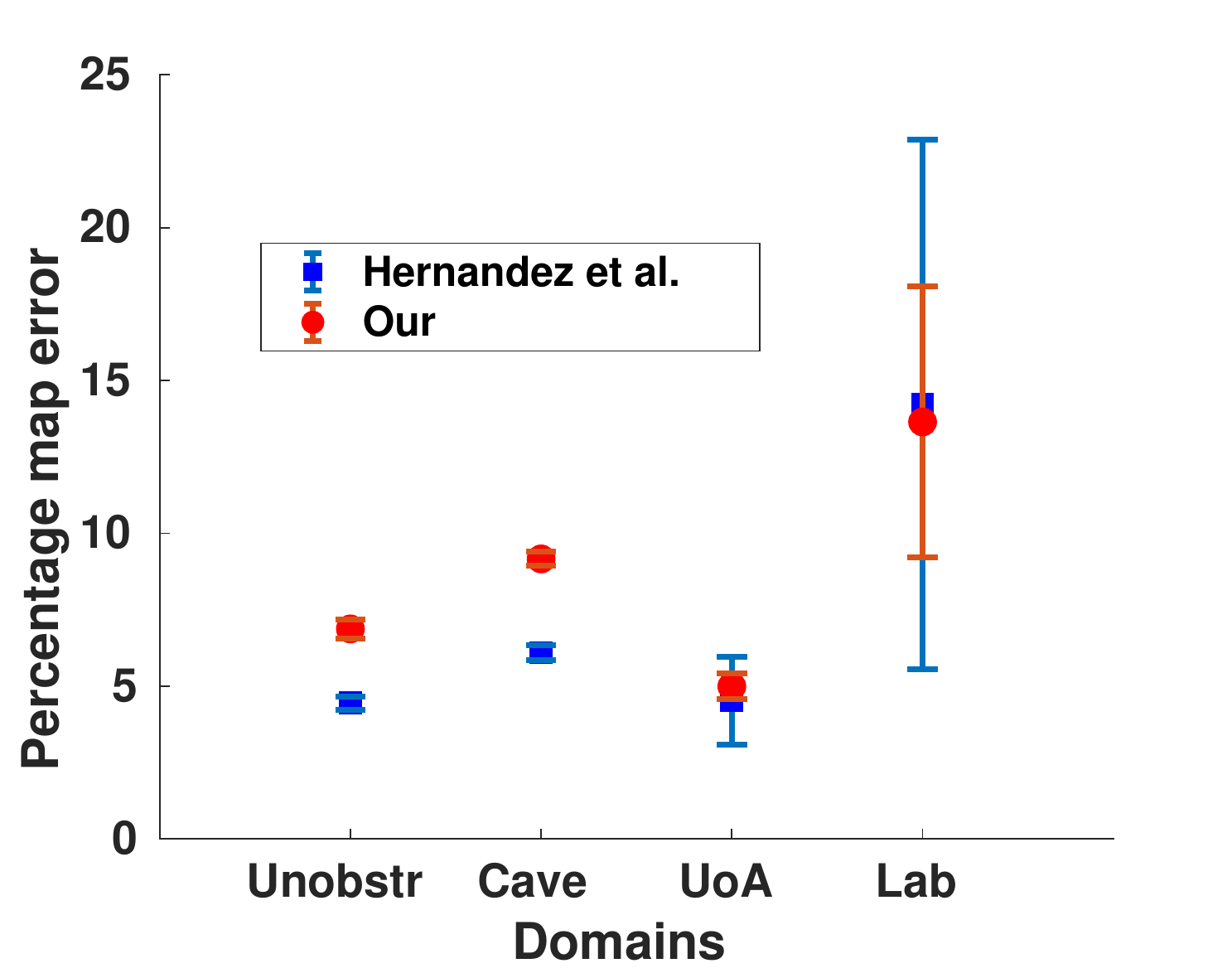}

\caption{\fnl{Comparison of our strategy with the {multi-robot mapping} strategy  in Hern{\'a}ndez \textit{et al}. \cite{hernandez2020real}. The comparison is performed for the environments in \autoref{subfig:plain}, \autoref{subfig:mycave}, \autoref{subfig:uoa_robotics_lab}, and \autoref{subfig:autolab}. {For each environment, the dot and square represent the median of the maximum error in the maps produced by our strategy and by the one in \cite{hernandez2020real}, respectively, over 10 simulation trials. The corresponding error bars indicate the} largest and smallest maximum errors over the trials.}}

\label{fig:map error comparison}
\end{figure}

%%%%%%%%%%%%%%%%%%%%%%%%%%%%%%%%%%%%%%%%%%%%%%%%%%%%%%%%%%%%%%%%%%%%%%%%%%%%

\begin{figure}[!t]

		\centering	
		\begin{tabular}{cc}
		%\hspace{-3mm}	
		\subcaptionbox{\label{subfig:alpha}}{\includegraphics[width=0.45\linewidth, height=0.5\linewidth]{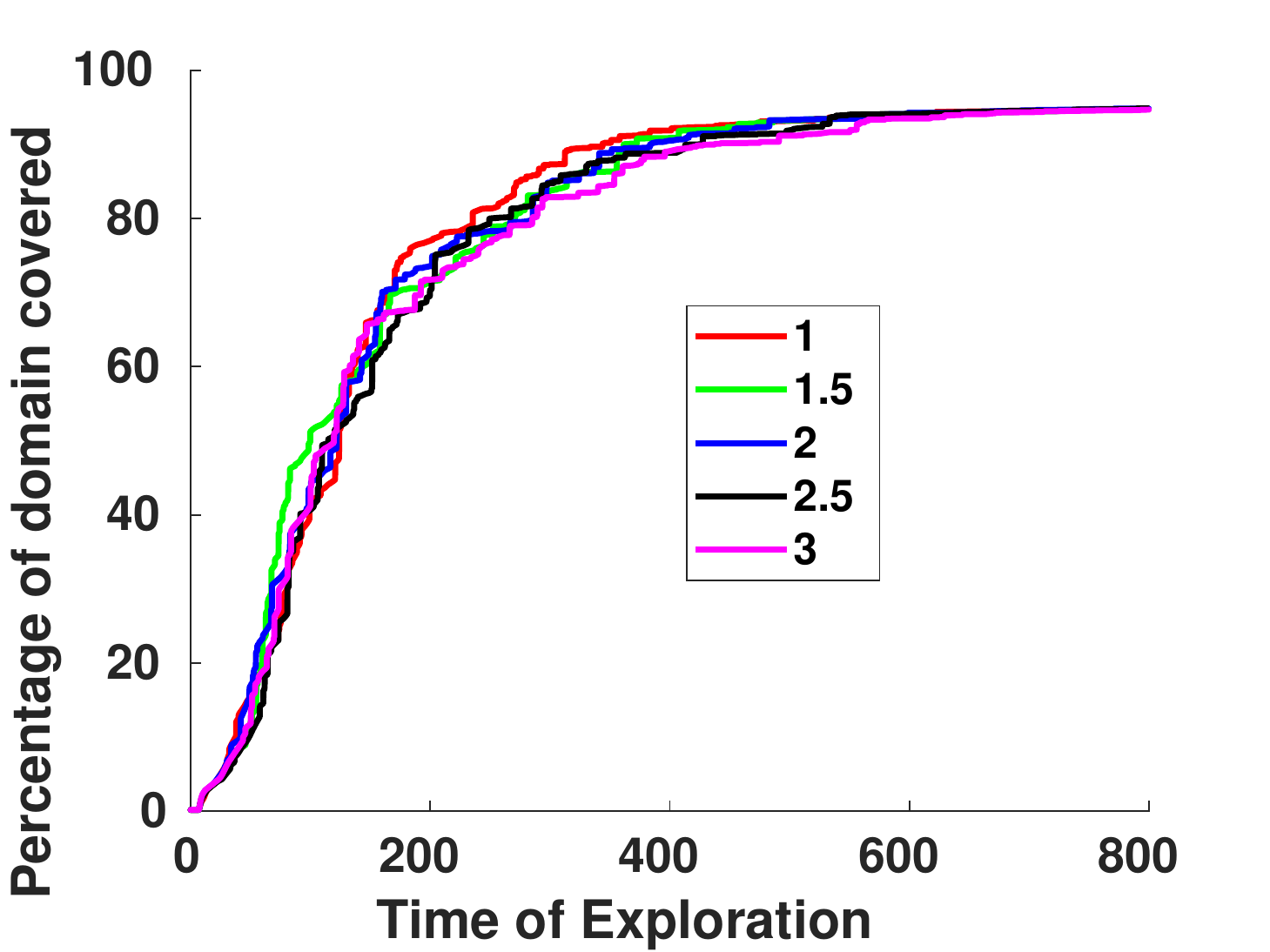}}
		&   
		
		\subcaptionbox{\label{subfig:sigma}}{\includegraphics[width=0.45\linewidth, height=0.5\linewidth]{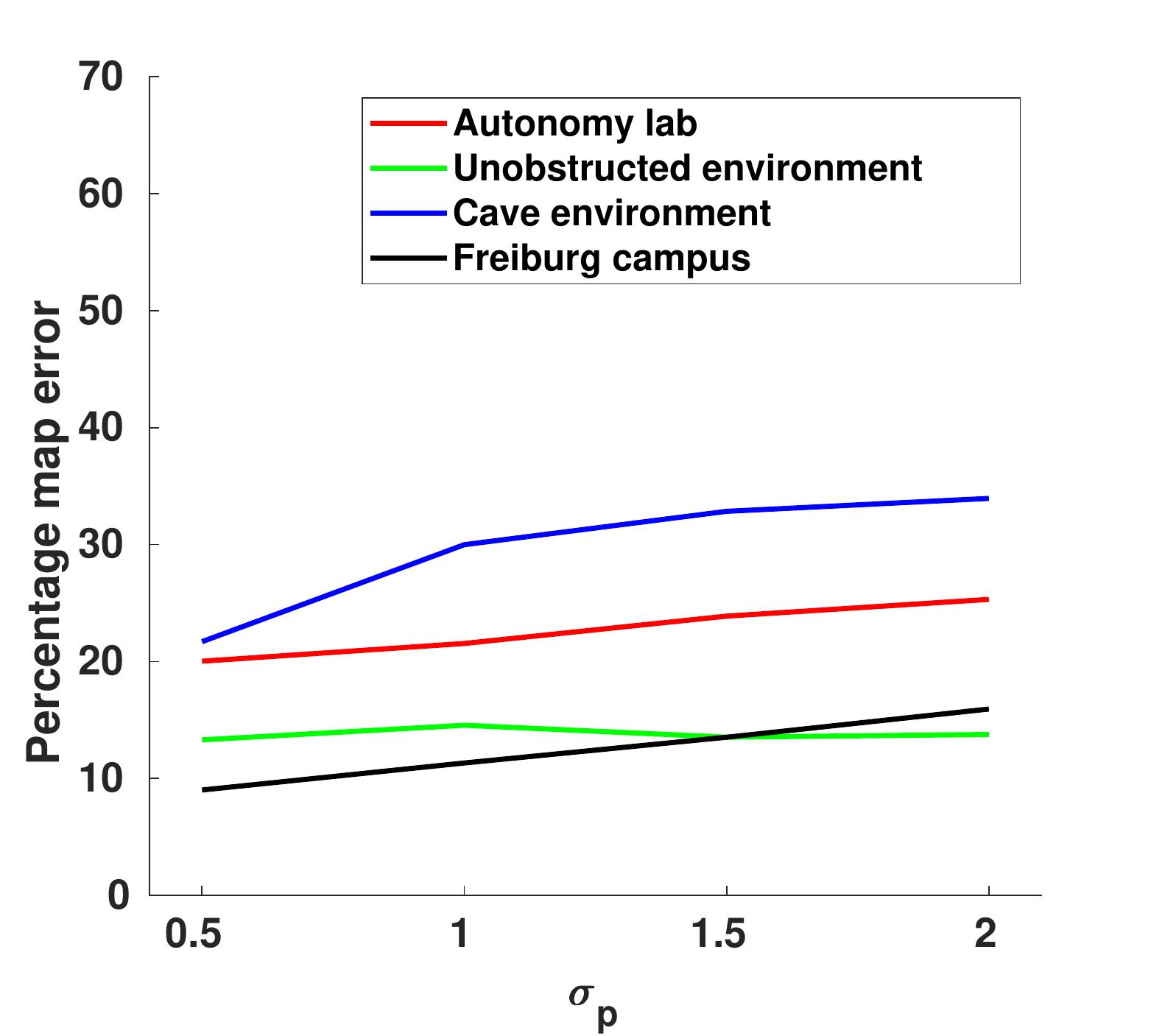}}
		
       \end{tabular}

\caption{{(a) Time evolution of the mean minimum percentage of the unobstructed environment (\autoref{subfig:plain}) covered by the robots for different $\alpha$ values. (b) The mean maximum percent error in the maps for different robot localization uncertainties ($\sigma_p$) in four environments.}}

\label{fig:validation}
\vspace{-5mm}
\end{figure}

%%%%%%%%%%%% Coverage %%%%%%%%%%%%%%%%%%%%%%%%%%%%%%%%%%%%%%%%

\begin{figure*}[!t]
	
	\begin{tabular}{ccc}
		\centering	
		\subcaptionbox{Unobstructed environment\label{subfig:cover plain}}{\includegraphics[width=.32\linewidth, height=.25\linewidth]{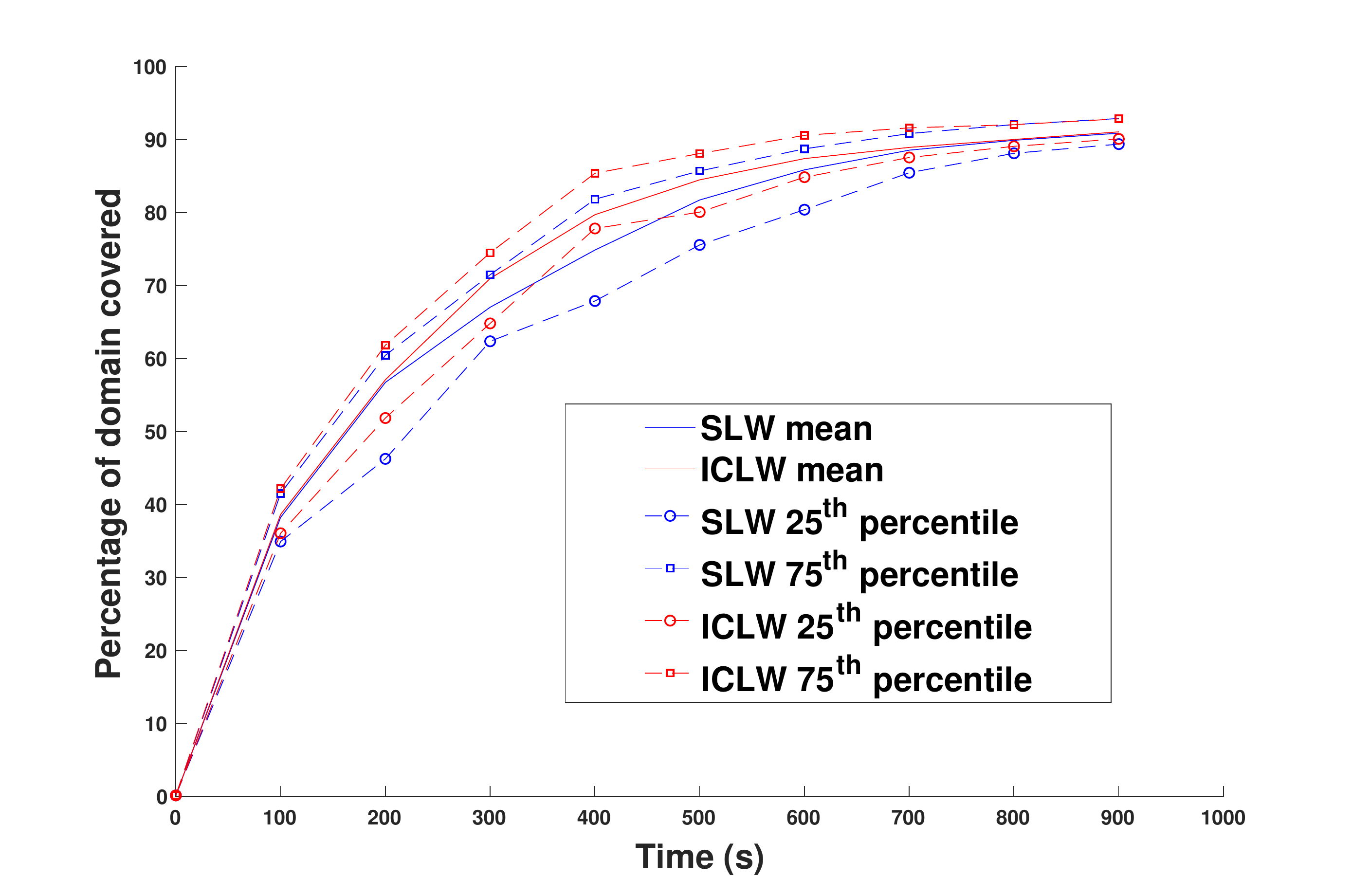}}
		
		&
		
		\subcaptionbox{Cave environment \label{subfig:cover cave}}{\includegraphics[width=.32\linewidth, height=.25\linewidth]{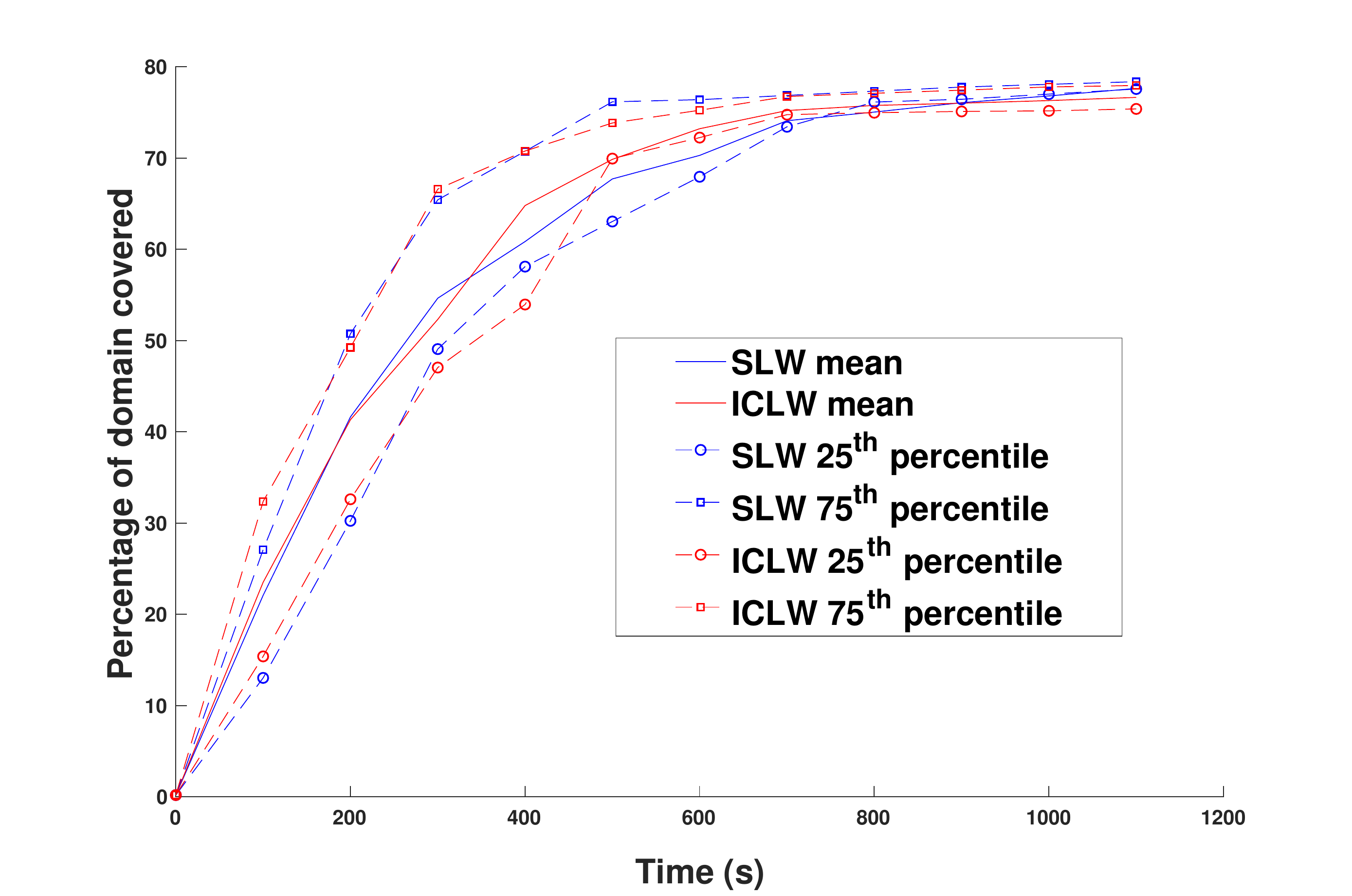}}
		&

		\subcaptionbox{Autonomy laboratory\label{subfig:cover autolab}}{\includegraphics[width=.32\linewidth, height=.25\linewidth]{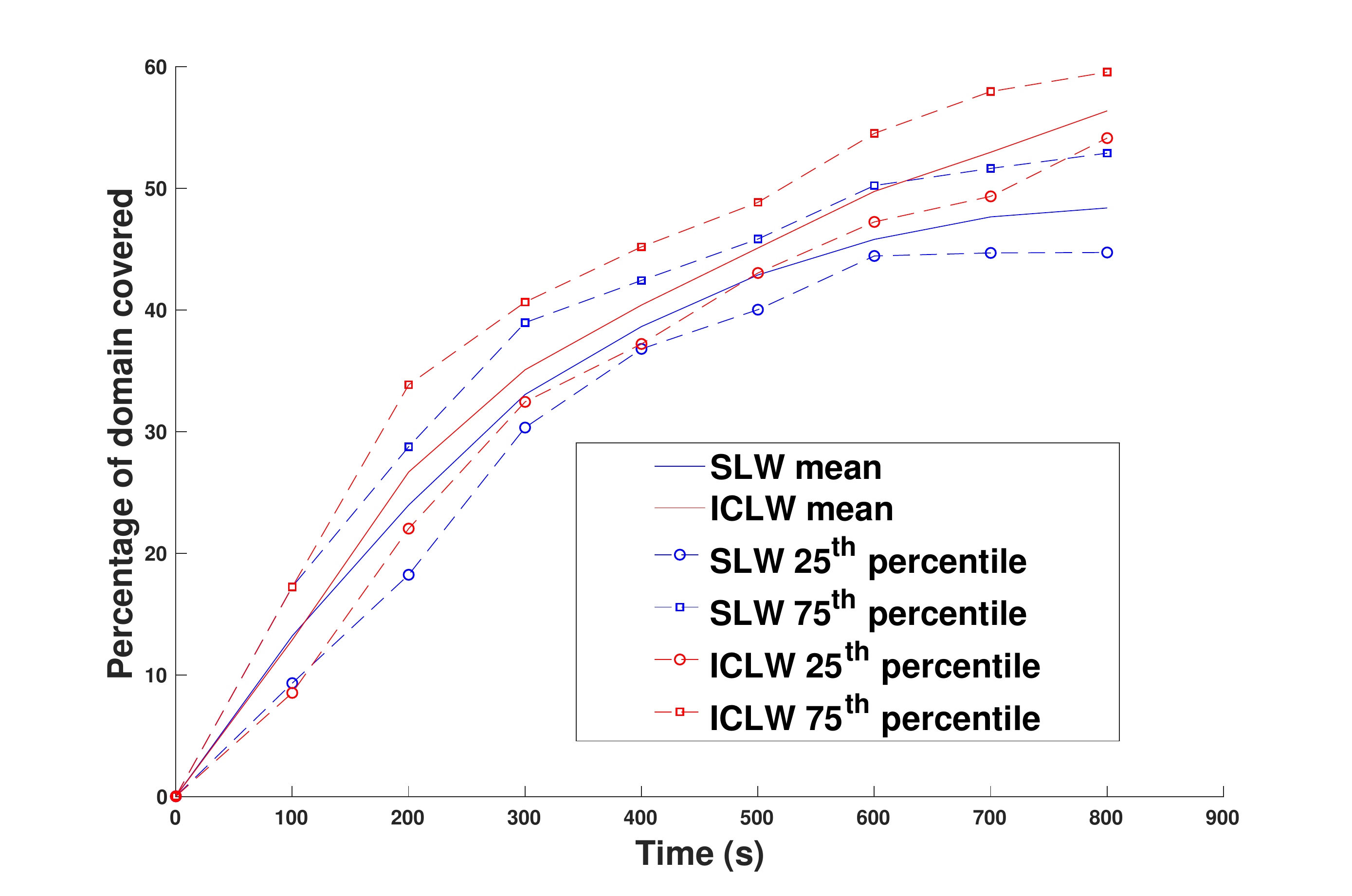}} 
	\end{tabular}
	\caption{Time evolution %variation 
	of the percentage of the domain area covered by the robots during the same simulations as in \autoref{fig:entropy}.
	%vs time of exploration, 
	%(a) Results of the simulations conducted on a cave environment with layout \autoref{subfig:mycave} and size $16m\times16m$, using a swarm of $5$ robots. (b) Results of the simulations conducted on an  autonomy lab with layout \autoref{subfig:autolab} and size $40m\times30m$, using a swarm of $20$ robots. (c) Results of the simulations conducted on a plain with layout \autoref{subfig:plain} and size $16m\times16m$, using a swarm of $5$ robots. 
	%The simulated experiments were repeated $30$ times.  
	%when robots explored the domain using L\'evy walk strategy and with our exploration strategy (\autoref{sec:Exploration based on information Correlated Levy Walk}). 
	%The x-axis of the plots indicates time of exploration. Y-axis represents the percentage of covered. 
	The plot legends are the same as in \autoref{fig:entropy}.
	%Legends of the plots are as follows: solid blue line :  mean percentage of domain covered for L\'evy walk strategy, solid red line : mean percentage of  domain covered for our information correlated exploration and mapping strategy(indicated as MI L\'evy walk in plot legends),  dash with circle blue line : $25^{th}$ percentile value of  trails in the case of standard L\'evy walk, dash with square blue line : $75^{th}$ percentile value of  trails in the case of standard L\'evy walk, dash with circle red line : $25^{th}$ percentile value of  trails in the case of our strategy(MI L\'evy walk), dash with square blue line : $75^{th}$ percentile value of  trails in the case of our strategy(MI L\'evy walk).
	}
	\label{fig:coverage}
\end{figure*}

%%%%%%%%%%%%%%%%%%%%%%%%%%%%%%%%%%%%%%%%%%%%%%%%%%%%%%%%%%%%%%%%%%%%%%%%%%%%

%%%%%%%%%%%% Barcode %%%%%%%%%%%%%%%%%%%%%%%%%%%%%%%%%%%%%%%%

\begin{figure*}[!t]
	
	\begin{tabular}{cccc}
		\centering

		\subcaptionbox{$H_0$, Cave environment \label{subfig:bar_0_cave}}{\includegraphics[width=.22\linewidth, height=.20\linewidth]{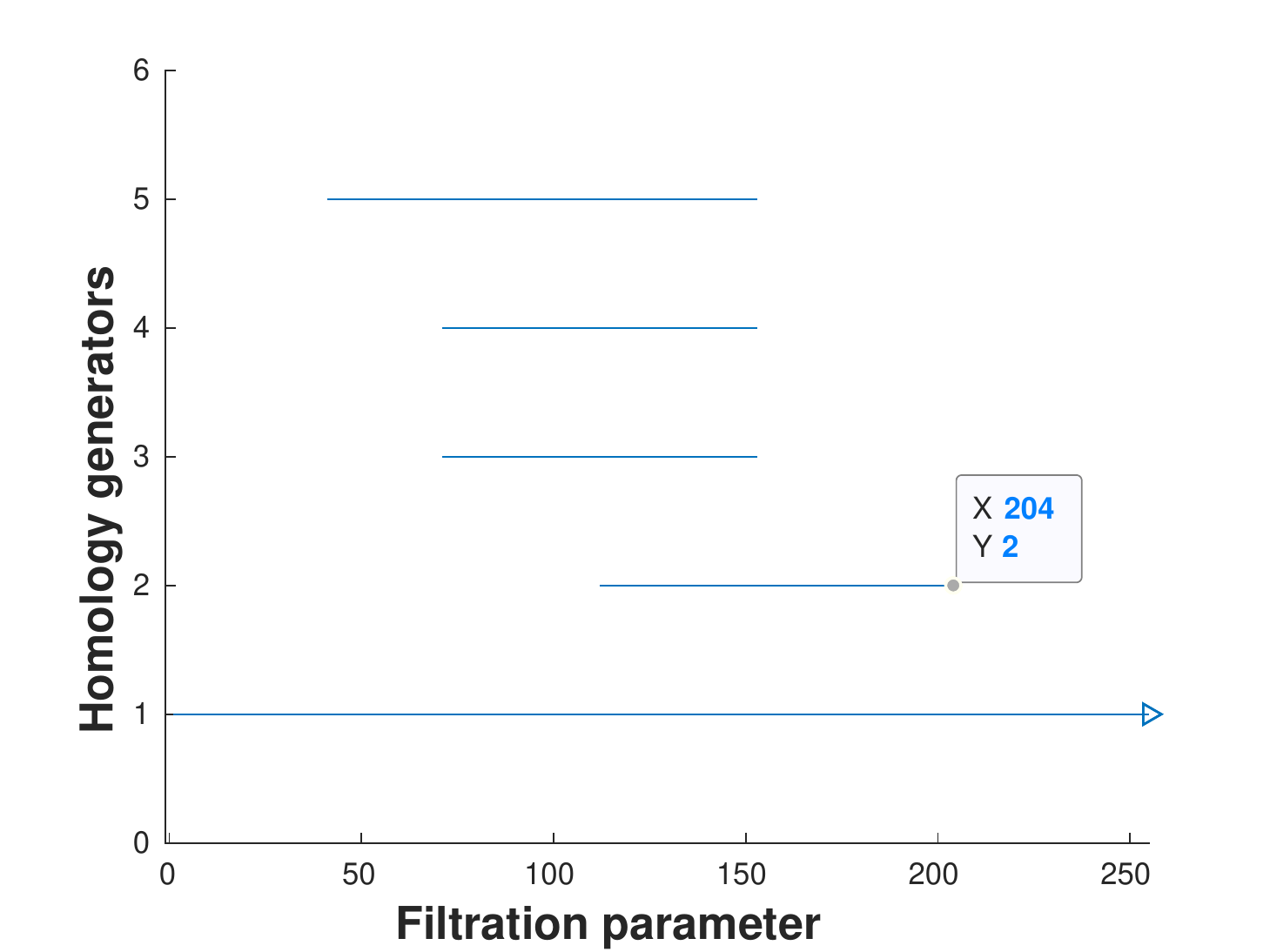}}
		&

		\subcaptionbox{$H_1$, Cave environment \label{subfig:bar_1_cave}}{\includegraphics[width=.22\linewidth, height=.20\linewidth]{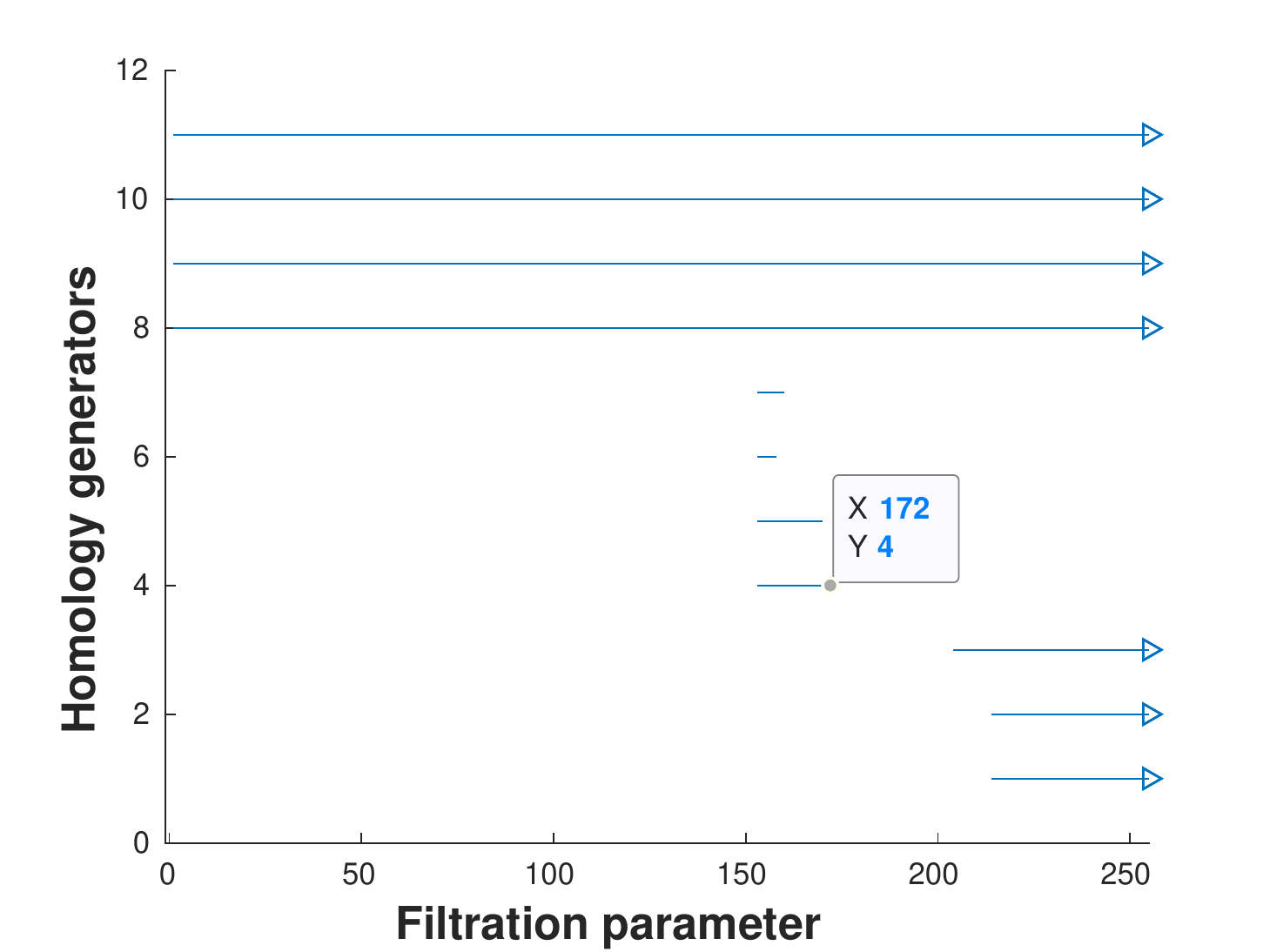}} 
		
		&

		\subcaptionbox{$H_0$, Autonomy lab \label{subfig:bar_0_autolab}}{\includegraphics[width=.22\linewidth, height=.20\linewidth]{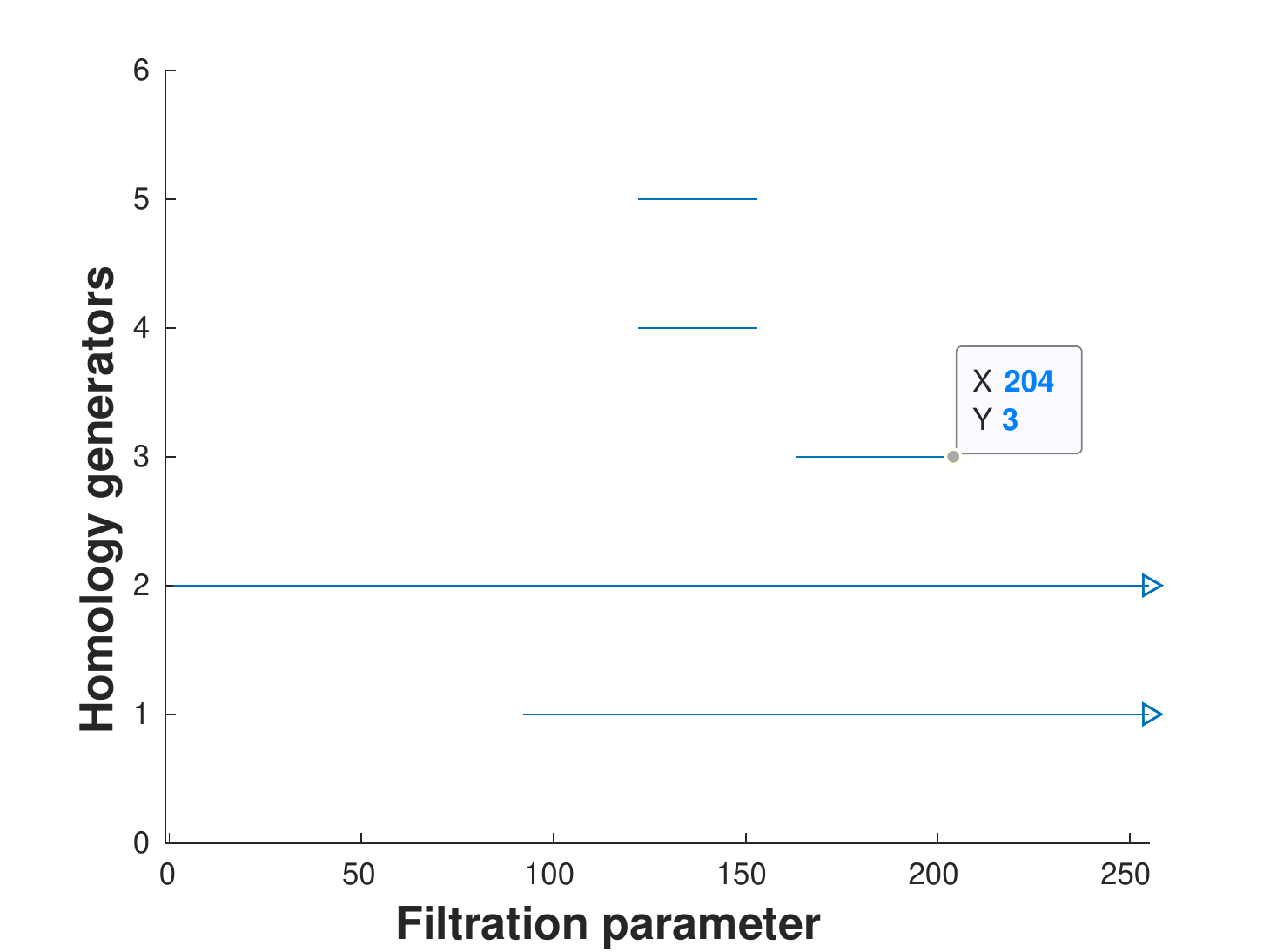}}
		&

		\subcaptionbox{$H_1$, Autonomy lab \label{subfig:bar_1_autolab}}{\includegraphics[width=.22\linewidth, height=.20\linewidth]{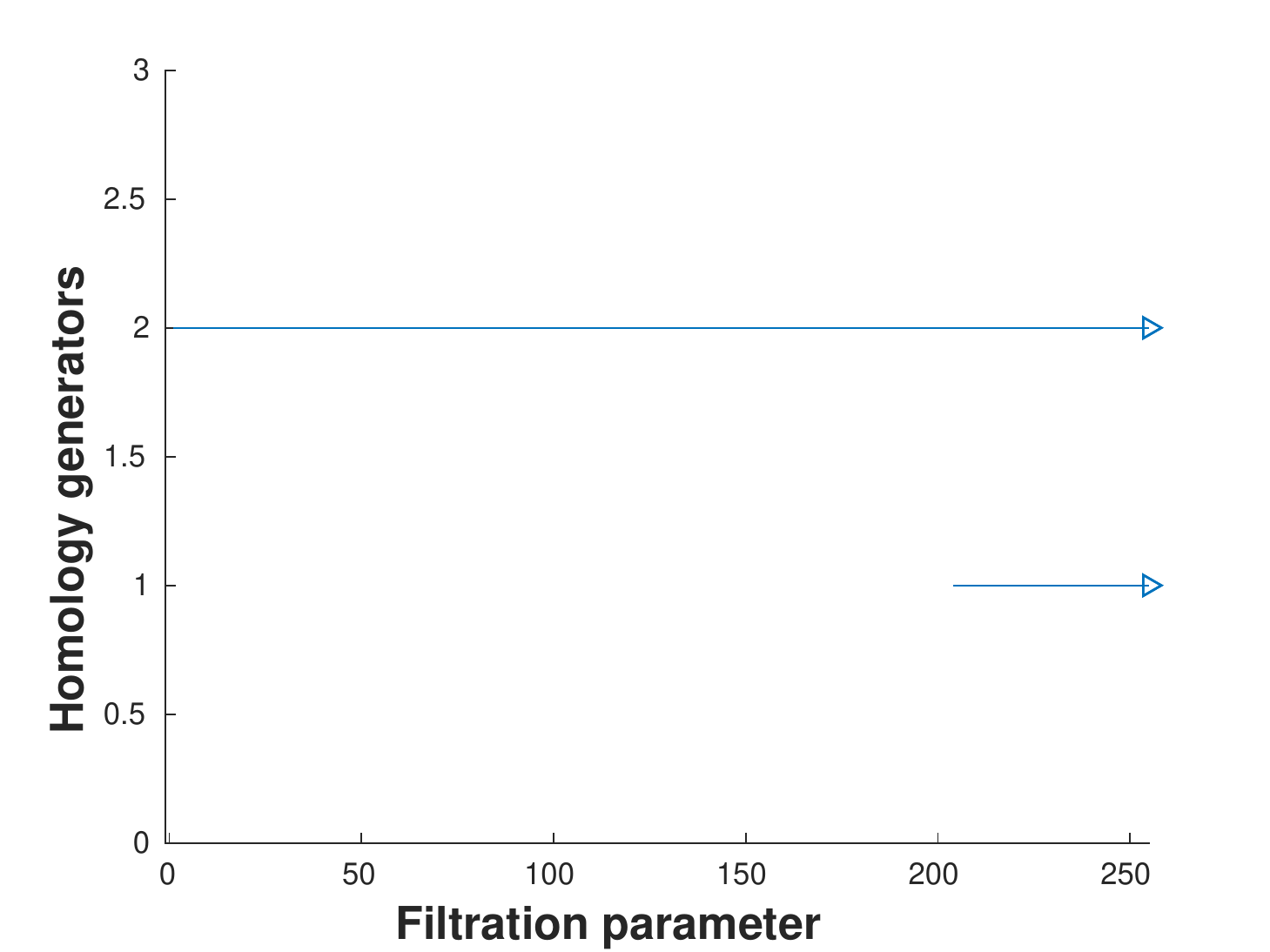}}
		
	\end{tabular}
	\caption{Barcode diagrams of the occupancy grid maps in \autoref{subfig:mycave map} (\autoref{subfig:bar_0_cave}, \autoref{subfig:bar_1_cave}) and \autoref{subfig:autolab map} (\autoref{subfig:bar_0_autolab}, \autoref{subfig:bar_1_autolab}). The filtration parameter $\varpi$ along the $x$-axes represents the pixel intensity of grid cells $m^i_j$ in the map, defined as the values of $\mathbb{P}_{m^i_j}$ scaled between 0 and 255. The filtration parameter ranges from 0 to 250.  
	 %The pixel intensity of each grid cell $m^i_j$ in the map is the value of $\mathbb{P}_{m^i_j}$ scaled between 0 and 255. 
	 The $y$-axes show the number of homology generators for dimensions zero (\autoref{subfig:bar_0_cave}, \autoref{subfig:bar_0_autolab}) and one (\autoref{subfig:bar_1_cave}, \autoref{subfig:bar_1_autolab}).} %\autoref{subfig:bar_0_cave} and \autoref{subfig:bar_1_cave} are barcode diagrams generated for map shown in \autoref{subfig:mycave map}.
	 %\autoref{subfig:bar_0_autolab} and \autoref{subfig:bar_1_autolab} are barcode diagrams generated for map shown in \autoref{subfig:autolab map}. 
	\label{fig:barcode}
\end{figure*}

%%%%%%%%%%%%%%%%%%%%%%%%%%%%%%%%%%%%%%%%%%%%%%%%%%%%%%%%%%%%%%%%%%%%%%%%%%%%

\section{SIMULATION RESULTS}
\label{sec:simulation}

In this section, we validate our distributed mapping technique with kinematic robots in five simulated environments with different sizes, shapes, and layouts, shown in \autoref{fig:worlds}, using the robot simulator Stage \cite{Vaughan2008}. The robots are controlled with velocity commands and have a maximum speed of $40$ cm/s, and they are equipped with on-board laser range sensors with a maximum range of $2$ m and a field-of-view of $180^{\circ}$.
%\rag{The field of view of the laser range sensor is an arc of central angle $180^{\circ}$, such that the direction of motion of the robot bisects this arc.} 
Each robot can communicate with any robot located within a circle of radius $2$ m. We set $p_f = 0.1$, $p_a = 0.5$, and $p_{hit} = 0.9$ in \autoref{eqn:reflectance model} and \autoref{eqn:unreflectance model}. In order for the robots to perform the information correlated L\'evy walk in the superdiffusive regime, {the L\'evy exponent $\alpha$ was set to $1.5$.}
%We used the same parameter value for the experiments.

%the proposed approach of information based exploration and distributed mapping technique is assessed in simulation and the results obtained are presented. 
%We implemented the proposed methodology using 

% For all simulations and experiments, 

For each environment in \autoref{fig:worlds}, we simulated a swarm of $N_R$ robots that explored the domain for $t_f$ seconds. \autoref{fig:generated maps} shows the occupancy grid map that was generated by an arbitrary robot in the swarm after this amount of time.
%post exploration map stored in a robot belonging to the swarm which explored the domains depicted in \autoref{fig:worlds}. 
%Each map in \autoref{fig:generated maps} are outcomes of exploring domains of various size, shapes and layouts, as given in the description of \autoref{fig:generated maps}. 
It is evident that each occupancy grid map is a reasonably accurate estimate of the corresponding environment in \autoref{fig:worlds}. {Although %in the simulations and experiments 
%we stopped the mapping procedure after the robots explored the domain for a predefined amount of time (tf seconds). 
in this work, the robots stop exploring the domain and sharing their maps after a predefined time ($t_f$ seconds), they could alternatively be programmed to autonomously halt the mapping procedure once their stored occupancy grid maps stop changing significantly when they acquire new range measurements or maps communicated by neighboring robots. }

Using a simulation of $N_R = 5$ robots exploring the  cave environment (\autoref{subfig:mycave}) as an example, we illustrate the evolution of a robot's occupancy map over time as it updates the map with new laser range sensor measurements and with the maps communicated by its neighbors.  
\autoref{fig:screenshots} shows a sequence of snapshots of a simulation for this environment. The occupancy grid maps that are constructed by one of the robots in the swarm
%, indicated by a green dotted circle, 
at the corresponding times are displayed in \autoref{fig:snapshots}. {Note that the choice of the robots' initial positions is arbitrary; the robots could ultimately produce a map of the environment starting from any initial positions, due to the diffusive nature of their exploratory random walks. \autoref{fig:collision} depicts a collision avoidance maneuver performed by the robot in the green circle: upon encountering an obstacle, the robot follows a new heading that it computes from \autoref{eqn:MI objective}.} The accompanying multimedia attachment shows the video of this simulation with the occupancy grid maps of two robots evolving over time as the simulation progresses. The video demonstrates that the robots combine their individual maps when they are close enough to communicate, and that their maps eventually converge to a common occupancy map.
%\spr{(Summarize key events in the evolution of both maps: when two robots communicate, their maps update, and the maps of both robots ultimately converge to a common map)} 
%\rag{
%The video shows that the evolution of maps in two robots while exploring the domain. 
%From the video, observe how the maps of the robots change when they communicate with other robots in the domain and eventually converging to a common map. }
%overlaid at the corners of the simulated environment. 
%This video and \autoref{fig:snapshots} 
%it is interesting to examine how 
%demonstrate how the robot's occupancy map gets updated when it receive new laser range sensor measurements and when it interacts with its neighbors. 

We quantified the degree of  consensus on the occupancy map and the coverage performance of the swarm for simulations of the campus environment in \autoref{subfig:frieburg}. After running a simulation with $N_R = 50$ robots, we represented the set of occupancy probabilities $\mathbf{\bar{P}}_{M^i}$ computed by each robot $R^i$ at time $t_f = 3600$ s as a matrix $\mathbf{P}_{M^i}$ with the dimensions of the occupancy grid map $M^i$. \autoref{subfig:map_cons} plots the normalized 2-norm $\|\mathbf{P}_{M^i}\|_2 / \max_i \|\mathbf{P}_{M^i}\|_2$, $i=1,...,50$, which consolidates each robot's set of occupancy probabilities into a single metric for comparison. The figure shows that the value of this metric has low variability across the robots, ranging at most $\sim1.25\%$ from the maximum value for all but two robots.  The occupancy grid maps of eight of the robots at time $t_f = 3600$ s are shown in \autoref{fig:map consensus}. The similarity among the eight maps indicates that consensus has been achieved. In addition, we investigated the effect of the swarm size on coverage performance by simulating swarms of $N_R =$ 10, 20, 30, 40, and 50 robots for $t_f = 3600$ s, running 10 simulations for each value of $N_R$. \autoref{subfig:map_cov} plots the percentage of the domain area that was covered by the robots after 3600 s for each swarm size $N_R$. {The percentage of the domain covered is defined as the percentage of the total number of grid cells in the domain, including cells that overlap obstacles, for which the robots have obtained distance measurements.}
%(i.e., the percentage of grid cells for which the robots obtained distance measurements)
%the robots accessed through their respective laser range sensors)
 Each red dot marks the median percent coverage over 10 simulations, and the lower and upper error bars represent the $25^{th}$ and $75^{th}$ percentiles, respectively.  The figure shows that for the same exploration time $t_f$, the percent coverage rises with increasing swarm size and saturates {to approximately $40\%$ at around $N_R = 40$. This relatively low percent coverage is due to the fact that in the campus environment, the robots cannot traverse the buildings, which function as obstacles; the percentage of the total number of grid cells that are accessible to the robots is about $40\%$.}
%asymptotic consensus on the occupancy grid map, we conducted 
 %The plots show that the norms of the occupancy grid maps of all robots are quite close. 
 %The effect of the swarm size on covering a domain having same size and layout as in \autoref{subfig:frieburg} is depicted in
%The x-axis represents the number of robots deployed for exploration and mapping. 
%The percentage coverage is calculated as the ratio of amount of area covered by the robot to the total area of the domain times $100$. 
%The error bars in the plot represent the $25^{th}$ and $75^{th}$ percentile of the data over the trials.  

We also compared our information correlated L\'evy walk (ICLW) robot exploration strategy to a standard L\'evy walk (SLW) strategy in terms of the entropy of the occupancy grid maps, defined in \autoref{eqn:map entropy}, and the percent coverage of the domain.
\autoref{fig:entropy} plots the time evolution of the resulting occupancy grid maps' entropy for three simulated environments that robots explored using both SLW and ICLW strategies. For every environment, 30 simulations were run for each exploration strategy. {\autoref{fig:autolab_levy} shows the map of the autonomy laboratory environment (\autoref{subfig:autolab}) that was generated by robots using the SLW exploration strategy, for comparison to the map obtained using the ICLW strategy in \autoref{subfig:autolab map}. } %when robots explored the domain using a standard L\'evy walk (SLW) strategy and with our exploration strategy (\autoref{sec:Exploration based on information Correlated Levy Walk}). 
%The simulations were repeated $10$ times. 
%standard L\'evy walk based exploration. 
 \autoref{fig:coverage} plots the time evolution of the percentage of each environment that was covered by the robots during the same simulations.  
%delineates the variation in domain converge over time of exploration for the two exploration strategies considered earlier. 
%By careful inspection of \autoref{fig:coverage}, we find that

 \autoref{fig:entropy} and \autoref{fig:coverage} show that when the robots perform the ICLW in the autonomy laboratory environment, the occupancy map entropy decreases more quickly and the percent coverage increases more quickly than when they perform the SLW. %information based exploration
%than when they perform the SLW strategy.
 %The figure shows that our exploration strategy performs better in covering the domain compared to a standard L\'evy walk. 
%From \autoref{fig:entropy} and \autoref{fig:coverage}, we find that, 
For the unobstructed and cave environments, the figures show that both the ICLW and SLW exploration strategies yield similar performance in terms of map entropy and percent coverage over time. Recall that these two environments, each with area 256 m$^2$, are significantly smaller than the 1200 m$^2$ autonomy laboratory.  % \spr{(Can you rephrase the following explanation in terms of the robot headings, and why the headings computed by the optimization problem are approximately uniform for small environments?)}
Since these environments are relatively small, the robots that explore them have a high rate of encounters with other robots and with the boundaries of the free space, which produces frequent changes in the robots' headings.
%and each encounter initiates a change in the robot's heading. 
This causes the distribution of robot headings in these environments to more closely resemble a uniform distribution, as in the SLW. 

\fnl{Additionally, we compared our strategy with the multi-robot mapping strategy presented in Hern{\'a}ndez \textit{et al}. \cite{hernandez2020real}. {Like our strategy, the one in \cite{hernandez2020real} uses multiple robots to generate occupancy grid maps; unlike ours, it is not an identity-free strategy.} For both strategies, we ran 10 simulations on each of four environments shown in \autoref{fig:worlds} with the same parameters used to generate the maps in \autoref{fig:generated maps}.  {In each simulation, we obtained maps from all the robots exploring the domain.}
%(grid cell error = (actual map grid cell - robot map grid cell)./actual map grid cell $\times$ 100, 
%mean absolute error is mean over the grid cells) for each trail 
%and computed the mean. This is a worst case scenario.} 
We selected the robot map with the maximum error, defined as the percent absolute error between the actual map and the robot-generated map. \autoref{fig:map error comparison} plots the median of the maximum errors obtained from the simulations, with error bars indicating the largest and smallest maximum errors over the trials. 
The results show that the performance of our strategy, in terms of the median of the maximum error, is comparable with that of the strategy presented in 
 \cite{hernandez2020real}. Our strategy produces a slightly higher median error for the unobstructed (\autoref{subfig:plain}) and cave (\autoref{subfig:mycave}) environments, but yields a smaller range of error for the University of Auckland (UoA) robotics laboratory (\autoref{subfig:uoa_robotics_lab}) and autonomy laboratory (\autoref{subfig:autolab}) environments.}
%them to have a behavior similar to the standard L\'evy walk.}
%our strategy does not significantly outperform the standard L\'evy walk exploration technique.
%This should be because these domains are small, 
 
{To investigate the effect of the L\'evy  exponent $\alpha$ on the robots' coverage of the domain, we performed simulations of $N_R=10$ robots exploring the unobstructed environment for $t_f=800$ seconds with $\alpha \in \{1.0, 1.5, 2.0, 2.5, 3.0\}$, running $10$ simulations for each value of $\alpha$. For each   $\alpha$ value, \autoref{subfig:alpha} plots the time evolution of the %\spr{simulation?, \sout{trial,}} 
%we measured the 
minimum percentage of the domain covered by the 10 robots, averaged over the 10 simulations for that $\alpha$ value.
%attained by the robots at each time instant. 
%\autoref{subfig:alpha} depicts the mean over the minimum percentage of the domain covered attained by the robots 
%over the trails at various time instants. 
%\spr{(Is this the mean over 10 trials of the minimum $\%$ of coverage over the 10 robots?)} \rag{yes} 
Interestingly, \autoref{subfig:alpha} shows that there is not a significant effect of $\alpha$ on coverage of this environment for $1\leq \alpha \leq 3$. (A detailed investigation of the invariance of coverage to $\alpha$ in this range is beyond the scope of this paper.)}
%A detailed investigation of the reason for the invariance of coverage to $\alpha$ is beyond the scope of this paper.

% we have run simulations with imperfect robot localization, in which we added zero-mean Gaussian noise with different standard deviations to the $x$ and $y$ coordinates of the robots' positions, and have included the results in the revised paper ({Figure 14(b)} and the corresponding text in Section VI, paragraph 8). These simulation results  demonstrate the robustness of our approach to localization uncertainty.
Furthermore, to test the robustness of our mapping strategy to uncertainty in robot localization, we conducted simulations in which we added zero-mean Gaussian noise with standard deviation $\sigma_p$ to the $x$ and $y$ coordinates of the robots' positions. 
For each value $\sigma_p \in \{0.5, 1.0, 1.5, 2.0\}$ meters, we ran 10 simulations on each of four environments shown in \autoref{fig:worlds} with the same parameters used to generate the maps in \autoref{fig:generated maps}. 
%The simulations were conducted for various environments shown in \autoref{fig:worlds} using their corresponding parameters used to generate the maps in \autoref{fig:generated maps}.
%For each value of $\sigma_p \in \{0.5, 1.0, 1.5, 2.0\}$ in meters, we conducted 10 simulation trails on each domain and 
%on each trail we computed the percentage mean absolute error of the map for each robot.
%\spr{(confusing:) The maximum percentage mean absolute error was selected from robots in each trial and the mean of this value over various trials for the different $\sigma_p$ values were plotted.} 
In each simulation, we obtained maps from all the robots exploring the domain. We selected the robot map with the maximum error, defined as the percent absolute error between the actual map and the robot-generated map. 
%(grid cell error = (actual map grid cell - robot map grid cell)./actual map grid cell $\times$ 100, 
%mean absolute error is mean over the grid cells) for each trail 
%and computed the mean. This is a worst case scenario.} 
\autoref{subfig:sigma} plots the mean of the maximum errors obtained from the simulations. The unobstructed environment is least affected by the noise, since it does not contain any obstacles. For the other environments, the mapping error increases with an increase in standard deviation, as expected. Observing the blue line (cave environment) in the plot, we find that the highest mapping error is only about $35\%$, even though the 2-meter standard deviation in the robot position is a substantial fraction of the environment dimensions (16 m $\times$ 16 m).  

Finally, we tested our TDA-based adaptive thresholding technique on the occupancy grid maps in \autoref{subfig:mycave map} and \autoref{subfig:autolab map}, which were generated for the cave and autonomy laboratory environments, respectively.
%in order to demonstrate usefulness of the TDA-based adaptive thresholding  on occupancy grid maps to isolate occupied grid cells from free ones described in \autoref{sec:post processing OCG}. 
%We apply the technique on maps shown \autoref{subfig:mycave map} and \autoref{subfig:autolab map} and 
The resulting barcode diagrams 
are shown in \autoref{fig:barcode}.
A topological feature is considered to be {\it persistent} if its corresponding arrow in the barcode diagram spans the entire range of the filtration parameter, $\varpi \in [0, 250]$. The single arrow in \autoref{subfig:bar_0_cave} and the four arrows that span $\varpi \in [0, 250]$ in \autoref{subfig:bar_1_cave}  correctly indicate that \autoref{subfig:mycave map} 
%the map of \autoref{subfig:mycave}
has one connected component and four obstacles (topological holes), respectively. Similarly, the single  arrows that span $\varpi \in [0, 250]$ in \autoref{subfig:bar_0_autolab} and  \autoref{subfig:bar_1_autolab}  correctly indicate that \autoref{subfig:autolab map} 
%the map of \autoref{subfig:mycave}
has one connected component and one obstacle, respectively. 
In \autoref{subfig:bar_0_cave} and \autoref{subfig:bar_1_cave}, the maximum filtration value for which all non-arrow line segments (which represent non-persistent topological features) terminate is $\varpi = 204$, meaning that grid cells with pixel intensity greater than $204$ should be considered occupied.
%Comparing the upper limits %termination points 
%of the non-arrow line segments in \autoref{subfig:bar_0_cave} and \autoref{subfig:bar_1_cave}, we find that the maximum filtration value for which all the line segments representing non-persistent topological features terminate is $204$, i.e., grid cells with pixel intensity greater than $204$ should be occupied. 
Therefore, any grid cell $m^i_j$ with an occupancy probability $\mathbb{P}_{m^i_j} \geq (204/255) = 0.8$ is designated as an occupied cell. We arrive at a similar conclusion from \autoref{subfig:bar_0_autolab} and \autoref{subfig:bar_1_autolab}.
%Again comparing the termination points of the line segments (non arrows) in \autoref{subfig:bar_0_autolab} and \autoref{subfig:bar_1_autolab}, we find that grid cells with pixel intensity greater than $204$ should be occupied. 

% \begin{figure}[!t]
% 	\centering	
% 	\includegraphics[width=.85\linewidth, height=.65\linewidth]{Figures/experiments/overhead_2.png}
% 	\caption{Overhead view of the experimental arena with three rectangular obstacles and three robots in their initial positions for each trial.}  
% 	\label{fig:arena}
% \end{figure}

\begin{figure*}[!t]
	\begin{tabular}{cc}
		\centering	
		\subcaptionbox{Physical environment\label{subfig:arena plain}}{\includegraphics[width=.45\linewidth, height=.32\linewidth]{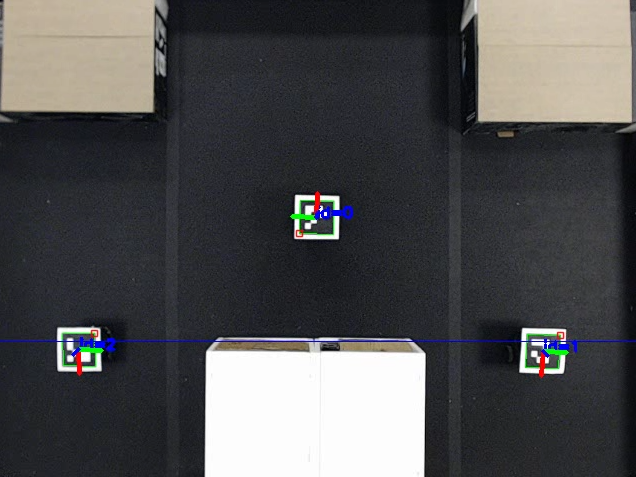}}
		~~~
		&
		~~~
		\subcaptionbox{Gazebo environment \label{subfig:gazebo cave}}{\includegraphics[width=.45\linewidth, height=.32\linewidth]{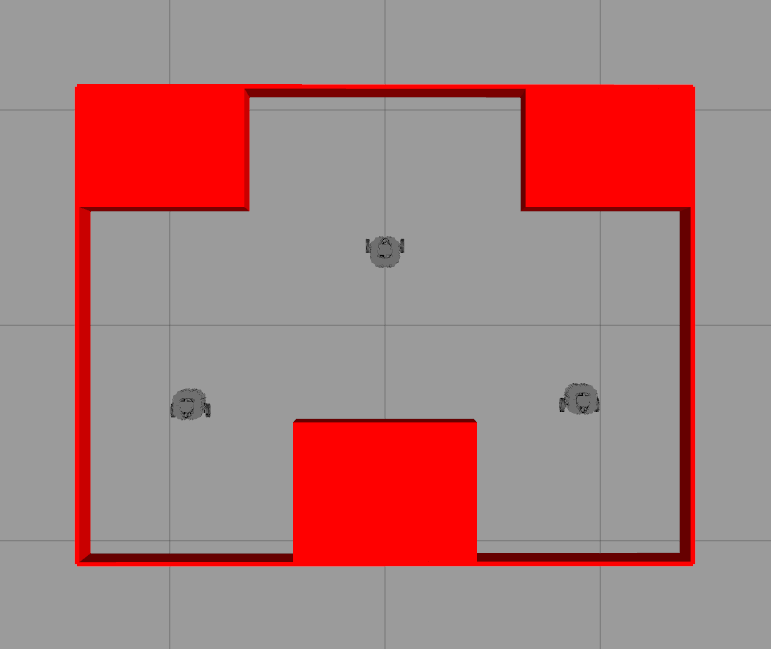}}
	\end{tabular}
	\caption{Overhead view of the physical experimental arena and \emph{Gazebo} simulation arena with three rectangular obstacles and three robots in their initial positions for each trial.}
	\label{fig:arena}
	\vspace{-6mm}
\end{figure*}

\section{ROBOT EXPERIMENTS}
\label{sec:experiments}
%To validate the theory and simulations established in the previous sections, 
We also validated our mapping approach with physical experiments using small differential-drive robots.
%equipped with Lidar sensors. 
In this section, we describe the robot testbed, software framework, and experimental results.  
%was formulated. 
%We will expand upon the equipment, obstacle setup, assumptions, and results used in the following sections. 

\subsection{Experimental Setup}
\label{subsec:experimental_setup}

Three \emph{Turtlebot3 Burger} robots were used to conduct the experiments. The robots maneuver by means of two \emph{Dynamixel XL430-W250} actuators in a differential-drive configuration. Each robot is equipped with a \emph{Raspberry Pi 3} computer for high-level computation, such as image processing, networking, and controlling a 360$^{\circ}$ Lidar sensor. Additionally, each robot has an \emph{OpenCR} controller board with a microcontroller and various sensors. For high-level control, each robot's Raspberry Pi runs the Dashing Diademata version of \emph{ros2}, the next generation of the Robot Operating System (ROS) middleware, on Ubuntu 18.04 IoT Server. The robots have a communication range of 40 cm.
%in which to share their map with others.
 
The experiments were conducted in a 2.78 m $\times$ 2.14 m arena containing three rectangular obstacles of different sizes, 
%with different lengths and widths, 
shown in \autoref{fig:arena}. All obstacles and the bordering walls of the arena had the same height, which was chosen to be high enough for the robots to detect these features with their Lidar sensors. %To \sout{monitor the robots} record the experiments, 
An \emph{ArUco} fiducial marker was fixed to the top of each robot, outside the range of the Lidar sensor, with a 3D-printed mount.
%on top of each robot without disrupting its Lidar sensor readings. 
To track the ArUco markers during the experiments, a Microsoft LifeCam camera 
%(resolution of 1920 $\times$ 1080 pixels) [CORRECT?] (zmk: No, it is 640x480)
was mounted to the ceiling over the arena and connected to a central computer.
It should be noted that the size of the arena was limited by the ceiling height, which restricted the camera's field-of-view. In addition, camera option restrictions imposed by the OS (Ubuntu) on the central computer limited the resolution of the camera image.

The code used in the simulations in Section \ref{sec:simulation} was modified for implementation on the robots. %presented by the arena. 
As in the simulations, we set $p_f = 0.1$, $p_a = 0.5$, and $p_{hit} = 0.9$ in \autoref{eqn:reflectance model} and \autoref{eqn:unreflectance model}, {and the L\'evy exponent $\alpha$ was set to $1.5$.} 
To account for the relatively small size of the arena, each robot's Lidar sensor range was limited to a maximum distance of 40 cm and a field-of-view of 180$^{\circ}$ instead of the default 2 m maximum distance and 360$^{\circ}$ field-of-view. 
%The sensor range of the Lidar was limited to 180$^{\circ}$ 
%instead of the full 360$^{\circ}$. 

%For consistency in testing, the same positions were used for each trial (see \autoref{fig:arena})
%shows the starting locations of each agent in the arena. 
%The robots may move anywhere in the arena and will avoid obstacles that are within their 180$^{\circ}$ sensing range.

\begin{figure*}[!t]
	
	\begin{tabular}{cccc}
	\hspace{-4mm}
		\centering	
		\subcaptionbox{%Iteration 1
		Maps at 0.5 min (exp) \label{subfig:exp_before}}{\includegraphics[width=.22\linewidth, height=0.5\linewidth]{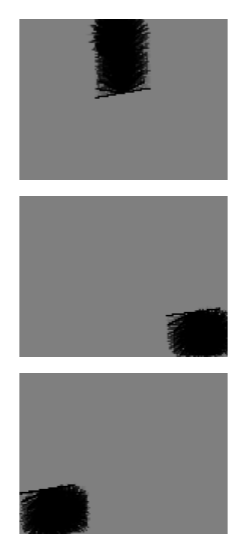}}
		
		&
		
		\subcaptionbox{%Iteration 20
		Maps at 10 min (exp)
		\label{subfig:exp_after}}{\includegraphics[width=.22\linewidth, height=0.5\linewidth]{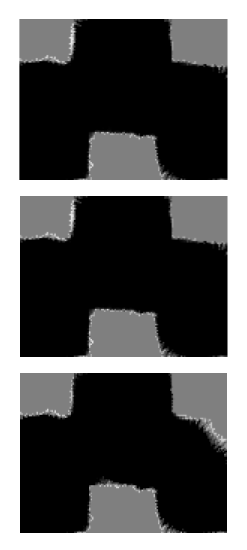}}
		
		&
		
		\subcaptionbox{%Iteration 1
		Maps at 0.5 min (sim) \label{subfig:sim_before}}{\includegraphics[width=.22\linewidth, height=0.5\linewidth]{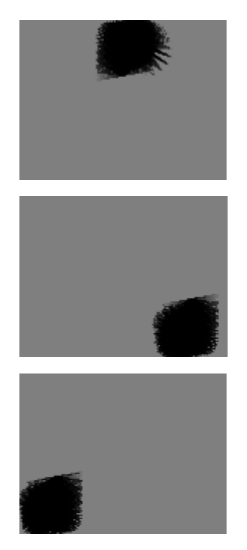}}
		
		&
		
		\subcaptionbox{%Iteration 20
		Maps at 10 min (sim)
		\label{subfig:sim_after}}{\includegraphics[width=.22\linewidth, height=0.5\linewidth]{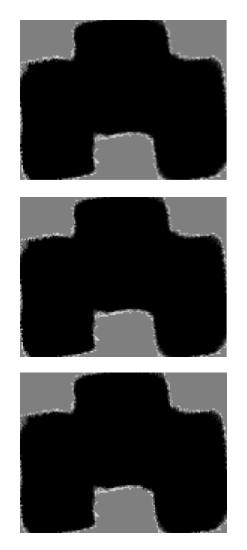}}
	\end{tabular}
	\caption{{Occupancy grid maps generated by robot $R^0$ {\it (top row)}, robot $R^1$ {\it (middle row)}, and robot $R^2$ {\it (bottom row)} at 0.5 min  and  10 min during an experimental trial {\it (left two columns)} and a \emph{Gazebo} simulation trial {\it (right two columns)}.} Robot labels are shown in \autoref{fig:arena}, detected by the overhead camera from the ArUco markers.}
	%Every 30 seconds a map is saved along. (a) are the first maps saved by the robots, and (b) are the last.}
	\label{fig:exp_map}
\end{figure*}

\begin{figure}[!t]
	
	\begin{tabular}{c}
% 		\centering	
		%\hspace{-6mm}	
		\subcaptionbox{Entropy of occupancy maps  \label{subfig:exp_entropy}}{\includegraphics[width=.98\linewidth, height=.5\linewidth]{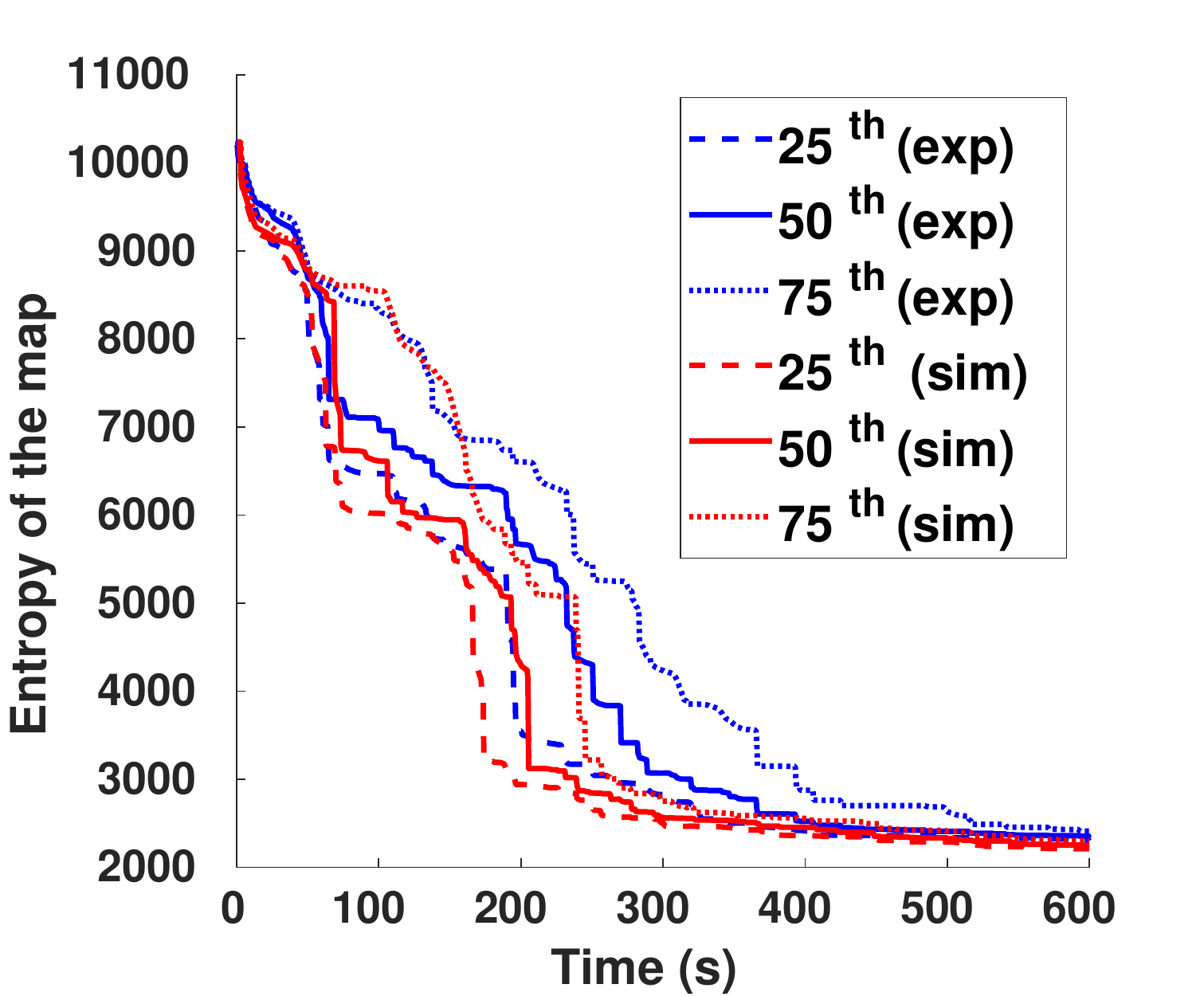}} 
		\\
		\subcaptionbox{Percentage coverage \label{subfig:exp_coverage}}{\includegraphics[width=.98\linewidth, height=.5\linewidth]{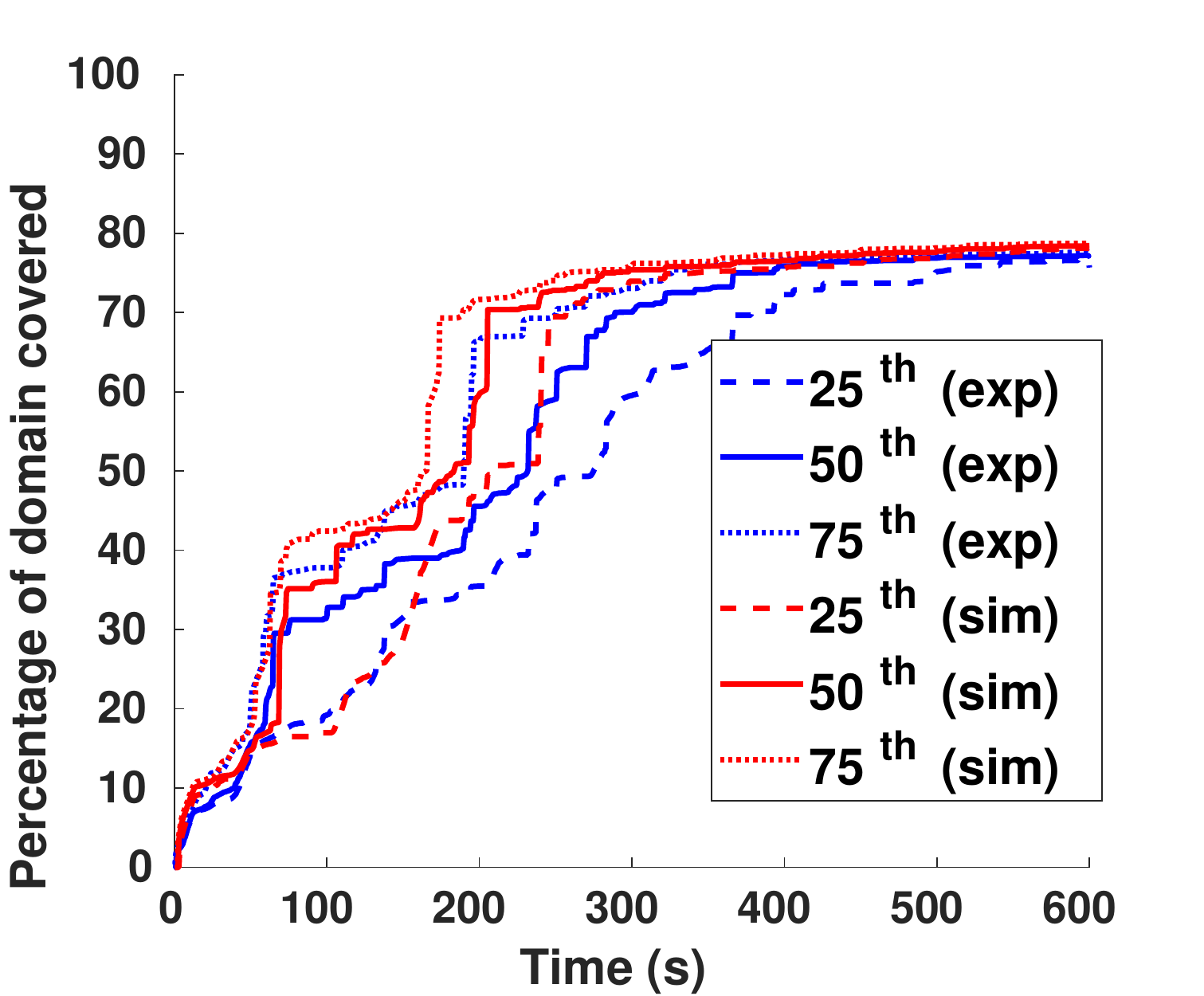}} 
	\end{tabular}
	\caption{{Time evolution during 10 experimental {\it (blue)} and simulation {\it (red)} trials of} (a) the entropy $H(M^i)$
	of occupancy maps $M^i$, $i = 1, 2, 3$, and (b) the percentage of the domain area covered by the robots.
	The plots show the $25^{th}$ percentile {\it (dashed line)}, $50^{th}$ percentile {\it (solid line)}, and $75^{th}$ percentile {\it (dotted line)} of these quantities from the 10 trials.}
	\label{fig:exp_results_combined}
	\vspace{-15mm}
\end{figure}

\subsection{Software Architecture}
\label{subsec:software_arch}

By using ros2, we were able to distribute much of the computing necessary for the experiments. To do this, ros2 utilizes the concept of \emph{nodes}, which contain code for executing specific tasks. For example, sensor information from a robot's Lidar is processed and used to construct the robot's occupancy grid map in a node. The information generated in this node can then be sent to other nodes. For our experiments, we distributed nodes across the three robots and a central computer. The robots were all given the same nodes, but were assigned unique identifiers to individualize their information.

Compared to the original ROS, ros2 does not have a \emph{ROS Master} which connects nodes that are distributed across multiple machines. Instead, each component within the system communicates using a Data Distribution Service (DDS), making the system decentralized and improving its robustness to failure. If any robot or the central computer were to fail, the remaining components will continue to work until the failing component is returned to an operational status. In this work, the central computer ran nodes that acquire information from the overhead camera and disseminate the information to the robots. The overhead camera node calculated the position and orientation of each robot from its ArUco marker. The robots require this information to build their occupancy grid maps and to identify robots that are within their communication range for sharing maps.

Each %Turtlebot 3 Burger 
robot ran %contained 
three nodes: one to transmit %send and receive 
information to and from the OpenCR controller board, a second to collect and publish information from the Lidar sensor, and a third to provide the high-level control pertaining to the L\'evy walk exploration strategy. %\zmk{This last node had a planner which sent a signal to the OpenCR board node that drove the motors.} 
All robots communicated with the central computer and with each other using the on-board \emph{WiFi} modules on the Raspberry Pi 3.

\subsection{Experimental Results}
\label{subsec:experiment_results}

Ten experimental trials were run for 10 minutes each. To minimize robot interactions early on in the experiments, the robots were placed in different parts of the arena in the same locations at the beginning of each trial (see \autoref{fig:arena}). During every trial, each robot generated an occupancy grid map and computed its occupancy map entropy and percent coverage of the domain every 0.1 s. To conserve memory on the robot, however, the occupancy grid map image was saved only every 30 s. {In addition to the physical experiments, ten \emph{Gazebo} \cite{gazebo2004} simulations were run with the same parameters as the experiments 
%sing the same experimental procedures as described previously 
in a reproduction of the experimental testbed, shown in Figure \ref{subfig:gazebo cave}. These simulations provide a basis of comparison for the performance of our approach in a noise- 
%serve to validate our approach in a noise and 
and error-free environment.}  

%During every trial, each robot stored the occupancy grid map that it generated every 0.5 min and computed its occupancy map entropy and percent coverage of the domain every 0.1 s. \zmk{one last thing}
%Entropy and coverage data were collected  \zmk{yes, but all the information was recorded at the same time. These values were not calculated after the fact. We recorded the map every 30 seconds to save space on the robot.} at a higher frequency by each robot.
%The robots were placed in the same initial positions at the start of each trial to allow comparison of data across trials. 
\autoref{fig:exp_map} shows the occupancy grid maps of each robot at 0.5 min and 10 min for an experimental  trial and a simulation trial. The maps at the end of the trial (10 min) approximate the actual environment, and the similarity of these maps indicates that the robots have reached consensus. The accompanying multimedia attachment shows the video of the experimental trial with the time evolution of the robots' occupancy grid maps over the entire trial.
%each robot's first and last iteration of a trial run. 
\autoref{subfig:exp_entropy} and \autoref{subfig:exp_coverage} plot the time evolution of the occupancy grid maps' entropy 
%stored in a robot
and the percent coverage of the domain, respectively, over all trials.
%shown in \autoref{fig:arena} 
%over time of exploration respectively.
From \autoref{subfig:exp_entropy}, we see that the entropy of the occupancy maps is minimized 
%reduced to a minimum 
within about $500$ s 
%of exploration time
in all trials, indicating that the robots have achieved consensus on a map of the domain with minimum uncertainty. %around this time. %Also, comparing \autoref{subfig:exp_entropy} and
Moreover, \autoref{subfig:exp_coverage} shows that the percentage of the domain covered by the robots reaches its maximum value around this time in all trials. 
%when their occupancy grid maps attain the minimum entropy, 
%indicating that the robots have achieved consensus on a map of the domain which has minimum uncertainty.
{The figure %\ref{fig:exp_results_combined} also displays 
shows that the results from the ten Gazebo simulations are comparable to the results from the physical experiments, and their entropy and percent coverage values converge to nearly the same value at the final time. 
%In both cases, the values converge to nearly the same final result over time.
This similarity suggests that our approach is robust to noise and errors that occur in physical environments.}

\section{Conclusion}
\label{sec:conclusion}
In this work, we developed a novel distributed approach to generating an occupancy grid map of an unknown environment using a swarm of robots with local communication. The approach is scalable with the number of robots and robust to robot errors and failures. We also presented a robot exploration strategy, referred to as an information correlated L\'evy walk, 
that directs L\'evy-walking robots to regions that maximize the their information gain. In addition, we demonstrated that a technique based on topological data analysis, which we used in our earlier works to construct topological maps of unknown environments, can also be used for adaptive thresholding of occupancy grid maps. We validated our mapping methodology through simulations and physical robot experiments on domains with different geometries and sizes. %Overall, we believe this to be an effective distributed mapping approach for robotic swarms due to its scalability and robustness to robot failures. 

One direction for future work is to extend our mapping approach to the case where the robots have no global localization or are only capable of weak localization, which we define as %localization
pose estimation with bounded uncertainty, using a received signal strength indicator (RSSI) device \cite{Xie2015} or ultra-wideband (UWB) sensors \cite{Prorok:2014,alarifi2016ultra}. This will allow our approach to be applied in GPS-denied environments. 
%The extension of our mapping approach to the case of weak localization is a direction for future work. 
%weak or no global localization information and also analyze the technique to understand the kind of guarantee our strategy can provide in a weak localization setting. This could become a viable technology in the future as our approach can then be coupled with existing ultra wide band (UWB) or received signal strength indicator (RSSI) technologies to perform mapping on large complex environments.
Another future objective is to derive the partial differential equation model that describes the spatiotemporal evolution of the swarm population dynamics as the robots perform information correlated L\'evy walks. This will enable an analytical investigation of the performance of our exploration strategy.
%\rag{The resulting PDE model can then be used understand the diffusive property of the robot population analytically.}

% or


\appendices
\section{Derivation of formula to compute \texorpdfstring{$\mathbf{I}[\mathbf{c}^{i,a}_\tau; z^{i,a}_{\tau}]$}{MI}}
\label{app: formula derivation}

By definition, 
\begin{align*}
\mathbf{I}[\mathbf{c}^{i,a}_\tau; z^{i,a}_{\tau}] = \int_{ z\in z^{i,a}_{\tau}}\sum_{c\in\mathbf{c}^{i,a}_\tau}\mathbb{P}(c,z)\log_2 \left( \frac{\mathbb{P}(c,z)}{\mathbb{P}(c)\mathbb{P}(z)}\right) dz
\end{align*}
 
 This expression can be rewritten as,
 \begin{eqnarray}
 \mathbf{I}[\mathbf{c}^{i,a}_\tau; z^{i,a}_{\tau}] &{}={}& \int_{ z\in z^{i,a}_{\tau}}\sum_{c\in\mathbf{c}^{i,a}_\tau}\mathbb{P}(c,z)\log_2\left( \mathbb{P}(z,c)\right)dz \nonumber \\ &&{-}\: \int_{ z\in z^{i,a}_{\tau}}\sum_{c\in\mathbf{c}^{i,a}_\tau}\mathbb{P}(c,z)\log_2\left(\mathbb{P}(z)\mathbb{P}(c)\right)dz \nonumber
 \end{eqnarray}
 
 Define
 \begin{align*}
 \mathbf{P1}=\int_{ z\in z^{i,a}_{\tau}}\sum_{c\in\mathbf{c}^{i,a}_\tau}\mathbb{P}(c,z)\log_2\left( \mathbb{P}(z,c)\right)dz
 \end{align*}
 and
 \begin{align*}
\mathbf{P2}=\int_{ z\in z^{i,a}_{\tau}}\sum_{c\in\mathbf{c}^{i,a}_\tau}\mathbb{P}(c,z)\log_2\left(\mathbb{P}(z)\mathbb{P}(c)\right)dz
 \end{align*}
 
 %Now we focus on  $\mathbf{P1}$, 
 Expanding $\mathbf{P1}$  yields, 
 \begin{eqnarray}
 \mathbf{P1} &{}={}& \int_{ z\in z^{i,a}_{\tau}}\sum_{p=0}^{|\mathbf{c}^{i,a}_\tau|}( \mathbb{P}(z|e_p)\mathbb{P}(e_p)\log_2\left(\mathbb{P}(z|e_p) \right) \nonumber\\ && ~~~~~~~~~~~~{+}~\: \mathbb{P}(z|e_p) \mathbb{P}(e_p)\log_2\left(\mathbb{P}(e_p) \right))dz \nonumber
 \end{eqnarray}
 Rearranging this expression, we obtain
 \begin{eqnarray}
 \mathbf{P1} &{}={}&  \sum_{p=0}^{|\mathbf{c}^{i,a}_\tau|} \mathbb{P}(e_p) \int_{ z\in z^{i,a}_{\tau}}\mathbb{P}(z|e_p)\log_2(\mathbb{P}(z|e_p))dz \nonumber \\ &&{+}\: \sum_{p=0}^{|\mathbf{c}^{i,a}_\tau|} \mathbb{P}(e_p)\log_2(\mathbb{P}(e_p))\int_{ z\in z^{i,a}_{\tau}} \mathbb{P}(z|e_p)dz \nonumber
 \end{eqnarray}
Using the fact that $\int_{ z\in z^{i,a}_{\tau}} \mathbb{P}(z|e_p)dz = 1$
%is unity 
and substituting $\mathbb{P}(z|e_k)$, defined as the Gaussian forward measurement model in \autoref{eqn:forward sensor model}, into the above equation, we find the following expression for $\mathbf{P1}$ after simplification:
 \begin{equation}
 \label{eqn:part1 simplified}
 \mathbf{P1} = -\log_2(\sqrt{2\pi}\sigma) -0.5 + \sum_{p=0}^{|\mathbf{c}^{i,a}_\tau|} \mathbb{P}(e_p)\log_2(\mathbb{P}(e_p))  \nonumber
 \end{equation}
 
 The expression for $\mathbf{P2}$ can be split into the sum of the following two terms, which we denote as $\mathbf{P2}_a$ and $\mathbf{P2}_b$, respectively:
 \begin{eqnarray}
 \mathbf{P2} &{}={}& \int_{ z\in z^{i,a}_{\tau}}\sum_{c\in\mathbf{c}^{i,a}_\tau}\mathbb{P}(c,z)\log_2(\mathbb{P}(c))dz \nonumber \\ 
 &&{+}\:\int_{ z\in z^{i,a}_{\tau}}\sum_{c\in\mathbf{c}^{i,a}_\tau} \mathbb{P}(c,z)\log_2(\mathbb{P}(z))dz \nonumber
 \end{eqnarray}
 $\mathbf{P2}_a$ can be expanded as, 
 \begin{align*}
     \mathbf{P2}_a = \int_{ z\in z^{i,a}_{\tau}}\sum_{p=0}^{|\mathbf{c}^{i,a}_\tau|}\mathbb{P}(z|e_p)\mathbb{P}(e_p)\log_2(\mathbb{P}(e_p))dz \nonumber \\
 = \sum_{p=0}^{|\mathbf{c}^{i,a}_\tau|} \mathbb{P}(e_p)\log_2(\mathbb{P}(e_p))\int_{ z\in z^{i,a}_{\tau}}\mathbb{P}(z|e_p)dz
 \end{align*}
Since $\sum_{c\in\mathbf{c}^{i,a}_\tau} \mathbb{P}(c,z)$ reduces to $\mathbb{P}(z)$ (by marginalization over $c$), $\mathbf{P2}_b$ can be expressed as,
 \begin{align*}
 \mathbf{P2}_b = \int_{ z\in z^{i,a}_{\tau}}\mathbb{P}(z)\log_2(\mathbb{P}(z))dz
 \end{align*} 
 
 Finally, using our derived equations for $\mathbf{P1}$, $\mathbf{P2}$ and the fact that $\int_{ z\in z^{i,a}_{\tau}}\mathbb{P}(z|e_p)dz = 1$, we obtain the following formula:
 \begin{equation}
 \label{eqn:app: MI formula}
 \mathbf{I}[\mathbf{c}^{i,a}_\tau; z^{i,a}_{\tau}] =  -\int_{ z\in z^{i,a}_{\tau}}\mathbb{P}(z)\log_2(\mathbb{P}(z))dz - \log_2(\sqrt{2\pi}\sigma) -0.5 \nonumber
 \end{equation}
 \qed

\section{Proof of \autoref{thm: consensus}}
\label{app: proof of consensus theorem}

\begin{proof}

We abbreviate $u(m^i_j, \mathbf{x}^i_k,\mathbf{z}^i_k)$ as $u^i_j(k)$  and define $\mathbb{L}x = -\log_2(x)$ for $x > 0$.
%Let $\mathbb{L}\mathbb{P}_{m^i_j}(k + 1)$ and $\mathbb{L}u_e^{i,j}(k) $ denote $-\log_2(\mathbb{P}_{m^i_j}(k + 1)) $ and $-\log_2(u(m^i_j, \mathbf{x}^i_k,\mathbf{z}^i_k))$, respectively.
Taking the negative logarithm of both sides of \autoref{eqn:grid cell update rule}, we obtain: 
\begin{align}
\label{eqn: app:log of grid update}
 \mathbb{L}\mathbb{P}_{m^i_j}(k + 1) = \sum_{\hat{n} \in \mathbb{N}^i_{k} \cup i}a_{i \hat{n}}(k)\left( \mathbb{L}\mathbb{P}_{m^{\hat{n}}_j}(k) \right) + \mathbb{L} u^i_j(k)
\end{align}
Let $\mathbb{L}\mathbb{P}_{m_j}[k] = [\mathbb{L}\mathbb{P}_{m^1_j}(k)~...~\mathbb{L}\mathbb{P}_{m^{N_R}_j}(k)]^T$ and $\mathbb{L}\mathbf{u}_{j}[k] = [\mathbb{L} u^1_j(k)~...~\mathbb{L} u^{N_R}_j(k)]^T$.
%If we stack the variables $\mathbb{L}\mathbb{P}_{m^i_j}(k)$ and $\mathbb{L} u_e^{i,j}(k)$ for every robot index $i \in \{1, ..., N_R\}$ into column vectors and denote as $ \mathbb{L}\mathbb{P}_{m_j}[k] $ and $ \mathbb{L}\mathbf{u}_e^{j}[k]$ respectively
Then  \autoref{eqn: app:log of grid update} can be written in the following form:
\begin{align}
\label{eqn:app: matrix update rule}
\mathbb{L}\mathbb{P}_{m_j}[k + 1] ~=~ \mathbb{A}[k] \left(\mathbb{L}\mathbb{P}_{m_j}[k] \right) + \mathbb{L}\mathbf{u}_{j}[k], 
\end{align} 
where $\mathbb{A}[k]$ is the adjacency matrix %\cite{GodsilRoyle2001} 
of the time-varying robot communication  graph $\mathbb{G}(k)$, as defined in \autoref{subsec:Protocol for occupancy grid sharing}.

Given an initial time step $k_0 \geq 0$, we define $\Phi_ {\mathbb{A}}[k, k_0] = \mathbb{A}[k]\mathbb{A}[k-1]\cdots\mathbb{A}[k_0]$. Then the information dynamics described in \autoref{eqn:app: matrix update rule} can be expanded as: 
\begin{align}
\label{eqn:app: matrix update rule with phi}
\mathbb{L}\mathbb{P}_{m_j}[k + 1] = \Phi_ {\mathbb{A}}[k, 0] (\mathbb{L}\mathbb{P}_{m_j}[0]) + \sum_{d \in \mathbf{d}}\Phi_ {\mathbb{A}}[k, d] (\mathbb{L}{\mathbf{u}_{j}[k]}) %\mathbf{u}_e^{j}[d].
\end{align}

If  \hyperref[ass:graph connectedness]{Assumption~\ref*{ass:graph connectedness}} holds and $\mathbb{A}[k]$ is a doubly stochastic matrix for each $k$, then by \cite[Theorem 1]{Kingston2006}, we have that
\begin{align*}
\lim_{k \rightarrow \infty} \Phi_ {\mathbb{A}}[k, k_0] = \frac{1}{N_R} \mathbf{1}\mathbf{1}^{T},
\end{align*}
where $\mathbf{1} \in \mathbb{R}^{N_R}$ is a column vector of ones. 
%In addition, according to 
By \cite[Theorem 11.6]{FB-LNS}, $\Phi_ {\mathbb{A}}[k, k_0]$ converges exponentially to $ \frac{1}{N_R} \mathbf{1}\mathbf{1}^{T}$.  \hyperref[ass: pairwise interaction]{Assumption~\ref*{ass: pairwise interaction}} ensures that $\mathbb{A}[k]$ is doubly stochastic for all $k$. 
%pairwise interaction assumption
%\hyperref[ass: pairwise interaction]{Assumption~\ref*{ass: pairwise interaction}} is one way to ensure that the adjacency matrix corresponding to $\mathbb{G}(k)$ is doubly stochastic.
Taking the limit of \autoref{eqn:app: matrix update rule with phi} as $k \rightarrow \infty$ and using the above result yields,
\begin{eqnarray}
\lim_{k \rightarrow \infty} \mathbb{L}\mathbb{P}_{m_j}[k] &{}={}&  \mathbf{1}\left(\frac{1}{N_R}\mathbf{1}^{T} (\mathbb{L}\mathbb{P}_{m_j}[0])\right) \nonumber\\
&&{+}~ \: \mathbf{1}\left(\sum_{d \in \mathbf{d}}\frac{1}{N_R}\mathbf{1}^{T} (\mathbb{L}{\mathbf{u}_{j}[k]}) \right)
%\mathbf{u}_e^{j}[d]\right) 
\nonumber
\end{eqnarray}

 The product of $\frac{1}{N_R}\mathbf{1}^{T}$ and a column vector with $N_R$ non-negative elements is equal to the arithmetic mean of the elements of the vector.
 %the corresponding column vector. 
Denoting $<\cdot>$ as the arithmetic mean operator, we can therefore rewrite the above equation as: % in compact form as,
\begin{align}
\label{eqn:app:vector_consensus}
\lim_{k \rightarrow \infty} \mathbb{L}\mathbb{P}_{m_j}[k] ~=~ \mathbf{1} \left\langle \mathbb{L}\mathbb{P}_{m_j}[0]\right\rangle  
+ \mathbf{1} \sum_{d \in \mathbf{d}}\left\langle\mathbb{L}\mathbf{u}_{j}[k]
%\mathbf{u}_e^{j}[d]
\right\rangle  
\end{align}
 \hyperref[ass: finite support]{Assumption~\ref*{ass: finite support}} is required to ensure the convergence of the sum $\sum_{d \in \mathbf{d}}\left\langle\mathbb{L}{\mathbf{u}_{j}[k]}
 %\mathbf{u}_e^{j}[d]
 \right\rangle$. 
 %\hyperref[ass: finite support]{Assumption~\ref*{ass: finite support}} is used to establish convergence of a sequence summation. 

  By \autoref{eqn:app:vector_consensus}, 
  the limit of the $i^{th}$ element of $\mathbb{L}\mathbb{P}_{m_j}[k]$ as $k \rightarrow \infty$ is given by:
  %we are interested in 
  %\spr{the expression for each component of $\lim_{k \rightarrow \infty} \mathbb{L}\mathbb{P}_{m_j}[k]$, which is expressed as:}
% analyzing \autoref{eqn:app:vector_consensus} component-wise. Therefore, expressing  \autoref{eqn:app:vector_consensus} component-wise  yields,
\begin{align*}
\lim_{k \rightarrow \infty} \mathbb{L}\mathbb{P}_{m^i_j}(k) ~=~   \left\langle \mathbb{L}\mathbb{P}_{m_j}[0]\right\rangle  
+  \sum_{d \in \mathbf{d}}\left\langle\mathbb{L} {\mathbf{u}_{j}[d]}
%\mathbf{u}_e^{j}[d]
\right\rangle  
\end{align*}
Since $\mathbf{d}$ is a finite set and $\Phi_ {\mathbb{A}}[k, k_0]$ converges exponentially to $ \frac{1}{N_R} \mathbf{1}\mathbf{1}^{T}$, the above equation converges exponentially.  %\rag{From Equation 19, it is easy to see that if $\Phi_ {\mathbb{A}}[k, k_0]$ converges exponentially then, the above equation also converges exponentially.}

Finally, by taking the exponential of both sides of the above equation, we obtain the desired result {(\autoref{eqn: asymptotic result}):}
\begin{align*}
\lim_{k \rightarrow \infty} \mathbb{P}_{m^i_j}(k) ~=~  <\mathbb{P}_{m_j}[0]>_{gm} \hspace{1mm} \cdot \hspace{1mm} \prod_{d \in \mathbf{d}} <\mathbf{u}_j[d]>_{gm}
\end{align*}

For inaccessible grid cells $m^i_a$, $\mathbf{u}_a[d] = \mathbf{1}$ for all $d$ since none of the robots' %useful 
 measurements provide any information about the occupancy probability of these cells. Therefore, %in this scenario
\autoref{eqn: asymptotic result} reduces to
\begin{align}
   \lim_{k \rightarrow \infty} \mathbb{P}_{m^i_{\bar{a}}}(k) ~=~  <\mathbb{P}_{m_{\bar{a}}}[0]>_{gm}.
\end{align}

Now we consider the set of  accessible grid cells. Let the elements of the set $\mathbf{d}$ be denoted by $d_1, d_2, \cdots, d_\gamma$, where $\gamma$ is the cardinality of $\mathbf{d}$. Then, 
\begin{eqnarray}
    \prod_{d \in \mathbf{d}} <\mathbf{u}_a[d]>_{gm}  = & \nonumber \\ 
     <\mathbf{u}_a[d_1]>_{gm} ~~ \cdot & <\mathbf{u}_a[d_2]>_{gm} ~~\cdots~~ <\mathbf{u}_a[d_\gamma]>_{gm} \nonumber
\end{eqnarray}
Using the definition of geometric mean, the above equation can be written as, 
\begin{align*}
    \prod_{d \in \mathbf{d}} <\mathbf{u}_a[d]>_{gm} = \sqrt[N_R]{\prod_{q = 1}^{N_R} \mathbf{u}_a[d_1](q)  \cdots \prod_{q = 1}^{N_R} \mathbf{u}_a[d_\gamma](q)}~,
\end{align*}
where $\mathbf{u}_a[d_1](q)$  is the $q^{th}$ element of the vector $\mathbf{u}_a[d_1]$. The right-hand side of the above equation contains a product of $N_R \gamma$ positive real numbers. Since under \hyperref[ass: finite support]{Assumption~\ref*{ass: finite support}A}, all elements of this product are equal to one except for the $N_R$ values $\bar{u}_a^1, ..., \bar{u}_a^{N_R}$, 
%there are only $N_R$ elements which are not equal to one, 
the above equation can be reduced to 
\begin{align*}
    \prod_{d \in \mathbf{d}} <\mathbf{u}_a[d]>_{gm} ~=~ \sqrt[N_R]{\bar{u}^1_a \cdot \bar{u}^2_a ~\cdots~ \bar{u}^{N_R}_a}.
\end{align*}
Therefore,
\begin{align*}
    \prod_{d \in \mathbf{d}} <\mathbf{u}_a[d]>_{gm} ~=~ <[\bar{u}^1_a ~ \cdots ~ \bar{u}^{N_R}_a]^T>_{gm}.
\end{align*}
Combining the above result with \autoref{eqn: asymptotic result} yields,
\begin{align*}
    \lim_{k \rightarrow \infty} \mathbb{P}_{m^i_a}(k) ~=~  <\mathbb{P}_{m_a}[0]>_{gm} \hspace{1mm} \cdot \hspace{1mm} <[\bar{u}^1_a ~ \cdots ~ \bar{u}^{N_R}_a]^T>_{gm}.
\end{align*}

\end{proof}
%\section*{Acknowledgment}

\ifCLASSOPTIONcaptionsoff
  \newpage
\fi

%\bibliographystyle{IEEEtran}
%\bibliography{reference.bib,IEEEabrv.bib}


\begin{IEEEbiography}[{\includegraphics[width=1in,height=1.25in,clip,keepaspectratio]{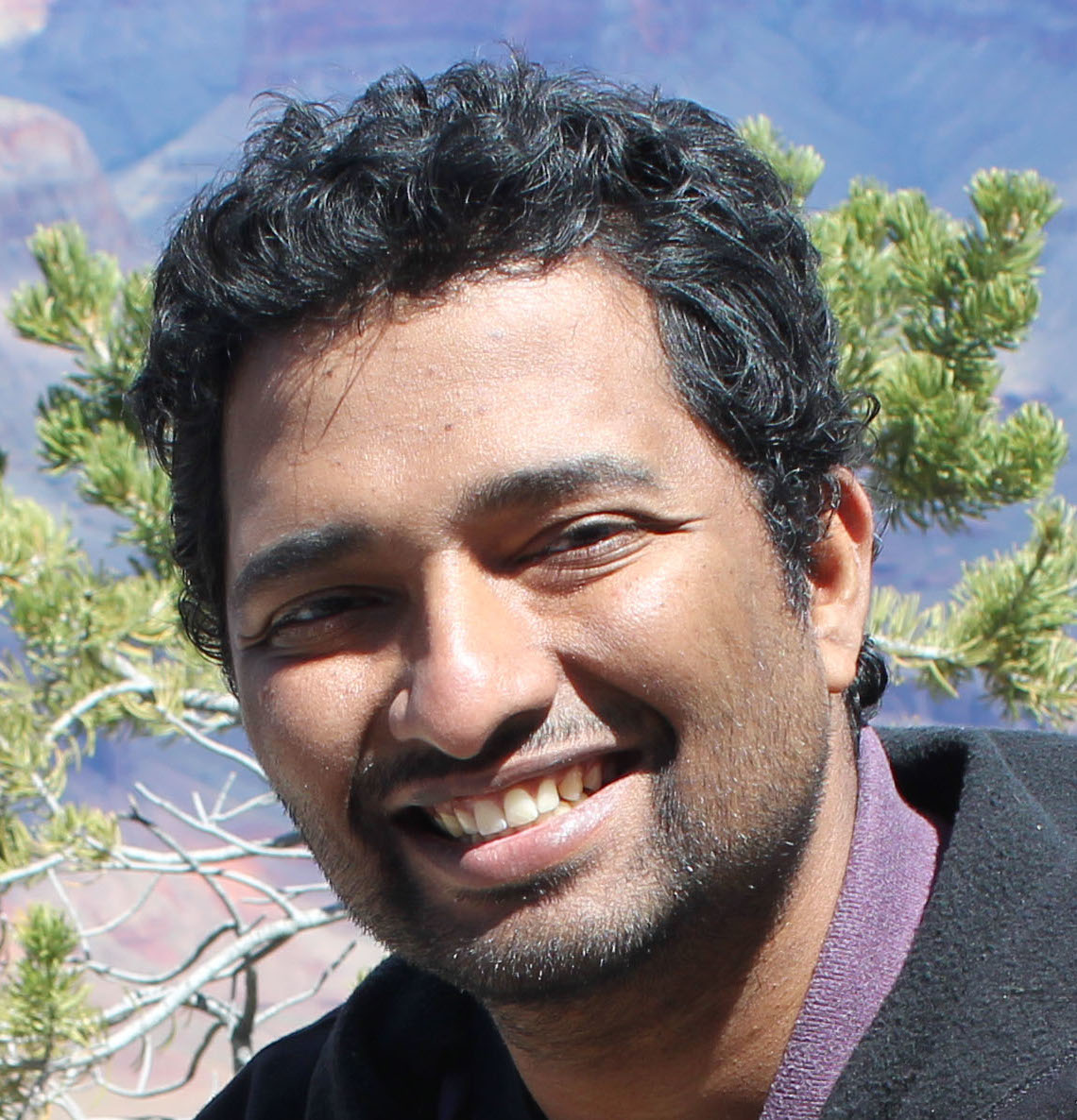}}]{Ragesh Kumar Ramachandran} received the B.Tech degree in civil engineering from National Institute of Technology Calicut, India, in 2011, and the M.S. and Ph.D. degrees in mechanical engineering from Arizona State University, Tempe, AZ, USA, in 2018. Since 2018, he has been a postdoctoral researcher in computer science at the University of Southern California. His current research interests include the design and analysis of algorithms for solving problems in robotic swarms and multi-robot systems, network and graph theory, optimal and nonlinear control theory, and applied topology and differential geometry.
\end{IEEEbiography}

\begin{IEEEbiography}[{\includegraphics[width=1in,height=1.25in,clip,keepaspectratio]{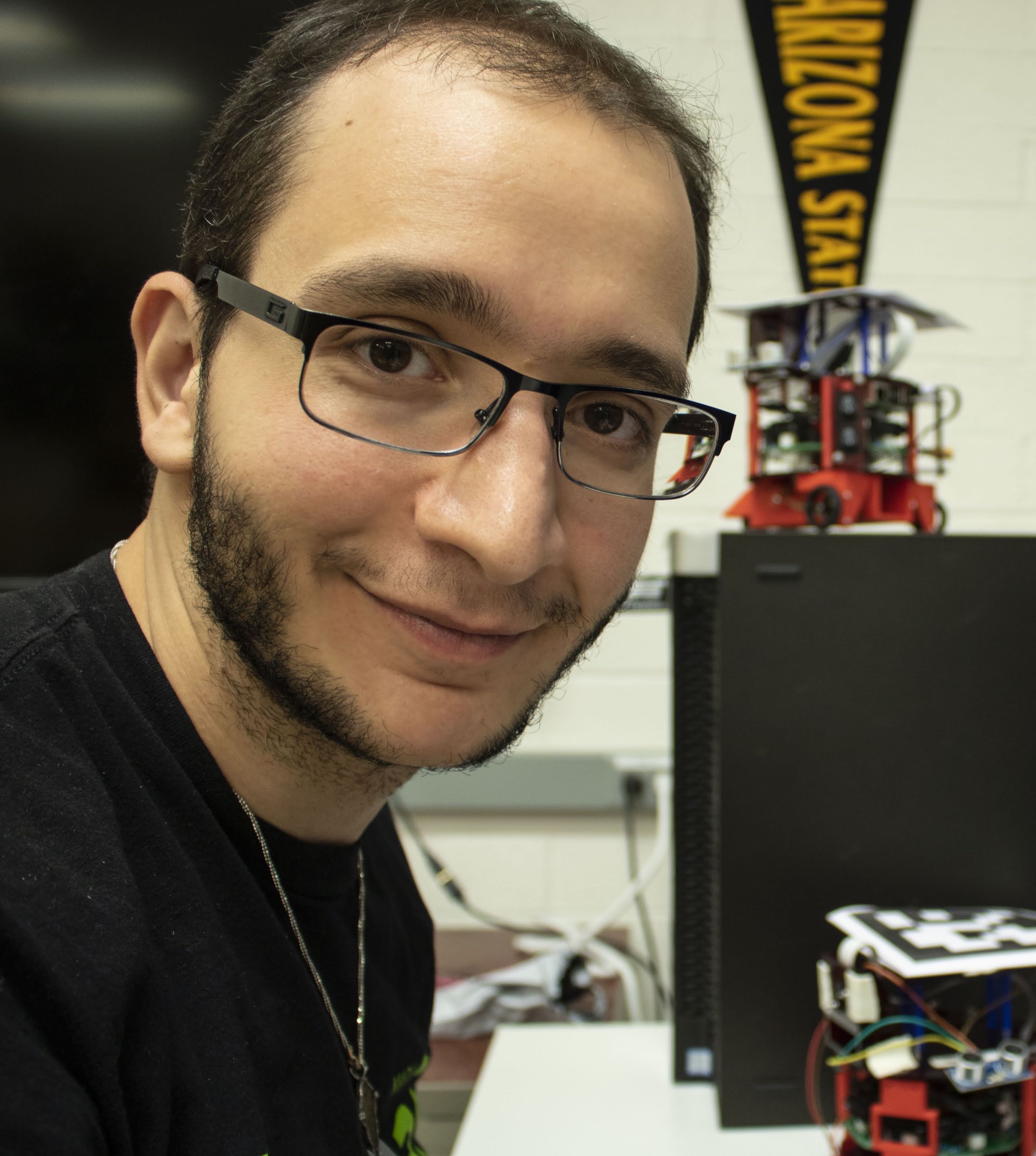}}]{Zahi Kakish} received the B.S. degree in mechanical polymer engineering from the University of Akron, Akron, OH, in 2013. He received the M.S. degree in mechanical engineering from the University of Akron in 2015. Currently, he is a Ph.D. student in mechanical engineering at Arizona State University, where his research interests include swarm intelligence, human-swarm interaction, and multi-agent reinforcement learning.
\end{IEEEbiography}

\begin{IEEEbiography}[{\includegraphics[width=1in,height=1.25in,clip,keepaspectratio]{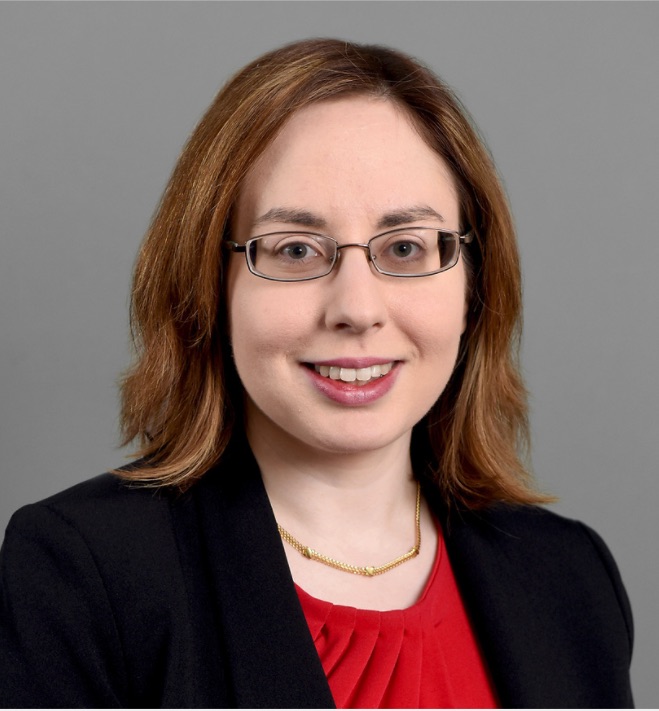}}]{Spring Berman} (M'07) received the M.S.E. and Ph.D. degrees in mechanical engineering and applied mechanics from the University of Pennsylvania in 2008 and 2010, respectively. From 2010 to 2012, she was a postdoctoral researcher in computer science at Harvard University. Since 2012, she has been a professor of mechanical and aerospace engineering at Arizona State University. Her research focuses on the analysis of behaviors in biological and engineered collectives and the synthesis of control strategies for robotic swarms. She received the 2014 DARPA Young Faculty Award and the 2016 ONR Young Investigator Award.
\end{IEEEbiography}


%\end{IEEEbiography}

%% insert where needed to balance the two columns on the last page with
%%\newpage


%\vfill


\end{document}